\documentclass{article}

\usepackage[T1]{fontenc}  %
\usepackage{lmodern}      %
\usepackage{amsmath}
\usepackage{amssymb}
\usepackage{graphicx}
\usepackage{array}
\usepackage{booktabs}
\usepackage{placeins}     %
\usepackage[font=small,labelfont=bf]{caption}
\usepackage[round,semicolon]{natbib}  %
\usepackage{xcolor}
\usepackage{hyperref}
\usepackage{doi}  %
\definecolor{linkblue}{rgb}{0 0.4235 0.6392}
\hypersetup{colorlinks, linkcolor=linkblue, citecolor=linkblue, urlcolor=linkblue,
  pdftitle={Prune, Don't Merge: Expert Compression of Mixture-of-Experts LLMs on Multilingual Production Tasks},
  pdfauthor={Tomas Rasymas, Povilas Kvedaras}}
\newif\ifsamples
\IfFileExists{dataset_samples_preamble.tex}{\usepackage{listings}
\usepackage{textcomp}%
\usepackage{tcolorbox}
\tcbuselibrary{breakable}
\definecolor{samplebreak}{gray}{0.6}%

\newtcolorbox{samplebox}[2][Anonymised production example]{%
  breakable,
  colback=black!2, colframe=black!45, colbacktitle=black!10, coltitle=black,
  fonttitle=\sffamily\bfseries\footnotesize,
  title={#2\hfill{\mdseries\itshape #1}},
  boxrule=0.4pt, arc=1pt,
  left=3pt, right=3pt, top=2pt, bottom=2pt, toptitle=1pt, bottomtitle=1pt,
  before skip=4pt, after skip=6pt,
}

\lstset{
  basicstyle=\ttfamily\scriptsize,
  breaklines=true,
  breakatwhitespace=false,
  columns=fullflexible,
  keepspaces=true,
  showstringspaces=false,
  upquote=true,
  aboveskip=0pt, belowskip=0pt,
  xleftmargin=0pt, xrightmargin=0pt,
  frame=none,
  postbreak=\mbox{\textcolor{samplebreak}{$\hookrightarrow$}\space},
  extendedchars=true,
  literate=
    {á}{{\'a}}1 {é}{{\'e}}1 {í}{{\'i}}1 {ó}{{\'o}}1 {ú}{{\'u}}1
    {Á}{{\'A}}1 {É}{{\'E}}1 {Í}{{\'I}}1 {Ó}{{\'O}}1 {Ú}{{\'U}}1
    {à}{{\`a}}1 {è}{{\`e}}1 {ì}{{\`i}}1 {ò}{{\`o}}1 {ù}{{\`u}}1
    {â}{{\^a}}1 {ê}{{\^e}}1 {î}{{\^i}}1 {ô}{{\^o}}1 {û}{{\^u}}1
    {ä}{{\"a}}1 {ë}{{\"e}}1 {ï}{{\"i}}1 {ö}{{\"o}}1 {ü}{{\"u}}1
    {Ä}{{\"A}}1 {Ö}{{\"O}}1 {Ü}{{\"U}}1 {ß}{{\ss}}1
    {ç}{{\c c}}1 {Ç}{{\c C}}1 {ñ}{{\~n}}1 {Ñ}{{\~N}}1 {ã}{{\~a}}1 {õ}{{\~o}}1,
}
\samplestrue}{}

\makeatletter
\newcommand{\V}[1]{\ifcsname V@#1\endcsname\csname V@#1\endcsname\else
  \textbf{\textcolor{red}{??#1??}}\PackageWarning{numbers}{Undefined value #1}\fi}
\makeatother
\providecommand{\defV}[2]{\expandafter\def\csname V@#1\endcsname{#2}}
\defV{n-rows}{2{,}000}
\defV{n-products}{5}
\defV{n-models}{10}
\defV{calib-tokens-q}{2.96}
\defV{calib-tokens-n}{3.02}
\defV{calib-samples}{5{,}666}
\defV{langs-usage}{19}
\defV{enshare-usage}{53}
\defV{langs-brand}{24}
\defV{enshare-brand}{24}
\defV{langs-categ}{33}
\defV{enshare-categ}{14}
\defV{langs-domains}{17}
\defV{enshare-domains}{17}
\defV{langs-reach}{35}
\defV{enshare-reach}{17}
\defV{langs-min}{17}
\defV{langs-max}{35}
\defV{medtok-q-usage}{23}
\defV{medtok-n-usage}{23}
\defV{medtok-q-brand}{24}
\defV{medtok-n-brand}{19}
\defV{medtok-q-categ}{11}
\defV{medtok-n-categ}{10}
\defV{medtok-q-domains}{20}
\defV{medtok-n-domains}{28}
\defV{medtok-q-reach}{2{,}724}
\defV{medtok-n-reach}{1{,}520}
\defV{val-q-reap25-usage}{98.5}
\defV{d-q-reap25-usage}{\ensuremath{-}0.2}
\defV{dlo-q-reap25-usage}{\ensuremath{-}0.6}
\defV{dhi-q-reap25-usage}{0.0}
\defV{ret-q-reap25-usage}{99.7}
\defV{val-q-reap50-usage}{97.8}
\defV{d-q-reap50-usage}{\ensuremath{-}1.0}
\defV{dlo-q-reap50-usage}{\ensuremath{-}1.5}
\defV{dhi-q-reap50-usage}{\ensuremath{-}0.4}
\defV{ret-q-reap50-usage}{99.0}
\defV{val-q-ream25-usage}{98.2}
\defV{d-q-ream25-usage}{\ensuremath{-}0.5}
\defV{dlo-q-ream25-usage}{\ensuremath{-}1.0}
\defV{dhi-q-ream25-usage}{0.0}
\defV{ret-q-ream25-usage}{99.5}
\defV{val-q-ream50-usage}{97.1}
\defV{d-q-ream50-usage}{\ensuremath{-}1.7}
\defV{dlo-q-ream50-usage}{\ensuremath{-}2.4}
\defV{dhi-q-ream50-usage}{\ensuremath{-}1.0}
\defV{ret-q-ream50-usage}{98.3}
\defV{val-n-reap25-usage}{96.5}
\defV{d-n-reap25-usage}{+0.5}
\defV{dlo-n-reap25-usage}{0.0}
\defV{dhi-n-reap25-usage}{+1.1}
\defV{ret-n-reap25-usage}{100.6}
\defV{val-n-reap50-usage}{94.7}
\defV{d-n-reap50-usage}{\ensuremath{-}1.3}
\defV{dlo-n-reap50-usage}{\ensuremath{-}2.1}
\defV{dhi-n-reap50-usage}{\ensuremath{-}0.5}
\defV{ret-n-reap50-usage}{98.6}
\defV{val-n-ream25-usage}{95.5}
\defV{d-n-ream25-usage}{\ensuremath{-}0.5}
\defV{dlo-n-ream25-usage}{\ensuremath{-}1.2}
\defV{dhi-n-ream25-usage}{+0.2}
\defV{ret-n-ream25-usage}{99.5}
\defV{val-n-ream50-usage}{91.5}
\defV{d-n-ream50-usage}{\ensuremath{-}4.4}
\defV{dlo-n-ream50-usage}{\ensuremath{-}5.4}
\defV{dhi-n-ream50-usage}{\ensuremath{-}3.4}
\defV{ret-n-ream50-usage}{95.4}
\defV{val-q-reap25-brand}{68.7}
\defV{d-q-reap25-brand}{+3.9}
\defV{dlo-q-reap25-brand}{+2.7}
\defV{dhi-q-reap25-brand}{+5.1}
\defV{ret-q-reap25-brand}{106.0}
\defV{val-q-reap50-brand}{69.0}
\defV{d-q-reap50-brand}{+4.2}
\defV{dlo-q-reap50-brand}{+2.6}
\defV{dhi-q-reap50-brand}{+5.8}
\defV{ret-q-reap50-brand}{106.5}
\defV{val-q-ream25-brand}{64.5}
\defV{d-q-ream25-brand}{\ensuremath{-}0.3}
\defV{dlo-q-ream25-brand}{\ensuremath{-}1.6}
\defV{dhi-q-ream25-brand}{+1.0}
\defV{ret-q-ream25-brand}{99.5}
\defV{val-q-ream50-brand}{70.7}
\defV{d-q-ream50-brand}{+5.9}
\defV{dlo-q-ream50-brand}{+4.1}
\defV{dhi-q-ream50-brand}{+7.6}
\defV{ret-q-ream50-brand}{109.0}
\defV{val-n-reap25-brand}{45.8}
\defV{d-n-reap25-brand}{\ensuremath{-}4.0}
\defV{dlo-n-reap25-brand}{\ensuremath{-}5.5}
\defV{dhi-n-reap25-brand}{\ensuremath{-}2.5}
\defV{ret-n-reap25-brand}{92.0}
\defV{val-n-reap50-brand}{28.6}
\defV{d-n-reap50-brand}{\ensuremath{-}21.1}
\defV{dlo-n-reap50-brand}{\ensuremath{-}23.1}
\defV{dhi-n-reap50-brand}{\ensuremath{-}19.1}
\defV{ret-n-reap50-brand}{57.6}
\defV{val-n-ream25-brand}{39.8}
\defV{d-n-ream25-brand}{\ensuremath{-}10.0}
\defV{dlo-n-ream25-brand}{\ensuremath{-}11.8}
\defV{dhi-n-ream25-brand}{\ensuremath{-}8.0}
\defV{ret-n-ream25-brand}{80.0}
\defV{val-n-ream50-brand}{34.1}
\defV{d-n-ream50-brand}{\ensuremath{-}15.6}
\defV{dlo-n-ream50-brand}{\ensuremath{-}17.7}
\defV{dhi-n-ream50-brand}{\ensuremath{-}13.6}
\defV{ret-n-ream50-brand}{68.5}
\defV{val-q-reap25-categ}{80.7}
\defV{d-q-reap25-categ}{\ensuremath{-}1.0}
\defV{dlo-q-reap25-categ}{\ensuremath{-}2.2}
\defV{dhi-q-reap25-categ}{+0.4}
\defV{ret-q-reap25-categ}{98.8}
\defV{val-q-reap50-categ}{75.0}
\defV{d-q-reap50-categ}{\ensuremath{-}6.7}
\defV{dlo-q-reap50-categ}{\ensuremath{-}8.6}
\defV{dhi-q-reap50-categ}{\ensuremath{-}4.9}
\defV{ret-q-reap50-categ}{91.8}
\defV{val-q-ream25-categ}{76.7}
\defV{d-q-ream25-categ}{\ensuremath{-}5.0}
\defV{dlo-q-ream25-categ}{\ensuremath{-}6.5}
\defV{dhi-q-ream25-categ}{\ensuremath{-}3.3}
\defV{ret-q-ream25-categ}{93.9}
\defV{val-q-ream50-categ}{71.2}
\defV{d-q-ream50-categ}{\ensuremath{-}10.4}
\defV{dlo-q-ream50-categ}{\ensuremath{-}12.5}
\defV{dhi-q-ream50-categ}{\ensuremath{-}8.4}
\defV{ret-q-ream50-categ}{87.3}
\defV{val-n-reap25-categ}{65.2}
\defV{d-n-reap25-categ}{\ensuremath{-}3.0}
\defV{dlo-n-reap25-categ}{\ensuremath{-}4.7}
\defV{dhi-n-reap25-categ}{\ensuremath{-}1.4}
\defV{ret-n-reap25-categ}{95.6}
\defV{val-n-reap50-categ}{56.9}
\defV{d-n-reap50-categ}{\ensuremath{-}11.3}
\defV{dlo-n-reap50-categ}{\ensuremath{-}13.4}
\defV{dhi-n-reap50-categ}{\ensuremath{-}9.2}
\defV{ret-n-reap50-categ}{83.4}
\defV{val-n-ream25-categ}{59.9}
\defV{d-n-ream25-categ}{\ensuremath{-}8.3}
\defV{dlo-n-ream25-categ}{\ensuremath{-}10.3}
\defV{dhi-n-ream25-categ}{\ensuremath{-}6.2}
\defV{ret-n-ream25-categ}{87.8}
\defV{val-n-ream50-categ}{49.9}
\defV{d-n-ream50-categ}{\ensuremath{-}18.3}
\defV{dlo-n-ream50-categ}{\ensuremath{-}20.5}
\defV{dhi-n-ream50-categ}{\ensuremath{-}16.0}
\defV{ret-n-ream50-categ}{73.1}
\defV{val-q-reap25-domains}{80.5}
\defV{d-q-reap25-domains}{\ensuremath{-}13.0}
\defV{dlo-q-reap25-domains}{\ensuremath{-}14.7}
\defV{dhi-q-reap25-domains}{\ensuremath{-}11.5}
\defV{ret-q-reap25-domains}{86.1}
\defV{val-q-reap50-domains}{71.3}
\defV{d-q-reap50-domains}{\ensuremath{-}22.2}
\defV{dlo-q-reap50-domains}{\ensuremath{-}24.3}
\defV{dhi-q-reap50-domains}{\ensuremath{-}20.2}
\defV{ret-q-reap50-domains}{76.2}
\defV{val-q-ream25-domains}{62.4}
\defV{d-q-ream25-domains}{\ensuremath{-}31.1}
\defV{dlo-q-ream25-domains}{\ensuremath{-}33.3}
\defV{dhi-q-ream25-domains}{\ensuremath{-}29.0}
\defV{ret-q-ream25-domains}{66.7}
\defV{val-q-ream50-domains}{63.5}
\defV{d-q-ream50-domains}{\ensuremath{-}30.1}
\defV{dlo-q-ream50-domains}{\ensuremath{-}32.3}
\defV{dhi-q-ream50-domains}{\ensuremath{-}27.9}
\defV{ret-q-ream50-domains}{67.8}
\defV{val-n-reap25-domains}{52.9}
\defV{d-n-reap25-domains}{\ensuremath{-}5.9}
\defV{dlo-n-reap25-domains}{\ensuremath{-}8.1}
\defV{dhi-n-reap25-domains}{\ensuremath{-}3.5}
\defV{ret-n-reap25-domains}{90.0}
\defV{val-n-reap50-domains}{36.8}
\defV{d-n-reap50-domains}{\ensuremath{-}22.0}
\defV{dlo-n-reap50-domains}{\ensuremath{-}24.4}
\defV{dhi-n-reap50-domains}{\ensuremath{-}19.4}
\defV{ret-n-reap50-domains}{62.6}
\defV{val-n-ream25-domains}{40.0}
\defV{d-n-ream25-domains}{\ensuremath{-}18.8}
\defV{dlo-n-ream25-domains}{\ensuremath{-}21.3}
\defV{dhi-n-ream25-domains}{\ensuremath{-}16.2}
\defV{ret-n-ream25-domains}{68.0}
\defV{val-n-ream50-domains}{31.7}
\defV{d-n-ream50-domains}{\ensuremath{-}27.1}
\defV{dlo-n-ream50-domains}{\ensuremath{-}29.6}
\defV{dhi-n-ream50-domains}{\ensuremath{-}24.4}
\defV{ret-n-ream50-domains}{54.0}
\defV{val-q-reap25-reach}{71.1}
\defV{d-q-reap25-reach}{+0.9}
\defV{dlo-q-reap25-reach}{\ensuremath{-}0.5}
\defV{dhi-q-reap25-reach}{+2.3}
\defV{ret-q-reap25-reach}{101.2}
\defV{val-q-reap50-reach}{45.3}
\defV{d-q-reap50-reach}{\ensuremath{-}25.0}
\defV{dlo-q-reap50-reach}{\ensuremath{-}26.8}
\defV{dhi-q-reap50-reach}{\ensuremath{-}23.1}
\defV{ret-q-reap50-reach}{64.4}
\defV{val-q-ream25-reach}{40.9}
\defV{d-q-ream25-reach}{\ensuremath{-}29.3}
\defV{dlo-q-ream25-reach}{\ensuremath{-}31.2}
\defV{dhi-q-ream25-reach}{\ensuremath{-}27.4}
\defV{ret-q-ream25-reach}{58.3}
\defV{val-q-ream50-reach}{25.1}
\defV{d-q-ream50-reach}{\ensuremath{-}45.1}
\defV{dlo-q-ream50-reach}{\ensuremath{-}46.9}
\defV{dhi-q-ream50-reach}{\ensuremath{-}43.3}
\defV{ret-q-ream50-reach}{35.8}
\defV{val-n-reap25-reach}{13.0}
\defV{d-n-reap25-reach}{+0.2}
\defV{dlo-n-reap25-reach}{\ensuremath{-}1.4}
\defV{dhi-n-reap25-reach}{+1.8}
\defV{ret-n-reap25-reach}{101.8}
\defV{val-n-reap50-reach}{2.9}
\defV{d-n-reap50-reach}{\ensuremath{-}9.8}
\defV{dlo-n-reap50-reach}{\ensuremath{-}11.2}
\defV{dhi-n-reap50-reach}{\ensuremath{-}8.5}
\defV{ret-n-reap50-reach}{23.0}
\defV{val-n-ream25-reach}{4.5}
\defV{d-n-ream25-reach}{\ensuremath{-}8.3}
\defV{dlo-n-ream25-reach}{\ensuremath{-}9.7}
\defV{dhi-n-ream25-reach}{\ensuremath{-}6.9}
\defV{ret-n-ream25-reach}{35.2}
\defV{val-n-ream50-reach}{0.4}
\defV{d-n-ream50-reach}{\ensuremath{-}12.4}
\defV{dlo-n-ream50-reach}{\ensuremath{-}13.7}
\defV{dhi-n-ream50-reach}{\ensuremath{-}11.2}
\defV{ret-n-ream50-reach}{3.0}
\defV{val-n-base-brand}{49.8}
\defV{val-n-base-categ}{68.2}
\defV{val-n-base-usage}{96.0}
\defV{val-n-base-domains}{58.8}
\defV{val-n-base-reach}{12.8}
\defV{val-q-base-brand}{64.8}
\defV{val-q-base-categ}{81.7}
\defV{val-q-base-usage}{98.8}
\defV{val-q-base-domains}{93.5}
\defV{val-q-base-reach}{70.2}
\defV{n-comp}{40}
\defV{n-sig}{32}
\defV{n-sig-worse}{29}
\defV{n-sig-better}{3}
\defV{n-ns}{8}
\defV{ret-q-reap25-disc}{101.5}
\defV{retlo-q-reap25-disc}{100.7}
\defV{rethi-q-reap25-disc}{102.4}
\defV{ret-q-reap25-gen}{93.6}
\defV{retlo-q-reap25-gen}{92.4}
\defV{rethi-q-reap25-gen}{95.0}
\defV{ret-q-reap50-disc}{99.1}
\defV{retlo-q-reap50-disc}{97.9}
\defV{rethi-q-reap50-disc}{100.3}
\defV{ret-q-reap50-gen}{70.3}
\defV{retlo-q-reap50-gen}{68.7}
\defV{rethi-q-reap50-gen}{72.0}
\defV{ret-q-ream25-disc}{97.7}
\defV{retlo-q-ream25-disc}{96.7}
\defV{rethi-q-ream25-disc}{98.6}
\defV{ret-q-ream25-gen}{62.5}
\defV{retlo-q-ream25-gen}{60.8}
\defV{rethi-q-ream25-gen}{64.2}
\defV{ret-q-ream50-disc}{98.2}
\defV{retlo-q-ream50-disc}{97.0}
\defV{rethi-q-ream50-disc}{99.5}
\defV{ret-q-ream50-gen}{51.8}
\defV{retlo-q-ream50-gen}{50.3}
\defV{rethi-q-ream50-gen}{53.4}
\defV{ret-n-reap25-disc}{96.0}
\defV{retlo-n-reap25-disc}{94.8}
\defV{rethi-n-reap25-disc}{97.3}
\defV{ret-n-reap25-gen}{95.9}
\defV{retlo-n-reap25-gen}{89.6}
\defV{rethi-n-reap25-gen}{103.0}
\defV{ret-n-reap50-disc}{79.9}
\defV{retlo-n-reap50-disc}{78.4}
\defV{rethi-n-reap50-disc}{81.4}
\defV{ret-n-reap50-gen}{42.8}
\defV{retlo-n-reap50-gen}{39.7}
\defV{rethi-n-reap50-gen}{46.1}
\defV{ret-n-ream25-disc}{89.1}
\defV{retlo-n-ream25-disc}{87.6}
\defV{rethi-n-ream25-disc}{90.7}
\defV{ret-n-ream25-gen}{51.6}
\defV{retlo-n-ream25-gen}{48.0}
\defV{rethi-n-ream25-gen}{55.7}
\defV{ret-n-ream50-disc}{79.0}
\defV{retlo-n-ream50-disc}{77.4}
\defV{rethi-n-ream50-disc}{80.6}
\defV{ret-n-ream50-gen}{28.5}
\defV{retlo-n-ream50-gen}{26.6}
\defV{rethi-n-ream50-gen}{30.5}
\defV{h-q25-usage}{+0.2}
\defV{hlo-q25-usage}{\ensuremath{-}0.2}
\defV{hhi-q25-usage}{+0.8}
\defV{hp-q25-usage}{p = 0.42}
\defV{h-q50-usage}{+0.7}
\defV{hlo-q50-usage}{0.0}
\defV{hhi-q50-usage}{+1.4}
\defV{hp-q50-usage}{p = 0.11}
\defV{h-n25-usage}{+1.0}
\defV{hlo-n25-usage}{+0.3}
\defV{hhi-n25-usage}{+1.7}
\defV{hp-n25-usage}{p = 0.026}
\defV{h-n50-usage}{+3.1}
\defV{hlo-n50-usage}{+2.2}
\defV{hhi-n50-usage}{+4.1}
\defV{hp-n50-usage}{p < 0.001}
\defV{h-q25-brand}{+4.2}
\defV{hlo-q25-brand}{+3.0}
\defV{hhi-q25-brand}{+5.5}
\defV{hp-q25-brand}{p < 0.001}
\defV{h-q50-brand}{\ensuremath{-}1.7}
\defV{hlo-q50-brand}{\ensuremath{-}2.9}
\defV{hhi-q50-brand}{\ensuremath{-}0.4}
\defV{hp-q50-brand}{p = 0.044}
\defV{h-n25-brand}{+6.0}
\defV{hlo-n25-brand}{+4.3}
\defV{hhi-n25-brand}{+7.6}
\defV{hp-n25-brand}{p < 0.001}
\defV{h-n50-brand}{\ensuremath{-}5.5}
\defV{hlo-n50-brand}{\ensuremath{-}7.2}
\defV{hhi-n50-brand}{\ensuremath{-}3.6}
\defV{hp-n50-brand}{p < 0.001}
\defV{h-q25-categ}{+4.0}
\defV{hlo-q25-categ}{+2.5}
\defV{hhi-q25-categ}{+5.5}
\defV{hp-q25-categ}{p < 0.001}
\defV{h-q50-categ}{+3.7}
\defV{hlo-q50-categ}{+1.9}
\defV{hhi-q50-categ}{+5.5}
\defV{hp-q50-categ}{p < 0.001}
\defV{h-n25-categ}{+5.3}
\defV{hlo-n25-categ}{+3.4}
\defV{hhi-n25-categ}{+7.2}
\defV{hp-n25-categ}{p < 0.001}
\defV{h-n50-categ}{+7.0}
\defV{hlo-n50-categ}{+5.0}
\defV{hhi-n50-categ}{+9.0}
\defV{hp-n50-categ}{p < 0.001}
\defV{h-q25-domains}{+18.1}
\defV{hlo-q25-domains}{+15.9}
\defV{hhi-q25-domains}{+20.3}
\defV{hp-q25-domains}{p < 0.001}
\defV{h-q50-domains}{+7.8}
\defV{hlo-q50-domains}{+5.7}
\defV{hhi-q50-domains}{+9.9}
\defV{hp-q50-domains}{p < 0.001}
\defV{h-n25-domains}{+12.9}
\defV{hlo-n25-domains}{+10.3}
\defV{hhi-n25-domains}{+15.6}
\defV{hp-n25-domains}{p < 0.001}
\defV{h-n50-domains}{+5.1}
\defV{hlo-n50-domains}{+2.7}
\defV{hhi-n50-domains}{+7.3}
\defV{hp-n50-domains}{p < 0.001}
\defV{h-q25-reach}{+30.2}
\defV{hlo-q25-reach}{+28.4}
\defV{hhi-q25-reach}{+32.0}
\defV{hp-q25-reach}{p < 0.001}
\defV{h-q50-reach}{+20.1}
\defV{hlo-q50-reach}{+18.0}
\defV{hhi-q50-reach}{+22.2}
\defV{hp-q50-reach}{p < 0.001}
\defV{h-n25-reach}{+8.5}
\defV{hlo-n25-reach}{+7.2}
\defV{hhi-n25-reach}{+9.9}
\defV{hp-n25-reach}{p < 0.001}
\defV{h-n50-reach}{+2.6}
\defV{hlo-n50-reach}{+2.0}
\defV{hhi-n50-reach}{+3.2}
\defV{hp-n50-reach}{p < 0.001}
\defV{h-total}{20}
\defV{h-reap}{16}
\defV{h-ream}{2}
\defV{h-ns}{2}
\defV{reach-rel-q-base}{70.2}
\defV{reach-prod-q-base}{26.1}
\defV{reach-passall-q-base}{88.7}
\defV{reach-failgate-q-base}{11.3}
\defV{reach-meanpass-q-base}{79.2}
\defV{reach-capped-q-base}{6}
\defV{reach-capped-pct-q-base}{0.3}
\defV{reach-jsoninvalid-q-base}{1.1}
\defV{reach-jsoninvalid-n-q-base}{22}
\defV{reach-medtok-q-base}{2{,}724}
\defV{gate-outputparseable-q-base}{1.1}
\defV{gate-outputparseable-n-q-base}{22}
\defV{gateprod-outputparseable-q-base}{1.1}
\defV{gateprod-outputparseable-n-q-base}{22}
\defV{gate-htmlpresent-q-base}{0.0}
\defV{gate-htmlpresent-n-q-base}{0}
\defV{gateprod-htmlpresent-q-base}{0.0}
\defV{gateprod-htmlpresent-n-q-base}{0}
\defV{gate-htmlcomplete-q-base}{0.1}
\defV{gate-htmlcomplete-n-q-base}{2}
\defV{gateprod-htmlcomplete-q-base}{61.5}
\defV{gateprod-htmlcomplete-n-q-base}{1{,}230}
\defV{gate-unsubscribelink-q-base}{0.1}
\defV{gate-unsubscribelink-n-q-base}{3}
\defV{gateprod-unsubscribelink-q-base}{0.2}
\defV{gateprod-unsubscribelink-n-q-base}{4}
\defV{gate-mobilelayoutintact-q-base}{10.0}
\defV{gate-mobilelayoutintact-n-q-base}{200}
\defV{gateprod-mobilelayoutintact-q-base}{10.0}
\defV{gateprod-mobilelayoutintact-n-q-base}{200}
\defV{reach-rel-q-reap25}{71.1}
\defV{reach-prod-q-reap25}{51.3}
\defV{reach-passall-q-reap25}{89.9}
\defV{reach-failgate-q-reap25}{10.1}
\defV{reach-meanpass-q-reap25}{79.1}
\defV{reach-capped-q-reap25}{14}
\defV{reach-capped-pct-q-reap25}{0.7}
\defV{reach-jsoninvalid-q-reap25}{1.0}
\defV{reach-jsoninvalid-n-q-reap25}{21}
\defV{reach-medtok-q-reap25}{2{,}670}
\defV{gate-outputparseable-q-reap25}{1.1}
\defV{gate-outputparseable-n-q-reap25}{21}
\defV{gateprod-outputparseable-q-reap25}{1.1}
\defV{gateprod-outputparseable-n-q-reap25}{21}
\defV{gate-htmlpresent-q-reap25}{0.0}
\defV{gate-htmlpresent-n-q-reap25}{0}
\defV{gateprod-htmlpresent-q-reap25}{0.0}
\defV{gateprod-htmlpresent-n-q-reap25}{0}
\defV{gate-htmlcomplete-q-reap25}{0.1}
\defV{gate-htmlcomplete-n-q-reap25}{3}
\defV{gateprod-htmlcomplete-q-reap25}{28.0}
\defV{gateprod-htmlcomplete-n-q-reap25}{560}
\defV{gate-unsubscribelink-q-reap25}{0.1}
\defV{gate-unsubscribelink-n-q-reap25}{2}
\defV{gateprod-unsubscribelink-q-reap25}{0.1}
\defV{gateprod-unsubscribelink-n-q-reap25}{3}
\defV{gate-mobilelayoutintact-q-reap25}{8.8}
\defV{gate-mobilelayoutintact-n-q-reap25}{177}
\defV{gateprod-mobilelayoutintact-q-reap25}{8.8}
\defV{gateprod-mobilelayoutintact-n-q-reap25}{177}
\defV{reach-loss-q-reap25}{\ensuremath{-}0.9}
\defV{reach-lossgate-q-reap25}{\ensuremath{-}0.9}
\defV{reach-lossqual-q-reap25}{0.1}
\defV{reach-rel-q-reap50}{45.3}
\defV{reach-prod-q-reap50}{37.7}
\defV{reach-passall-q-reap50}{59.2}
\defV{reach-failgate-q-reap50}{40.8}
\defV{reach-meanpass-q-reap50}{76.5}
\defV{reach-capped-q-reap50}{102}
\defV{reach-capped-pct-q-reap50}{5.1}
\defV{reach-jsoninvalid-q-reap50}{5.4}
\defV{reach-jsoninvalid-n-q-reap50}{108}
\defV{reach-medtok-q-reap50}{2{,}646}
\defV{gate-outputparseable-q-reap50}{5.4}
\defV{gate-outputparseable-n-q-reap50}{108}
\defV{gateprod-outputparseable-q-reap50}{5.4}
\defV{gateprod-outputparseable-n-q-reap50}{108}
\defV{gate-htmlpresent-q-reap50}{0.1}
\defV{gate-htmlpresent-n-q-reap50}{1}
\defV{gateprod-htmlpresent-q-reap50}{0.1}
\defV{gateprod-htmlpresent-n-q-reap50}{1}
\defV{gate-htmlcomplete-q-reap50}{0.4}
\defV{gate-htmlcomplete-n-q-reap50}{8}
\defV{gateprod-htmlcomplete-q-reap50}{14.3}
\defV{gateprod-htmlcomplete-n-q-reap50}{287}
\defV{gate-unsubscribelink-q-reap50}{0.1}
\defV{gate-unsubscribelink-n-q-reap50}{3}
\defV{gateprod-unsubscribelink-q-reap50}{0.2}
\defV{gateprod-unsubscribelink-n-q-reap50}{4}
\defV{gate-mobilelayoutintact-q-reap50}{35.1}
\defV{gate-mobilelayoutintact-n-q-reap50}{702}
\defV{gateprod-mobilelayoutintact-q-reap50}{35.1}
\defV{gateprod-mobilelayoutintact-n-q-reap50}{702}
\defV{reach-loss-q-reap50}{25.0}
\defV{reach-lossgate-q-reap50}{23.0}
\defV{reach-lossqual-q-reap50}{2.0}
\defV{reach-gateshare-q-reap50}{92.1}
\defV{reach-rel-q-ream25}{40.9}
\defV{reach-prod-q-ream25}{0.4}
\defV{reach-passall-q-ream25}{54.5}
\defV{reach-failgate-q-ream25}{45.5}
\defV{reach-meanpass-q-ream25}{75.0}
\defV{reach-capped-q-ream25}{61}
\defV{reach-capped-pct-q-ream25}{3.0}
\defV{reach-jsoninvalid-q-ream25}{5.1}
\defV{reach-jsoninvalid-n-q-ream25}{102}
\defV{reach-medtok-q-ream25}{2{,}228}
\defV{gate-outputparseable-q-ream25}{5.1}
\defV{gate-outputparseable-n-q-ream25}{102}
\defV{gateprod-outputparseable-q-ream25}{5.1}
\defV{gateprod-outputparseable-n-q-ream25}{102}
\defV{gate-htmlpresent-q-ream25}{0.0}
\defV{gate-htmlpresent-n-q-ream25}{0}
\defV{gateprod-htmlpresent-q-ream25}{0.0}
\defV{gateprod-htmlpresent-n-q-ream25}{0}
\defV{gate-htmlcomplete-q-ream25}{0.1}
\defV{gate-htmlcomplete-n-q-ream25}{2}
\defV{gateprod-htmlcomplete-q-ream25}{94.3}
\defV{gateprod-htmlcomplete-n-q-ream25}{1{,}886}
\defV{gate-unsubscribelink-q-ream25}{0.4}
\defV{gate-unsubscribelink-n-q-ream25}{9}
\defV{gateprod-unsubscribelink-q-ream25}{0.8}
\defV{gateprod-unsubscribelink-n-q-ream25}{15}
\defV{gate-mobilelayoutintact-q-ream25}{40.1}
\defV{gate-mobilelayoutintact-n-q-ream25}{802}
\defV{gateprod-mobilelayoutintact-q-ream25}{40.1}
\defV{gateprod-mobilelayoutintact-n-q-ream25}{802}
\defV{reach-loss-q-ream25}{29.3}
\defV{reach-lossgate-q-ream25}{26.3}
\defV{reach-lossqual-q-ream25}{3.0}
\defV{reach-gateshare-q-ream25}{89.8}
\defV{reach-rel-q-ream50}{25.1}
\defV{reach-prod-q-ream50}{1.9}
\defV{reach-passall-q-ream50}{34.4}
\defV{reach-failgate-q-ream50}{65.6}
\defV{reach-meanpass-q-ream50}{73.1}
\defV{reach-capped-q-ream50}{215}
\defV{reach-capped-pct-q-ream50}{10.8}
\defV{reach-jsoninvalid-q-ream50}{12.2}
\defV{reach-jsoninvalid-n-q-ream50}{245}
\defV{reach-medtok-q-ream50}{2{,}398}
\defV{gate-outputparseable-q-ream50}{12.2}
\defV{gate-outputparseable-n-q-ream50}{245}
\defV{gateprod-outputparseable-q-ream50}{12.2}
\defV{gateprod-outputparseable-n-q-ream50}{245}
\defV{gate-htmlpresent-q-ream50}{0.0}
\defV{gate-htmlpresent-n-q-ream50}{0}
\defV{gateprod-htmlpresent-q-ream50}{0.0}
\defV{gateprod-htmlpresent-n-q-ream50}{0}
\defV{gate-htmlcomplete-q-ream50}{2.6}
\defV{gate-htmlcomplete-n-q-ream50}{53}
\defV{gateprod-htmlcomplete-q-ream50}{82.5}
\defV{gateprod-htmlcomplete-n-q-ream50}{1{,}650}
\defV{gate-unsubscribelink-q-ream50}{0.4}
\defV{gate-unsubscribelink-n-q-ream50}{8}
\defV{gateprod-unsubscribelink-q-ream50}{0.5}
\defV{gateprod-unsubscribelink-n-q-ream50}{11}
\defV{gate-mobilelayoutintact-q-ream50}{51.7}
\defV{gate-mobilelayoutintact-n-q-ream50}{1{,}035}
\defV{gateprod-mobilelayoutintact-q-ream50}{51.7}
\defV{gateprod-mobilelayoutintact-n-q-ream50}{1{,}035}
\defV{reach-loss-q-ream50}{45.1}
\defV{reach-lossgate-q-ream50}{41.3}
\defV{reach-lossqual-q-ream50}{3.8}
\defV{reach-gateshare-q-ream50}{91.7}
\defV{reach-rel-n-base}{12.8}
\defV{reach-prod-n-base}{11.9}
\defV{reach-passall-n-base}{17.8}
\defV{reach-failgate-n-base}{82.2}
\defV{reach-meanpass-n-base}{72.0}
\defV{reach-capped-n-base}{38}
\defV{reach-capped-pct-n-base}{1.9}
\defV{reach-jsoninvalid-n-base}{27.2}
\defV{reach-jsoninvalid-n-n-base}{544}
\defV{reach-medtok-n-base}{1{,}520}
\defV{gate-outputparseable-n-base}{27.2}
\defV{gate-outputparseable-n-n-base}{544}
\defV{gateprod-outputparseable-n-base}{27.2}
\defV{gateprod-outputparseable-n-n-base}{544}
\defV{gate-htmlpresent-n-base}{0.1}
\defV{gate-htmlpresent-n-n-base}{3}
\defV{gateprod-htmlpresent-n-base}{0.1}
\defV{gateprod-htmlpresent-n-n-base}{3}
\defV{gate-htmlcomplete-n-base}{34.9}
\defV{gate-htmlcomplete-n-n-base}{698}
\defV{gateprod-htmlcomplete-n-base}{35.3}
\defV{gateprod-htmlcomplete-n-n-base}{706}
\defV{gate-unsubscribelink-n-base}{35.0}
\defV{gate-unsubscribelink-n-n-base}{701}
\defV{gateprod-unsubscribelink-n-base}{36.9}
\defV{gateprod-unsubscribelink-n-n-base}{737}
\defV{gate-mobilelayoutintact-n-base}{13.9}
\defV{gate-mobilelayoutintact-n-n-base}{278}
\defV{gateprod-mobilelayoutintact-n-base}{13.9}
\defV{gateprod-mobilelayoutintact-n-n-base}{278}
\defV{reach-rel-n-reap25}{13.0}
\defV{reach-prod-n-reap25}{12.1}
\defV{reach-passall-n-reap25}{17.9}
\defV{reach-failgate-n-reap25}{82.0}
\defV{reach-meanpass-n-reap25}{72.5}
\defV{reach-capped-n-reap25}{32}
\defV{reach-capped-pct-n-reap25}{1.6}
\defV{reach-jsoninvalid-n-reap25}{22.8}
\defV{reach-jsoninvalid-n-n-reap25}{455}
\defV{reach-medtok-n-reap25}{1{,}430}
\defV{gate-outputparseable-n-reap25}{22.8}
\defV{gate-outputparseable-n-n-reap25}{455}
\defV{gateprod-outputparseable-n-reap25}{22.8}
\defV{gateprod-outputparseable-n-n-reap25}{455}
\defV{gate-htmlpresent-n-reap25}{0.2}
\defV{gate-htmlpresent-n-n-reap25}{4}
\defV{gateprod-htmlpresent-n-reap25}{0.2}
\defV{gateprod-htmlpresent-n-n-reap25}{4}
\defV{gate-htmlcomplete-n-reap25}{32.3}
\defV{gate-htmlcomplete-n-n-reap25}{646}
\defV{gateprod-htmlcomplete-n-reap25}{32.8}
\defV{gateprod-htmlcomplete-n-n-reap25}{655}
\defV{gate-unsubscribelink-n-reap25}{40.3}
\defV{gate-unsubscribelink-n-n-reap25}{806}
\defV{gateprod-unsubscribelink-n-reap25}{42.5}
\defV{gateprod-unsubscribelink-n-n-reap25}{850}
\defV{gate-mobilelayoutintact-n-reap25}{15.2}
\defV{gate-mobilelayoutintact-n-n-reap25}{304}
\defV{gateprod-mobilelayoutintact-n-reap25}{15.2}
\defV{gateprod-mobilelayoutintact-n-n-reap25}{304}
\defV{reach-loss-n-reap25}{\ensuremath{-}0.2}
\defV{reach-lossgate-n-reap25}{\ensuremath{-}0.1}
\defV{reach-lossqual-n-reap25}{\ensuremath{-}0.1}
\defV{reach-rel-n-reap50}{2.9}
\defV{reach-prod-n-reap50}{2.6}
\defV{reach-passall-n-reap50}{4.3}
\defV{reach-failgate-n-reap50}{95.7}
\defV{reach-meanpass-n-reap50}{67.7}
\defV{reach-capped-n-reap50}{226}
\defV{reach-capped-pct-n-reap50}{11.3}
\defV{reach-jsoninvalid-n-reap50}{47.8}
\defV{reach-jsoninvalid-n-n-reap50}{956}
\defV{reach-medtok-n-reap50}{1{,}583}
\defV{gate-outputparseable-n-reap50}{47.8}
\defV{gate-outputparseable-n-n-reap50}{956}
\defV{gateprod-outputparseable-n-reap50}{47.8}
\defV{gateprod-outputparseable-n-n-reap50}{956}
\defV{gate-htmlpresent-n-reap50}{0.1}
\defV{gate-htmlpresent-n-n-reap50}{3}
\defV{gateprod-htmlpresent-n-reap50}{0.1}
\defV{gateprod-htmlpresent-n-n-reap50}{3}
\defV{gate-htmlcomplete-n-reap50}{27.9}
\defV{gate-htmlcomplete-n-n-reap50}{558}
\defV{gateprod-htmlcomplete-n-reap50}{28.1}
\defV{gateprod-htmlcomplete-n-n-reap50}{561}
\defV{gate-unsubscribelink-n-reap50}{36.4}
\defV{gate-unsubscribelink-n-n-reap50}{729}
\defV{gateprod-unsubscribelink-n-reap50}{37.5}
\defV{gateprod-unsubscribelink-n-n-reap50}{751}
\defV{gate-mobilelayoutintact-n-reap50}{16.2}
\defV{gate-mobilelayoutintact-n-n-reap50}{323}
\defV{gateprod-mobilelayoutintact-n-reap50}{16.2}
\defV{gateprod-mobilelayoutintact-n-n-reap50}{323}
\defV{reach-loss-n-reap50}{9.8}
\defV{reach-lossgate-n-reap50}{9.4}
\defV{reach-lossqual-n-reap50}{0.5}
\defV{reach-gateshare-n-reap50}{95.2}
\defV{reach-rel-n-ream25}{4.5}
\defV{reach-prod-n-ream25}{2.9}
\defV{reach-passall-n-ream25}{6.7}
\defV{reach-failgate-n-ream25}{93.3}
\defV{reach-meanpass-n-ream25}{67.7}
\defV{reach-capped-n-ream25}{166}
\defV{reach-capped-pct-n-ream25}{8.3}
\defV{reach-jsoninvalid-n-ream25}{34.0}
\defV{reach-jsoninvalid-n-n-ream25}{679}
\defV{reach-medtok-n-ream25}{1{,}176}
\defV{gate-outputparseable-n-ream25}{34.0}
\defV{gate-outputparseable-n-n-ream25}{679}
\defV{gateprod-outputparseable-n-ream25}{34.0}
\defV{gateprod-outputparseable-n-n-ream25}{679}
\defV{gate-htmlpresent-n-ream25}{0.4}
\defV{gate-htmlpresent-n-n-ream25}{7}
\defV{gateprod-htmlpresent-n-ream25}{0.4}
\defV{gateprod-htmlpresent-n-n-ream25}{7}
\defV{gate-htmlcomplete-n-ream25}{30.2}
\defV{gate-htmlcomplete-n-n-ream25}{604}
\defV{gateprod-htmlcomplete-n-ream25}{44.9}
\defV{gateprod-htmlcomplete-n-n-ream25}{898}
\defV{gate-unsubscribelink-n-ream25}{54.1}
\defV{gate-unsubscribelink-n-n-ream25}{1{,}083}
\defV{gateprod-unsubscribelink-n-ream25}{54.9}
\defV{gateprod-unsubscribelink-n-n-ream25}{1{,}098}
\defV{gate-mobilelayoutintact-n-ream25}{10.4}
\defV{gate-mobilelayoutintact-n-n-ream25}{209}
\defV{gateprod-mobilelayoutintact-n-ream25}{10.4}
\defV{gateprod-mobilelayoutintact-n-n-ream25}{209}
\defV{reach-loss-n-ream25}{8.3}
\defV{reach-lossgate-n-ream25}{7.8}
\defV{reach-lossqual-n-ream25}{0.5}
\defV{reach-gateshare-n-ream25}{93.7}
\defV{reach-rel-n-ream50}{0.4}
\defV{reach-prod-n-ream50}{0.1}
\defV{reach-passall-n-ream50}{0.6}
\defV{reach-failgate-n-ream50}{99.4}
\defV{reach-meanpass-n-ream50}{63.1}
\defV{reach-capped-n-ream50}{299}
\defV{reach-capped-pct-n-ream50}{14.9}
\defV{reach-jsoninvalid-n-ream50}{46.7}
\defV{reach-jsoninvalid-n-n-ream50}{933}
\defV{reach-medtok-n-ream50}{1{,}400}
\defV{gate-outputparseable-n-ream50}{46.7}
\defV{gate-outputparseable-n-n-ream50}{933}
\defV{gateprod-outputparseable-n-ream50}{46.7}
\defV{gateprod-outputparseable-n-n-ream50}{933}
\defV{gate-htmlpresent-n-ream50}{0.1}
\defV{gate-htmlpresent-n-n-ream50}{2}
\defV{gateprod-htmlpresent-n-ream50}{0.1}
\defV{gateprod-htmlpresent-n-n-ream50}{2}
\defV{gate-htmlcomplete-n-ream50}{46.4}
\defV{gate-htmlcomplete-n-n-ream50}{927}
\defV{gateprod-htmlcomplete-n-ream50}{49.5}
\defV{gateprod-htmlcomplete-n-n-ream50}{990}
\defV{gate-unsubscribelink-n-ream50}{43.4}
\defV{gate-unsubscribelink-n-n-ream50}{868}
\defV{gateprod-unsubscribelink-n-ream50}{44.9}
\defV{gateprod-unsubscribelink-n-n-ream50}{897}
\defV{gate-mobilelayoutintact-n-ream50}{15.2}
\defV{gate-mobilelayoutintact-n-n-ream50}{305}
\defV{gateprod-mobilelayoutintact-n-ream50}{15.2}
\defV{gateprod-mobilelayoutintact-n-n-ream50}{305}
\defV{reach-loss-n-ream50}{12.4}
\defV{reach-lossgate-n-ream50}{11.6}
\defV{reach-lossqual-n-ream50}{0.8}
\defV{reach-gateshare-n-ream50}{93.4}
\defV{reach-gateshare-min}{89.8}
\defV{reach-gateshare-max}{95.2}
\defV{reach-capped-max-pct}{14.9}
\defV{drift-q-base}{5.2}
\defV{passen-q-base}{98.2}
\defV{passnonen-q-base}{92.6}
\defV{passdrift-q-base}{34.5}
\defV{passnodrift-q-base}{95.8}
\defV{drift-q-reap25}{13.6}
\defV{passen-q-reap25}{99.4}
\defV{passnonen-q-reap25}{76.6}
\defV{passdrift-q-reap25}{14.7}
\defV{passnodrift-q-reap25}{86.3}
\defV{drift-q-reap50}{25.9}
\defV{passen-q-reap50}{96.8}
\defV{passnonen-q-reap50}{66.0}
\defV{passdrift-q-reap50}{7.0}
\defV{passnodrift-q-reap50}{86.7}
\defV{drift-q-ream25}{43.8}
\defV{passen-q-ream25}{97.7}
\defV{passnonen-q-ream25}{55.1}
\defV{passdrift-q-ream25}{4.7}
\defV{passnodrift-q-ream25}{94.4}
\defV{drift-q-ream50}{33.9}
\defV{passen-q-ream50}{95.6}
\defV{passnonen-q-ream50}{56.8}
\defV{passdrift-q-ream50}{2.8}
\defV{passnodrift-q-ream50}{84.5}
\defV{drift-n-base}{34.9}
\defV{passen-n-base}{92.7}
\defV{passnonen-n-base}{51.7}
\defV{passdrift-n-base}{18.7}
\defV{passnodrift-n-base}{69.4}
\defV{drift-n-reap25}{41.4}
\defV{passen-n-reap25}{93.3}
\defV{passnonen-n-reap25}{44.5}
\defV{passdrift-n-reap25}{19.9}
\defV{passnodrift-n-reap25}{61.9}
\defV{drift-n-reap50}{63.6}
\defV{passen-n-reap50}{96.2}
\defV{passnonen-n-reap50}{24.5}
\defV{passdrift-n-reap50}{12.7}
\defV{passnodrift-n-reap50}{45.1}
\defV{drift-n-ream25}{50.1}
\defV{passen-n-ream25}{86.3}
\defV{passnonen-n-ream25}{30.4}
\defV{passdrift-n-ream25}{15.8}
\defV{passnodrift-n-ream25}{45.0}
\defV{drift-n-ream50}{40.5}
\defV{passen-n-ream50}{78.9}
\defV{passnonen-n-ream50}{22.0}
\defV{passdrift-n-ream50}{6.8}
\defV{passnodrift-n-ream50}{32.3}
\defV{engopen-q-base}{3.4}
\defV{wronglang-q-base}{1.6}
\defV{domcapped-q-base}{0.0}
\defV{engopen-q-reap25}{11.6}
\defV{wronglang-q-reap25}{1.7}
\defV{domcapped-q-reap25}{0.0}
\defV{engopen-q-reap50}{17.1}
\defV{wronglang-q-reap50}{7.4}
\defV{domcapped-q-reap50}{0.1}
\defV{engopen-q-ream25}{41.7}
\defV{wronglang-q-ream25}{3.4}
\defV{domcapped-q-ream25}{0.0}
\defV{engopen-q-ream50}{26.3}
\defV{wronglang-q-ream50}{8.9}
\defV{domcapped-q-ream50}{0.2}
\defV{engopen-n-base}{32.3}
\defV{wronglang-n-base}{13.0}
\defV{domcapped-n-base}{0.0}
\defV{engopen-n-reap25}{40.0}
\defV{wronglang-n-reap25}{13.0}
\defV{domcapped-n-reap25}{0.0}
\defV{engopen-n-reap50}{60.5}
\defV{wronglang-n-reap50}{22.2}
\defV{domcapped-n-reap50}{0.8}
\defV{engopen-n-ream25}{41.6}
\defV{wronglang-n-ream25}{22.0}
\defV{domcapped-n-ream25}{0.2}
\defV{engopen-n-ream50}{32.0}
\defV{wronglang-n-ream50}{18.4}
\defV{domcapped-n-ream50}{1.0}
\defV{brand-detect-q-base}{87.2}
\defV{brand-lenient-q-base}{74.3}
\defV{brand-detect-q-reap25}{91.0}
\defV{brand-lenient-q-reap25}{77.8}
\defV{brand-detect-q-reap50}{91.4}
\defV{brand-lenient-q-reap50}{77.4}
\defV{brand-detect-q-ream25}{88.2}
\defV{brand-lenient-q-ream25}{74.1}
\defV{brand-detect-q-ream50}{95.2}
\defV{brand-lenient-q-ream50}{80.4}
\defV{brand-detect-n-base}{78.4}
\defV{brand-lenient-n-base}{55.7}
\defV{brand-detect-n-reap25}{72.0}
\defV{brand-lenient-n-reap25}{51.4}
\defV{brand-detect-n-reap50}{54.5}
\defV{brand-lenient-n-reap50}{32.0}
\defV{brand-detect-n-ream25}{61.6}
\defV{brand-lenient-n-ream25}{45.3}
\defV{brand-detect-n-ream50}{53.5}
\defV{brand-lenient-n-ream50}{38.2}
\defV{same-q-reap25-usage}{99.4}
\defV{same-q-reap25-brand}{83.0}
\defV{same-q-reap25-categ}{78.0}
\defV{lost-q-reap25-usage}{0.4}
\defV{gained-q-reap25-usage}{0.2}
\defV{lost-q-reap25-brand}{2.0}
\defV{gained-q-reap25-brand}{5.9}
\defV{lost-q-reap25-categ}{4.8}
\defV{gained-q-reap25-categ}{3.8}
\defV{lost-q-reap25-domains}{14.2}
\defV{gained-q-reap25-domains}{1.2}
\defV{same-q-reap50-usage}{98.4}
\defV{same-q-reap50-brand}{72.9}
\defV{same-q-reap50-categ}{59.8}
\defV{lost-q-reap50-usage}{1.2}
\defV{gained-q-reap50-usage}{0.3}
\defV{lost-q-reap50-brand}{5.1}
\defV{gained-q-reap50-brand}{9.4}
\defV{lost-q-reap50-categ}{12.4}
\defV{gained-q-reap50-categ}{5.7}
\defV{lost-q-reap50-domains}{24.1}
\defV{gained-q-reap50-domains}{1.8}
\defV{same-q-ream25-usage}{98.7}
\defV{same-q-ream25-brand}{78.3}
\defV{same-q-ream25-categ}{67.0}
\defV{lost-q-ream25-usage}{0.9}
\defV{gained-q-ream25-usage}{0.4}
\defV{lost-q-ream25-brand}{4.5}
\defV{gained-q-ream25-brand}{4.2}
\defV{lost-q-ream25-categ}{9.4}
\defV{gained-q-ream25-categ}{4.4}
\defV{lost-q-ream25-domains}{32.2}
\defV{gained-q-ream25-domains}{1.0}
\defV{same-q-ream50-usage}{97.5}
\defV{same-q-ream50-brand}{71.9}
\defV{same-q-ream50-categ}{53.8}
\defV{lost-q-ream50-usage}{2.1}
\defV{gained-q-ream50-usage}{0.4}
\defV{lost-q-ream50-brand}{5.1}
\defV{gained-q-ream50-brand}{11.0}
\defV{lost-q-ream50-categ}{16.2}
\defV{gained-q-ream50-categ}{5.8}
\defV{lost-q-ream50-domains}{32.4}
\defV{gained-q-ream50-domains}{2.3}
\defV{same-n-reap25-usage}{98.4}
\defV{same-n-reap25-brand}{58.9}
\defV{same-n-reap25-categ}{67.4}
\defV{lost-n-reap25-usage}{0.4}
\defV{gained-n-reap25-usage}{1.0}
\defV{lost-n-reap25-brand}{7.8}
\defV{gained-n-reap25-brand}{3.8}
\defV{lost-n-reap25-categ}{8.5}
\defV{gained-n-reap25-categ}{5.4}
\defV{lost-n-reap25-domains}{17.5}
\defV{gained-n-reap25-domains}{11.6}
\defV{same-n-reap50-usage}{96.4}
\defV{same-n-reap50-brand}{34.9}
\defV{same-n-reap50-categ}{49.6}
\defV{lost-n-reap50-usage}{2.4}
\defV{gained-n-reap50-usage}{1.0}
\defV{lost-n-reap50-brand}{23.0}
\defV{gained-n-reap50-brand}{2.0}
\defV{lost-n-reap50-categ}{17.5}
\defV{gained-n-reap50-categ}{6.2}
\defV{lost-n-reap50-domains}{29.8}
\defV{gained-n-reap50-domains}{7.8}
\defV{same-n-ream25-usage}{97.2}
\defV{same-n-ream25-brand}{46.1}
\defV{same-n-ream25-categ}{54.6}
\defV{lost-n-ream25-usage}{1.6}
\defV{gained-n-ream25-usage}{1.2}
\defV{lost-n-ream25-brand}{14.8}
\defV{gained-n-ream25-brand}{4.9}
\defV{lost-n-ream25-categ}{15.2}
\defV{gained-n-ream25-categ}{6.9}
\defV{lost-n-ream25-domains}{28.4}
\defV{gained-n-ream25-domains}{9.6}
\defV{same-n-ream50-usage}{94.4}
\defV{same-n-ream50-brand}{37.4}
\defV{same-n-ream50-categ}{43.3}
\defV{lost-n-ream50-usage}{5.0}
\defV{gained-n-ream50-usage}{0.5}
\defV{lost-n-ream50-brand}{20.2}
\defV{gained-n-ream50-brand}{4.6}
\defV{lost-n-ream50-categ}{24.2}
\defV{gained-n-ream50-categ}{5.9}
\defV{lost-n-ream50-domains}{34.7}
\defV{gained-n-ream50-domains}{7.6}
\defV{lang-cells}{240}
\defV{lang-sig}{154}
\defV{lang-sig-worse}{138}
\defV{lang-sig-better}{16}
\defV{langmean-brand-en}{\ensuremath{-}2.6}
\defV{langmean-brand-es}{\ensuremath{-}5.6}
\defV{langmean-brand-fr}{\ensuremath{-}3.7}
\defV{langmean-brand-id}{\ensuremath{-}5.7}
\defV{langmean-brand-other}{\ensuremath{-}4.0}
\defV{langmean-brand-pt}{\ensuremath{-}6.2}
\defV{langmean-brand-nonen}{\ensuremath{-}5.1}
\defV{langmean-categ-en}{\ensuremath{-}6.8}
\defV{langmean-categ-es}{\ensuremath{-}4.9}
\defV{langmean-categ-fr}{\ensuremath{-}7.6}
\defV{langmean-categ-id}{\ensuremath{-}13.7}
\defV{langmean-categ-other}{\ensuremath{-}9.5}
\defV{langmean-categ-pt}{\ensuremath{-}6.8}
\defV{langmean-categ-nonen}{\ensuremath{-}8.5}
\defV{langmean-usage-en}{\ensuremath{-}1.0}
\defV{langmean-usage-es}{\ensuremath{-}2.0}
\defV{langmean-usage-fr}{+0.3}
\defV{langmean-usage-id}{\ensuremath{-}2.0}
\defV{langmean-usage-other}{\ensuremath{-}1.1}
\defV{langmean-usage-pt}{\ensuremath{-}1.1}
\defV{langmean-usage-nonen}{\ensuremath{-}1.2}
\defV{langmean-domains-en}{\ensuremath{-}2.4}
\defV{langmean-domains-es}{\ensuremath{-}15.4}
\defV{langmean-domains-fr}{\ensuremath{-}31.4}
\defV{langmean-domains-id}{\ensuremath{-}18.6}
\defV{langmean-domains-other}{\ensuremath{-}34.2}
\defV{langmean-domains-pt}{\ensuremath{-}24.8}
\defV{langmean-domains-nonen}{\ensuremath{-}24.9}
\defV{langmean-reach-en}{\ensuremath{-}15.6}
\defV{langmean-reach-es}{\ensuremath{-}16.8}
\defV{langmean-reach-fr}{\ensuremath{-}18.7}
\defV{langmean-reach-id}{\ensuremath{-}9.7}
\defV{langmean-reach-other}{\ensuremath{-}15.3}
\defV{langmean-reach-pt}{\ensuremath{-}16.9}
\defV{langmean-reach-nonen}{\ensuremath{-}15.5}
\defV{lang-q-reap25-usage-en}{\ensuremath{-}0.1}
\defV{lang-q-reap25-usage-pt}{0.0}
\defV{lang-q-reap25-usage-es}{0.0}
\defV{lang-q-reap25-usage-fr}{0.0}
\defV{lang-q-reap25-usage-id}{0.0}
\defV{lang-q-reap25-usage-other}{\ensuremath{-}1.4}
\defV{lang-q-reap50-usage-en}{\ensuremath{-}0.5}
\defV{lang-q-reap50-usage-pt}{\ensuremath{-}0.8}
\defV{lang-q-reap50-usage-es}{\ensuremath{-}3.4}
\defV{lang-q-reap50-usage-fr}{\ensuremath{-}0.9}
\defV{lang-q-reap50-usage-id}{+0.8}
\defV{lang-q-reap50-usage-other}{\ensuremath{-}2.1}
\defV{lang-q-ream25-usage-en}{0.0}
\defV{lang-q-ream25-usage-pt}{\ensuremath{-}2.0}
\defV{lang-q-ream25-usage-es}{0.0}
\defV{lang-q-ream25-usage-fr}{\ensuremath{-}0.9}
\defV{lang-q-ream25-usage-id}{0.0}
\defV{lang-q-ream25-usage-other}{\ensuremath{-}1.4}
\defV{lang-q-ream50-usage-en}{\ensuremath{-}0.7}
\defV{lang-q-ream50-usage-pt}{\ensuremath{-}3.3}
\defV{lang-q-ream50-usage-es}{\ensuremath{-}4.0}
\defV{lang-q-ream50-usage-fr}{\ensuremath{-}0.9}
\defV{lang-q-ream50-usage-id}{\ensuremath{-}2.3}
\defV{lang-q-ream50-usage-other}{\ensuremath{-}2.5}
\defV{lang-n-reap25-usage-en}{+0.3}
\defV{lang-n-reap25-usage-pt}{+1.2}
\defV{lang-n-reap25-usage-es}{\ensuremath{-}1.1}
\defV{lang-n-reap25-usage-fr}{+0.9}
\defV{lang-n-reap25-usage-id}{0.0}
\defV{lang-n-reap25-usage-other}{+2.1}
\defV{lang-n-reap50-usage-en}{\ensuremath{-}1.0}
\defV{lang-n-reap50-usage-pt}{\ensuremath{-}0.8}
\defV{lang-n-reap50-usage-es}{\ensuremath{-}2.3}
\defV{lang-n-reap50-usage-fr}{+2.7}
\defV{lang-n-reap50-usage-id}{\ensuremath{-}9.1}
\defV{lang-n-reap50-usage-other}{0.0}
\defV{lang-n-ream25-usage-en}{\ensuremath{-}0.7}
\defV{lang-n-ream25-usage-pt}{+0.4}
\defV{lang-n-ream25-usage-es}{\ensuremath{-}2.3}
\defV{lang-n-ream25-usage-fr}{0.0}
\defV{lang-n-ream25-usage-id}{+0.8}
\defV{lang-n-ream25-usage-other}{0.0}
\defV{lang-n-ream50-usage-en}{\ensuremath{-}5.7}
\defV{lang-n-ream50-usage-pt}{\ensuremath{-}3.3}
\defV{lang-n-ream50-usage-es}{\ensuremath{-}2.8}
\defV{lang-n-ream50-usage-fr}{+1.8}
\defV{lang-n-ream50-usage-id}{\ensuremath{-}6.1}
\defV{lang-n-ream50-usage-other}{\ensuremath{-}3.2}
\defV{lang-q-reap25-brand-en}{+4.2}
\defV{lang-q-reap25-brand-pt}{+3.5}
\defV{lang-q-reap25-brand-es}{+2.7}
\defV{lang-q-reap25-brand-fr}{+3.6}
\defV{lang-q-reap25-brand-id}{+1.8}
\defV{lang-q-reap25-brand-other}{+6.6}
\defV{lang-q-reap50-brand-en}{+5.0}
\defV{lang-q-reap50-brand-pt}{+5.0}
\defV{lang-q-reap50-brand-es}{+4.3}
\defV{lang-q-reap50-brand-fr}{+3.0}
\defV{lang-q-reap50-brand-id}{+4.5}
\defV{lang-q-reap50-brand-other}{+2.2}
\defV{lang-q-ream25-brand-en}{+0.4}
\defV{lang-q-ream25-brand-pt}{+0.4}
\defV{lang-q-ream25-brand-es}{\ensuremath{-}2.5}
\defV{lang-q-ream25-brand-fr}{+1.2}
\defV{lang-q-ream25-brand-id}{\ensuremath{-}3.6}
\defV{lang-q-ream25-brand-other}{+0.9}
\defV{lang-q-ream50-brand-en}{+8.3}
\defV{lang-q-ream50-brand-pt}{+4.4}
\defV{lang-q-ream50-brand-es}{+6.1}
\defV{lang-q-ream50-brand-fr}{+6.1}
\defV{lang-q-ream50-brand-id}{+4.5}
\defV{lang-q-ream50-brand-other}{+4.4}
\defV{lang-n-reap25-brand-en}{\ensuremath{-}0.4}
\defV{lang-n-reap25-brand-pt}{\ensuremath{-}6.7}
\defV{lang-n-reap25-brand-es}{\ensuremath{-}3.8}
\defV{lang-n-reap25-brand-fr}{\ensuremath{-}2.4}
\defV{lang-n-reap25-brand-id}{\ensuremath{-}10.7}
\defV{lang-n-reap25-brand-other}{\ensuremath{-}4.1}
\defV{lang-n-reap50-brand-en}{\ensuremath{-}17.7}
\defV{lang-n-reap50-brand-pt}{\ensuremath{-}24.9}
\defV{lang-n-reap50-brand-es}{\ensuremath{-}21.8}
\defV{lang-n-reap50-brand-fr}{\ensuremath{-}18.2}
\defV{lang-n-reap50-brand-id}{\ensuremath{-}21.4}
\defV{lang-n-reap50-brand-other}{\ensuremath{-}20.9}
\defV{lang-n-ream25-brand-en}{\ensuremath{-}7.9}
\defV{lang-n-ream25-brand-pt}{\ensuremath{-}13.1}
\defV{lang-n-ream25-brand-es}{\ensuremath{-}11.5}
\defV{lang-n-ream25-brand-fr}{\ensuremath{-}8.5}
\defV{lang-n-ream25-brand-id}{\ensuremath{-}7.1}
\defV{lang-n-ream25-brand-other}{\ensuremath{-}7.9}
\defV{lang-n-ream50-brand-en}{\ensuremath{-}12.5}
\defV{lang-n-ream50-brand-pt}{\ensuremath{-}18.3}
\defV{lang-n-ream50-brand-es}{\ensuremath{-}18.7}
\defV{lang-n-ream50-brand-fr}{\ensuremath{-}14.5}
\defV{lang-n-ream50-brand-id}{\ensuremath{-}13.4}
\defV{lang-n-ream50-brand-other}{\ensuremath{-}13.6}
\defV{lang-q-reap25-categ-en}{+2.2}
\defV{lang-q-reap25-categ-pt}{+0.7}
\defV{lang-q-reap25-categ-es}{\ensuremath{-}1.1}
\defV{lang-q-reap25-categ-fr}{\ensuremath{-}2.2}
\defV{lang-q-reap25-categ-id}{\ensuremath{-}3.1}
\defV{lang-q-reap25-categ-other}{\ensuremath{-}1.9}
\defV{lang-q-reap50-categ-en}{\ensuremath{-}5.5}
\defV{lang-q-reap50-categ-pt}{\ensuremath{-}5.9}
\defV{lang-q-reap50-categ-es}{\ensuremath{-}7.0}
\defV{lang-q-reap50-categ-fr}{\ensuremath{-}1.8}
\defV{lang-q-reap50-categ-id}{\ensuremath{-}4.6}
\defV{lang-q-reap50-categ-other}{\ensuremath{-}9.0}
\defV{lang-q-ream25-categ-en}{+2.2}
\defV{lang-q-ream25-categ-pt}{\ensuremath{-}4.1}
\defV{lang-q-ream25-categ-es}{\ensuremath{-}5.9}
\defV{lang-q-ream25-categ-fr}{+0.4}
\defV{lang-q-ream25-categ-id}{\ensuremath{-}7.7}
\defV{lang-q-ream25-categ-other}{\ensuremath{-}8.7}
\defV{lang-q-ream50-categ-en}{\ensuremath{-}4.4}
\defV{lang-q-ream50-categ-pt}{\ensuremath{-}12.2}
\defV{lang-q-ream50-categ-es}{\ensuremath{-}5.1}
\defV{lang-q-ream50-categ-fr}{\ensuremath{-}5.2}
\defV{lang-q-ream50-categ-id}{\ensuremath{-}10.8}
\defV{lang-q-ream50-categ-other}{\ensuremath{-}15.1}
\defV{lang-n-reap25-categ-en}{\ensuremath{-}0.7}
\defV{lang-n-reap25-categ-pt}{\ensuremath{-}4.4}
\defV{lang-n-reap25-categ-es}{0.0}
\defV{lang-n-reap25-categ-fr}{\ensuremath{-}3.0}
\defV{lang-n-reap25-categ-id}{\ensuremath{-}7.7}
\defV{lang-n-reap25-categ-other}{\ensuremath{-}3.9}
\defV{lang-n-reap50-categ-en}{\ensuremath{-}9.2}
\defV{lang-n-reap50-categ-pt}{\ensuremath{-}5.9}
\defV{lang-n-reap50-categ-es}{\ensuremath{-}7.0}
\defV{lang-n-reap50-categ-fr}{\ensuremath{-}17.0}
\defV{lang-n-reap50-categ-id}{\ensuremath{-}20.0}
\defV{lang-n-reap50-categ-other}{\ensuremath{-}12.6}
\defV{lang-n-ream25-categ-en}{\ensuremath{-}13.6}
\defV{lang-n-ream25-categ-pt}{\ensuremath{-}5.9}
\defV{lang-n-ream25-categ-es}{\ensuremath{-}4.0}
\defV{lang-n-ream25-categ-fr}{\ensuremath{-}10.0}
\defV{lang-n-ream25-categ-id}{\ensuremath{-}21.5}
\defV{lang-n-ream25-categ-other}{\ensuremath{-}7.2}
\defV{lang-n-ream50-categ-en}{\ensuremath{-}25.0}
\defV{lang-n-ream50-categ-pt}{\ensuremath{-}16.6}
\defV{lang-n-ream50-categ-es}{\ensuremath{-}8.8}
\defV{lang-n-ream50-categ-fr}{\ensuremath{-}22.1}
\defV{lang-n-ream50-categ-id}{\ensuremath{-}33.8}
\defV{lang-n-ream50-categ-other}{\ensuremath{-}17.3}
\defV{lang-q-reap25-domains-en}{+1.2}
\defV{lang-q-reap25-domains-pt}{\ensuremath{-}9.4}
\defV{lang-q-reap25-domains-es}{\ensuremath{-}13.5}
\defV{lang-q-reap25-domains-fr}{\ensuremath{-}13.0}
\defV{lang-q-reap25-domains-id}{\ensuremath{-}42.8}
\defV{lang-q-reap25-domains-other}{\ensuremath{-}6.3}
\defV{lang-q-reap50-domains-en}{\ensuremath{-}1.5}
\defV{lang-q-reap50-domains-pt}{\ensuremath{-}15.2}
\defV{lang-q-reap50-domains-es}{\ensuremath{-}25.2}
\defV{lang-q-reap50-domains-fr}{\ensuremath{-}45.4}
\defV{lang-q-reap50-domains-id}{\ensuremath{-}8.0}
\defV{lang-q-reap50-domains-other}{\ensuremath{-}40.3}
\defV{lang-q-ream25-domains-en}{\ensuremath{-}0.6}
\defV{lang-q-ream25-domains-pt}{\ensuremath{-}68.4}
\defV{lang-q-ream25-domains-es}{\ensuremath{-}35.8}
\defV{lang-q-ream25-domains-fr}{\ensuremath{-}40.0}
\defV{lang-q-ream25-domains-id}{\ensuremath{-}33.8}
\defV{lang-q-ream25-domains-other}{\ensuremath{-}18.0}
\defV{lang-q-ream50-domains-en}{\ensuremath{-}2.6}
\defV{lang-q-ream50-domains-pt}{\ensuremath{-}45.9}
\defV{lang-q-ream50-domains-es}{\ensuremath{-}12.3}
\defV{lang-q-ream50-domains-fr}{\ensuremath{-}48.1}
\defV{lang-q-ream50-domains-id}{\ensuremath{-}7.4}
\defV{lang-q-ream50-domains-other}{\ensuremath{-}58.9}
\defV{lang-n-reap25-domains-en}{+0.6}
\defV{lang-n-reap25-domains-pt}{\ensuremath{-}8.8}
\defV{lang-n-reap25-domains-es}{\ensuremath{-}3.5}
\defV{lang-n-reap25-domains-fr}{\ensuremath{-}2.7}
\defV{lang-n-reap25-domains-id}{+5.8}
\defV{lang-n-reap25-domains-other}{\ensuremath{-}19.0}
\defV{lang-n-reap50-domains-en}{+3.5}
\defV{lang-n-reap50-domains-pt}{\ensuremath{-}21.6}
\defV{lang-n-reap50-domains-es}{\ensuremath{-}17.0}
\defV{lang-n-reap50-domains-fr}{\ensuremath{-}23.8}
\defV{lang-n-reap50-domains-id}{\ensuremath{-}11.6}
\defV{lang-n-reap50-domains-other}{\ensuremath{-}50.1}
\defV{lang-n-ream25-domains-en}{\ensuremath{-}6.4}
\defV{lang-n-ream25-domains-pt}{\ensuremath{-}15.8}
\defV{lang-n-ream25-domains-es}{\ensuremath{-}3.8}
\defV{lang-n-ream25-domains-fr}{\ensuremath{-}40.0}
\defV{lang-n-ream25-domains-id}{\ensuremath{-}22.2}
\defV{lang-n-ream25-domains-other}{\ensuremath{-}30.1}
\defV{lang-n-ream50-domains-en}{\ensuremath{-}13.7}
\defV{lang-n-ream50-domains-pt}{\ensuremath{-}13.5}
\defV{lang-n-ream50-domains-es}{\ensuremath{-}12.0}
\defV{lang-n-ream50-domains-fr}{\ensuremath{-}38.4}
\defV{lang-n-ream50-domains-id}{\ensuremath{-}29.3}
\defV{lang-n-ream50-domains-other}{\ensuremath{-}51.1}
\defV{lang-q-reap25-reach-en}{+1.5}
\defV{lang-q-reap25-reach-pt}{+2.4}
\defV{lang-q-reap25-reach-es}{\ensuremath{-}0.4}
\defV{lang-q-reap25-reach-fr}{\ensuremath{-}0.2}
\defV{lang-q-reap25-reach-id}{+3.0}
\defV{lang-q-reap25-reach-other}{+0.5}
\defV{lang-q-reap50-reach-en}{\ensuremath{-}21.8}
\defV{lang-q-reap50-reach-pt}{\ensuremath{-}23.5}
\defV{lang-q-reap50-reach-es}{\ensuremath{-}26.7}
\defV{lang-q-reap50-reach-fr}{\ensuremath{-}31.5}
\defV{lang-q-reap50-reach-id}{\ensuremath{-}22.2}
\defV{lang-q-reap50-reach-other}{\ensuremath{-}24.0}
\defV{lang-q-ream25-reach-en}{\ensuremath{-}26.7}
\defV{lang-q-ream25-reach-pt}{\ensuremath{-}30.9}
\defV{lang-q-ream25-reach-es}{\ensuremath{-}33.7}
\defV{lang-q-ream25-reach-fr}{\ensuremath{-}33.1}
\defV{lang-q-ream25-reach-id}{\ensuremath{-}20.2}
\defV{lang-q-ream25-reach-other}{\ensuremath{-}27.1}
\defV{lang-q-ream50-reach-en}{\ensuremath{-}40.2}
\defV{lang-q-ream50-reach-pt}{\ensuremath{-}47.1}
\defV{lang-q-ream50-reach-es}{\ensuremath{-}44.4}
\defV{lang-q-ream50-reach-fr}{\ensuremath{-}49.8}
\defV{lang-q-ream50-reach-id}{\ensuremath{-}34.5}
\defV{lang-q-ream50-reach-other}{\ensuremath{-}46.7}
\defV{lang-n-reap25-reach-en}{+3.1}
\defV{lang-n-reap25-reach-pt}{\ensuremath{-}1.2}
\defV{lang-n-reap25-reach-es}{\ensuremath{-}0.5}
\defV{lang-n-reap25-reach-fr}{\ensuremath{-}1.9}
\defV{lang-n-reap25-reach-id}{+6.4}
\defV{lang-n-reap25-reach-other}{\ensuremath{-}0.2}
\defV{lang-n-reap50-reach-en}{\ensuremath{-}13.2}
\defV{lang-n-reap50-reach-pt}{\ensuremath{-}11.4}
\defV{lang-n-reap50-reach-es}{\ensuremath{-}9.8}
\defV{lang-n-reap50-reach-fr}{\ensuremath{-}11.1}
\defV{lang-n-reap50-reach-id}{\ensuremath{-}1.6}
\defV{lang-n-reap50-reach-other}{\ensuremath{-}7.8}
\defV{lang-n-ream25-reach-en}{\ensuremath{-}11.5}
\defV{lang-n-ream25-reach-pt}{\ensuremath{-}9.2}
\defV{lang-n-ream25-reach-es}{\ensuremath{-}7.1}
\defV{lang-n-ream25-reach-fr}{\ensuremath{-}8.6}
\defV{lang-n-ream25-reach-id}{\ensuremath{-}3.1}
\defV{lang-n-ream25-reach-other}{\ensuremath{-}7.3}
\defV{lang-n-ream50-reach-en}{\ensuremath{-}16.3}
\defV{lang-n-ream50-reach-pt}{\ensuremath{-}14.4}
\defV{lang-n-ream50-reach-es}{\ensuremath{-}11.8}
\defV{lang-n-ream50-reach-fr}{\ensuremath{-}13.2}
\defV{lang-n-ream50-reach-id}{\ensuremath{-}5.5}
\defV{lang-n-ream50-reach-other}{\ensuremath{-}10.1}
\defV{dom-en-min}{\ensuremath{-}13.7}
\defV{dom-en-max}{+3.5}
\defV{dom-en-q-min}{\ensuremath{-}2.6}
\defV{dom-en-q-max}{+1.2}
\defV{dom-nonen-min}{\ensuremath{-}39.2}
\defV{dom-nonen-max}{\ensuremath{-}5.6}
\defV{dom-worst-cell}{\ensuremath{-}68.4}
\defV{xppl-q-base-usage}{1.000}
\defV{ownppl-q-base-usage}{1.00}
\defV{refnll-q-base-usage}{0.00}
\defV{cappedex-q-base-usage}{0}
\defV{xppl-q-reap25-usage}{1.001}
\defV{ownppl-q-reap25-usage}{1.00}
\defV{refnll-q-reap25-usage}{0.01}
\defV{cappedex-q-reap25-usage}{0}
\defV{xppl-q-reap50-usage}{1.005}
\defV{ownppl-q-reap50-usage}{1.00}
\defV{refnll-q-reap50-usage}{0.01}
\defV{cappedex-q-reap50-usage}{0}
\defV{xppl-q-ream25-usage}{1.003}
\defV{ownppl-q-ream25-usage}{1.00}
\defV{refnll-q-ream25-usage}{0.01}
\defV{cappedex-q-ream25-usage}{0}
\defV{xppl-q-ream50-usage}{1.009}
\defV{ownppl-q-ream50-usage}{1.00}
\defV{refnll-q-ream50-usage}{0.01}
\defV{cappedex-q-ream50-usage}{0}
\defV{xppl-n-base-usage}{1.003}
\defV{ownppl-n-base-usage}{1.00}
\defV{refnll-n-base-usage}{0.01}
\defV{cappedex-n-base-usage}{0}
\defV{xppl-n-reap25-usage}{1.004}
\defV{ownppl-n-reap25-usage}{1.00}
\defV{refnll-n-reap25-usage}{0.01}
\defV{cappedex-n-reap25-usage}{0}
\defV{xppl-n-reap50-usage}{1.007}
\defV{ownppl-n-reap50-usage}{1.00}
\defV{refnll-n-reap50-usage}{0.01}
\defV{cappedex-n-reap50-usage}{0}
\defV{xppl-n-ream25-usage}{1.004}
\defV{ownppl-n-ream25-usage}{1.00}
\defV{refnll-n-ream25-usage}{0.01}
\defV{cappedex-n-ream25-usage}{0}
\defV{xppl-n-ream50-usage}{1.008}
\defV{ownppl-n-ream50-usage}{1.00}
\defV{refnll-n-ream50-usage}{0.01}
\defV{cappedex-n-ream50-usage}{0}
\defV{xppl-q-base-brand}{1.042}
\defV{ownppl-q-base-brand}{1.04}
\defV{refnll-q-base-brand}{2.01}
\defV{cappedex-q-base-brand}{0}
\defV{xppl-q-reap25-brand}{1.073}
\defV{ownppl-q-reap25-brand}{1.04}
\defV{refnll-q-reap25-brand}{1.98}
\defV{cappedex-q-reap25-brand}{0}
\defV{xppl-q-reap50-brand}{1.15}
\defV{ownppl-q-reap50-brand}{1.03}
\defV{refnll-q-reap50-brand}{1.76}
\defV{cappedex-q-reap50-brand}{0}
\defV{xppl-q-ream25-brand}{1.086}
\defV{ownppl-q-ream25-brand}{1.05}
\defV{refnll-q-ream25-brand}{2.05}
\defV{cappedex-q-ream25-brand}{0}
\defV{xppl-q-ream50-brand}{1.16}
\defV{ownppl-q-ream50-brand}{1.02}
\defV{refnll-q-ream50-brand}{1.84}
\defV{cappedex-q-ream50-brand}{0}
\defV{xppl-n-base-brand}{1.074}
\defV{ownppl-n-base-brand}{1.07}
\defV{refnll-n-base-brand}{1.24}
\defV{cappedex-n-base-brand}{0}
\defV{xppl-n-reap25-brand}{1.16}
\defV{ownppl-n-reap25-brand}{1.09}
\defV{refnll-n-reap25-brand}{1.28}
\defV{cappedex-n-reap25-brand}{0}
\defV{xppl-n-reap50-brand}{1.37}
\defV{ownppl-n-reap50-brand}{1.13}
\defV{refnll-n-reap50-brand}{1.27}
\defV{cappedex-n-reap50-brand}{0}
\defV{xppl-n-ream25-brand}{1.26}
\defV{ownppl-n-ream25-brand}{1.11}
\defV{refnll-n-ream25-brand}{1.20}
\defV{cappedex-n-ream25-brand}{0}
\defV{xppl-n-ream50-brand}{1.42}
\defV{ownppl-n-ream50-brand}{1.13}
\defV{refnll-n-ream50-brand}{1.17}
\defV{cappedex-n-ream50-brand}{0}
\defV{xppl-q-base-categ}{1.038}
\defV{ownppl-q-base-categ}{1.04}
\defV{refnll-q-base-categ}{1.71}
\defV{cappedex-q-base-categ}{0}
\defV{xppl-q-reap25-categ}{1.11}
\defV{ownppl-q-reap25-categ}{1.03}
\defV{refnll-q-reap25-categ}{1.69}
\defV{cappedex-q-reap25-categ}{0}
\defV{xppl-q-reap50-categ}{1.32}
\defV{ownppl-q-reap50-categ}{1.04}
\defV{refnll-q-reap50-categ}{1.53}
\defV{cappedex-q-reap50-categ}{0}
\defV{xppl-q-ream25-categ}{1.30}
\defV{ownppl-q-ream25-categ}{1.05}
\defV{refnll-q-ream25-categ}{1.76}
\defV{cappedex-q-ream25-categ}{0}
\defV{xppl-q-ream50-categ}{1.57}
\defV{ownppl-q-ream50-categ}{1.05}
\defV{refnll-q-ream50-categ}{1.50}
\defV{cappedex-q-ream50-categ}{0}
\defV{xppl-n-base-categ}{1.11}
\defV{ownppl-n-base-categ}{1.11}
\defV{refnll-n-base-categ}{0.90}
\defV{cappedex-n-base-categ}{0}
\defV{xppl-n-reap25-categ}{1.21}
\defV{ownppl-n-reap25-categ}{1.11}
\defV{refnll-n-reap25-categ}{0.91}
\defV{cappedex-n-reap25-categ}{1}
\defV{xppl-n-reap50-categ}{1.40}
\defV{ownppl-n-reap50-categ}{1.14}
\defV{refnll-n-reap50-categ}{0.96}
\defV{cappedex-n-reap50-categ}{0}
\defV{xppl-n-ream25-categ}{1.35}
\defV{ownppl-n-ream25-categ}{1.13}
\defV{refnll-n-ream25-categ}{0.95}
\defV{cappedex-n-ream25-categ}{0}
\defV{xppl-n-ream50-categ}{1.63}
\defV{ownppl-n-ream50-categ}{1.18}
\defV{refnll-n-ream50-categ}{1.06}
\defV{cappedex-n-ream50-categ}{0}
\defV{xppl-q-base-domains}{1.19}
\defV{ownppl-q-base-domains}{1.19}
\defV{refnll-q-base-domains}{2.00}
\defV{cappedex-q-base-domains}{0}
\defV{xppl-q-reap25-domains}{1.50}
\defV{ownppl-q-reap25-domains}{1.19}
\defV{refnll-q-reap25-domains}{2.09}
\defV{cappedex-q-reap25-domains}{0}
\defV{xppl-q-reap50-domains}{4.00}
\defV{ownppl-q-reap50-domains}{1.31}
\defV{refnll-q-reap50-domains}{2.25}
\defV{cappedex-q-reap50-domains}{2}
\defV{xppl-q-ream25-domains}{2.13}
\defV{ownppl-q-ream25-domains}{1.21}
\defV{refnll-q-ream25-domains}{2.35}
\defV{cappedex-q-ream25-domains}{0}
\defV{xppl-q-ream50-domains}{5.90}
\defV{ownppl-q-ream50-domains}{1.33}
\defV{refnll-q-ream50-domains}{2.49}
\defV{cappedex-q-ream50-domains}{4}
\defV{xppl-n-base-domains}{2.20}
\defV{ownppl-n-base-domains}{2.20}
\defV{refnll-n-base-domains}{1.53}
\defV{cappedex-n-base-domains}{0}
\defV{xppl-n-reap25-domains}{3.16}
\defV{ownppl-n-reap25-domains}{2.39}
\defV{refnll-n-reap25-domains}{1.57}
\defV{cappedex-n-reap25-domains}{1}
\defV{xppl-n-reap50-domains}{7.11}
\defV{ownppl-n-reap50-domains}{3.57}
\defV{refnll-n-reap50-domains}{1.79}
\defV{cappedex-n-reap50-domains}{16}
\defV{xppl-n-ream25-domains}{7.21}
\defV{ownppl-n-ream25-domains}{3.66}
\defV{refnll-n-ream25-domains}{1.89}
\defV{cappedex-n-ream25-domains}{3}
\defV{xppl-n-ream50-domains}{18.31}
\defV{ownppl-n-ream50-domains}{7.34}
\defV{refnll-n-ream50-domains}{2.11}
\defV{cappedex-n-ream50-domains}{20}
\defV{xppl-q-base-reach}{1.048}
\defV{ownppl-q-base-reach}{1.05}
\defV{refnll-q-base-reach}{0.32}
\defV{cappedex-q-base-reach}{6}
\defV{xppl-q-reap25-reach}{1.065}
\defV{ownppl-q-reap25-reach}{1.05}
\defV{refnll-q-reap25-reach}{0.32}
\defV{cappedex-q-reap25-reach}{14}
\defV{xppl-q-reap50-reach}{1.12}
\defV{ownppl-q-reap50-reach}{1.05}
\defV{refnll-q-reap50-reach}{0.35}
\defV{cappedex-q-reap50-reach}{102}
\defV{xppl-q-ream25-reach}{1.13}
\defV{ownppl-q-ream25-reach}{1.05}
\defV{refnll-q-ream25-reach}{0.37}
\defV{cappedex-q-ream25-reach}{61}
\defV{xppl-q-ream50-reach}{1.20}
\defV{ownppl-q-ream50-reach}{1.06}
\defV{refnll-q-ream50-reach}{0.39}
\defV{cappedex-q-ream50-reach}{215}
\defV{xppl-n-base-reach}{1.089}
\defV{ownppl-n-base-reach}{1.09}
\defV{refnll-n-base-reach}{0.33}
\defV{cappedex-n-base-reach}{38}
\defV{xppl-n-reap25-reach}{1.19}
\defV{ownppl-n-reap25-reach}{1.09}
\defV{refnll-n-reap25-reach}{0.34}
\defV{cappedex-n-reap25-reach}{32}
\defV{xppl-n-reap50-reach}{1.36}
\defV{ownppl-n-reap50-reach}{1.12}
\defV{refnll-n-reap50-reach}{0.39}
\defV{cappedex-n-reap50-reach}{226}
\defV{xppl-n-ream25-reach}{1.34}
\defV{ownppl-n-ream25-reach}{1.12}
\defV{refnll-n-ream25-reach}{0.38}
\defV{cappedex-n-ream25-reach}{166}
\defV{xppl-n-ream50-reach}{1.45}
\defV{ownppl-n-ream50-reach}{1.15}
\defV{refnll-n-ream50-reach}{0.43}
\defV{cappedex-n-ream50-reach}{299}
\defV{xppl-reap-le}{20}
\defV{xppl-reap-lt}{19}
\defV{xppl-cells}{20}
\defV{jsd-n-ream25-usage}{0.001}
\defV{jsdearly-n-ream25-usage}{0.000}
\defV{jsdlate-n-ream25-usage}{n/a}
\defV{jsd-n-ream50-usage}{0.002}
\defV{jsdearly-n-ream50-usage}{0.001}
\defV{jsdlate-n-ream50-usage}{n/a}
\defV{jsd-n-reap25-usage}{0.000}
\defV{jsdearly-n-reap25-usage}{0.000}
\defV{jsdlate-n-reap25-usage}{n/a}
\defV{jsd-n-reap50-usage}{0.001}
\defV{jsdearly-n-reap50-usage}{0.001}
\defV{jsdlate-n-reap50-usage}{n/a}
\defV{jsd-q-ream25-usage}{0.001}
\defV{jsdearly-q-ream25-usage}{0.000}
\defV{jsdlate-q-ream25-usage}{n/a}
\defV{jsd-q-ream50-usage}{0.001}
\defV{jsdearly-q-ream50-usage}{0.000}
\defV{jsdlate-q-ream50-usage}{n/a}
\defV{jsd-q-reap25-usage}{0.000}
\defV{jsdearly-q-reap25-usage}{0.000}
\defV{jsdlate-q-reap25-usage}{n/a}
\defV{jsd-q-reap50-usage}{0.000}
\defV{jsdearly-q-reap50-usage}{0.000}
\defV{jsdlate-q-reap50-usage}{n/a}
\defV{jsd-n-ream25-brand}{0.040}
\defV{jsdearly-n-ream25-brand}{0.063}
\defV{jsdlate-n-ream25-brand}{n/a}
\defV{jsd-n-ream50-brand}{0.059}
\defV{jsdearly-n-ream50-brand}{0.096}
\defV{jsdlate-n-ream50-brand}{n/a}
\defV{jsd-n-reap25-brand}{0.023}
\defV{jsdearly-n-reap25-brand}{0.039}
\defV{jsdlate-n-reap25-brand}{n/a}
\defV{jsd-n-reap50-brand}{0.052}
\defV{jsdearly-n-reap50-brand}{0.078}
\defV{jsdlate-n-reap50-brand}{n/a}
\defV{jsd-q-ream25-brand}{0.010}
\defV{jsdearly-q-ream25-brand}{0.011}
\defV{jsdlate-q-ream25-brand}{n/a}
\defV{jsd-q-ream50-brand}{0.029}
\defV{jsdearly-q-ream50-brand}{0.056}
\defV{jsdlate-q-ream50-brand}{n/a}
\defV{jsd-q-reap25-brand}{0.008}
\defV{jsdearly-q-reap25-brand}{0.012}
\defV{jsdlate-q-reap25-brand}{n/a}
\defV{jsd-q-reap50-brand}{0.024}
\defV{jsdearly-q-reap50-brand}{0.042}
\defV{jsdlate-q-reap50-brand}{n/a}
\defV{jsd-n-ream25-categ}{0.047}
\defV{jsdearly-n-ream25-categ}{0.058}
\defV{jsdlate-n-ream25-categ}{n/a}
\defV{jsd-n-ream50-categ}{0.071}
\defV{jsdearly-n-ream50-categ}{0.085}
\defV{jsdlate-n-ream50-categ}{n/a}
\defV{jsd-n-reap25-categ}{0.018}
\defV{jsdearly-n-reap25-categ}{0.033}
\defV{jsdlate-n-reap25-categ}{0.000}
\defV{jsd-n-reap50-categ}{0.051}
\defV{jsdearly-n-reap50-categ}{0.064}
\defV{jsdlate-n-reap50-categ}{n/a}
\defV{jsd-q-ream25-categ}{0.030}
\defV{jsdearly-q-ream25-categ}{0.036}
\defV{jsdlate-q-ream25-categ}{n/a}
\defV{jsd-q-ream50-categ}{0.049}
\defV{jsdearly-q-ream50-categ}{0.065}
\defV{jsdlate-q-ream50-categ}{n/a}
\defV{jsd-q-reap25-categ}{0.013}
\defV{jsdearly-q-reap25-categ}{0.017}
\defV{jsdlate-q-reap25-categ}{n/a}
\defV{jsd-q-reap50-categ}{0.034}
\defV{jsdearly-q-reap50-categ}{0.043}
\defV{jsdlate-q-reap50-categ}{n/a}
\defV{jsd-n-ream25-domains}{0.142}
\defV{jsdearly-n-ream25-domains}{0.146}
\defV{jsdlate-n-ream25-domains}{0.133}
\defV{jsd-n-ream50-domains}{0.123}
\defV{jsdearly-n-ream50-domains}{0.164}
\defV{jsdlate-n-ream50-domains}{0.062}
\defV{jsd-n-reap25-domains}{0.062}
\defV{jsdearly-n-reap25-domains}{0.036}
\defV{jsdlate-n-reap25-domains}{0.080}
\defV{jsd-n-reap50-domains}{0.060}
\defV{jsdearly-n-reap50-domains}{0.110}
\defV{jsdlate-n-reap50-domains}{0.001}
\defV{jsd-q-ream25-domains}{0.101}
\defV{jsdearly-q-ream25-domains}{0.103}
\defV{jsdlate-q-ream25-domains}{n/a}
\defV{jsd-q-ream50-domains}{0.215}
\defV{jsdearly-q-ream50-domains}{0.201}
\defV{jsdlate-q-ream50-domains}{n/a}
\defV{jsd-q-reap25-domains}{0.056}
\defV{jsdearly-q-reap25-domains}{0.050}
\defV{jsdlate-q-reap25-domains}{n/a}
\defV{jsd-q-reap50-domains}{0.168}
\defV{jsdearly-q-reap50-domains}{0.156}
\defV{jsdlate-q-reap50-domains}{n/a}
\defV{jsd-n-ream25-reach}{0.038}
\defV{jsdearly-n-ream25-reach}{0.105}
\defV{jsdlate-n-ream25-reach}{0.036}
\defV{jsd-n-ream50-reach}{0.048}
\defV{jsdearly-n-ream50-reach}{0.132}
\defV{jsdlate-n-ream50-reach}{0.044}
\defV{jsd-n-reap25-reach}{0.023}
\defV{jsdearly-n-reap25-reach}{0.046}
\defV{jsdlate-n-reap25-reach}{0.022}
\defV{jsd-n-reap50-reach}{0.045}
\defV{jsdearly-n-reap50-reach}{0.090}
\defV{jsdlate-n-reap50-reach}{0.043}
\defV{jsd-q-ream25-reach}{0.018}
\defV{jsdearly-q-ream25-reach}{0.061}
\defV{jsdlate-q-ream25-reach}{0.016}
\defV{jsd-q-ream50-reach}{0.029}
\defV{jsdearly-q-ream50-reach}{0.130}
\defV{jsdlate-q-ream50-reach}{0.026}
\defV{jsd-q-reap25-reach}{0.005}
\defV{jsdearly-q-reap25-reach}{0.022}
\defV{jsdlate-q-reap25-reach}{0.004}
\defV{jsd-q-reap50-reach}{0.018}
\defV{jsdearly-q-reap50-reach}{0.085}
\defV{jsdlate-q-reap50-reach}{0.016}
\defV{jsd-ream-gt}{20}
\defV{jsd-cells}{20}
\defV{div-q-base-domains}{0.378}
\defV{divin-q-base-domains}{1.000}
\defV{div-n-base-domains}{0.494}
\defV{divin-n-base-domains}{0.998}
\defV{div-q-reap25-domains}{0.359}
\defV{divin-q-reap25-domains}{1.000}
\defV{div-q-reap50-domains}{0.332}
\defV{divin-q-reap50-domains}{0.998}
\defV{div-q-ream25-domains}{0.372}
\defV{divin-q-ream25-domains}{1.000}
\defV{div-q-ream50-domains}{0.346}
\defV{divin-q-ream50-domains}{0.998}
\defV{div-n-reap25-domains}{0.492}
\defV{divin-n-reap25-domains}{0.995}
\defV{div-n-reap50-domains}{0.477}
\defV{divin-n-reap50-domains}{0.994}
\defV{div-n-ream25-domains}{0.557}
\defV{divin-n-ream25-domains}{0.997}
\defV{div-n-ream50-domains}{0.572}
\defV{divin-n-ream50-domains}{0.988}
\defV{div-q-base-reach}{0.690}
\defV{divin-q-base-reach}{0.950}
\defV{div-n-base-reach}{0.572}
\defV{divin-n-base-reach}{0.880}
\defV{div-q-reap25-reach}{0.698}
\defV{divin-q-reap25-reach}{0.950}
\defV{div-q-reap50-reach}{0.696}
\defV{divin-q-reap50-reach}{0.928}
\defV{div-q-ream25-reach}{0.702}
\defV{divin-q-ream25-reach}{0.934}
\defV{div-q-ream50-reach}{0.664}
\defV{divin-q-ream50-reach}{0.912}
\defV{div-n-reap25-reach}{0.622}
\defV{divin-n-reap25-reach}{0.907}
\defV{div-n-reap50-reach}{0.491}
\defV{divin-n-reap50-reach}{0.817}
\defV{div-n-ream25-reach}{0.625}
\defV{divin-n-ream25-reach}{0.899}
\defV{div-n-ream50-reach}{0.530}
\defV{divin-n-ream50-reach}{0.821}
\defV{erank-q-base-usage}{1{,}184}
\defV{nrank-q-base-usage}{2{,}046}
\defV{erank-q-base-brand}{1{,}242}
\defV{nrank-q-base-brand}{2{,}046}
\defV{erank-q-base-categ}{1{,}413}
\defV{nrank-q-base-categ}{2{,}046}
\defV{erank-q-base-domains}{892}
\defV{nrank-q-base-domains}{1{,}639}
\defV{erank-q-base-reach}{1{,}272}
\defV{nrank-q-base-reach}{2{,}046}
\defV{erank-q-reap25-usage}{1{,}165}
\defV{nrank-q-reap25-usage}{2{,}046}
\defV{erank-q-reap25-brand}{1{,}230}
\defV{nrank-q-reap25-brand}{2{,}046}
\defV{erank-q-reap25-categ}{1{,}396}
\defV{nrank-q-reap25-categ}{2{,}046}
\defV{erank-q-reap25-domains}{876}
\defV{nrank-q-reap25-domains}{1{,}588}
\defV{erank-q-reap25-reach}{1{,}257}
\defV{nrank-q-reap25-reach}{2{,}046}
\defV{erank-q-reap50-usage}{1{,}104}
\defV{nrank-q-reap50-usage}{2{,}046}
\defV{erank-q-reap50-brand}{1{,}180}
\defV{nrank-q-reap50-brand}{2{,}046}
\defV{erank-q-reap50-categ}{1{,}332}
\defV{nrank-q-reap50-categ}{2{,}046}
\defV{erank-q-reap50-domains}{839}
\defV{nrank-q-reap50-domains}{1{,}457}
\defV{erank-q-reap50-reach}{1{,}226}
\defV{nrank-q-reap50-reach}{2{,}046}
\defV{erank-q-ream25-usage}{1{,}154}
\defV{nrank-q-ream25-usage}{2{,}046}
\defV{erank-q-ream25-brand}{1{,}215}
\defV{nrank-q-ream25-brand}{2{,}046}
\defV{erank-q-ream25-categ}{1{,}379}
\defV{nrank-q-ream25-categ}{2{,}046}
\defV{erank-q-ream25-domains}{873}
\defV{nrank-q-ream25-domains}{1{,}597}
\defV{erank-q-ream25-reach}{1{,}241}
\defV{nrank-q-ream25-reach}{2{,}046}
\defV{erank-q-ream50-usage}{1{,}091}
\defV{nrank-q-ream50-usage}{2{,}046}
\defV{erank-q-ream50-brand}{1{,}150}
\defV{nrank-q-ream50-brand}{2{,}046}
\defV{erank-q-ream50-categ}{1{,}289}
\defV{nrank-q-ream50-categ}{2{,}046}
\defV{erank-q-ream50-domains}{832}
\defV{nrank-q-ream50-domains}{1{,}473}
\defV{erank-q-ream50-reach}{1{,}207}
\defV{nrank-q-ream50-reach}{2{,}046}
\defV{erank-n-base-usage}{1{,}709}
\defV{nrank-n-base-usage}{2{,}687}
\defV{erank-n-base-brand}{1{,}764}
\defV{nrank-n-base-brand}{2{,}687}
\defV{erank-n-base-categ}{1{,}899}
\defV{nrank-n-base-categ}{2{,}687}
\defV{erank-n-base-domains}{1{,}105}
\defV{nrank-n-base-domains}{2{,}191}
\defV{erank-n-base-reach}{1{,}911}
\defV{nrank-n-base-reach}{2{,}687}
\defV{erank-n-reap25-usage}{1{,}704}
\defV{nrank-n-reap25-usage}{2{,}687}
\defV{erank-n-reap25-brand}{1{,}757}
\defV{nrank-n-reap25-brand}{2{,}687}
\defV{erank-n-reap25-categ}{1{,}878}
\defV{nrank-n-reap25-categ}{2{,}687}
\defV{erank-n-reap25-domains}{1{,}092}
\defV{nrank-n-reap25-domains}{2{,}136}
\defV{erank-n-reap25-reach}{1{,}909}
\defV{nrank-n-reap25-reach}{2{,}687}
\defV{erank-n-reap50-usage}{1{,}688}
\defV{nrank-n-reap50-usage}{2{,}687}
\defV{erank-n-reap50-brand}{1{,}734}
\defV{nrank-n-reap50-brand}{2{,}687}
\defV{erank-n-reap50-categ}{1{,}824}
\defV{nrank-n-reap50-categ}{2{,}687}
\defV{erank-n-reap50-domains}{1{,}074}
\defV{nrank-n-reap50-domains}{2{,}094}
\defV{erank-n-reap50-reach}{1{,}909}
\defV{nrank-n-reap50-reach}{2{,}687}
\defV{erank-n-ream25-usage}{1{,}701}
\defV{nrank-n-ream25-usage}{2{,}687}
\defV{erank-n-ream25-brand}{1{,}754}
\defV{nrank-n-ream25-brand}{2{,}687}
\defV{erank-n-ream25-categ}{1{,}878}
\defV{nrank-n-ream25-categ}{2{,}687}
\defV{erank-n-ream25-domains}{1{,}130}
\defV{nrank-n-ream25-domains}{2{,}305}
\defV{erank-n-ream25-reach}{1{,}912}
\defV{nrank-n-ream25-reach}{2{,}687}
\defV{erank-n-ream50-usage}{1{,}709}
\defV{nrank-n-ream50-usage}{2{,}687}
\defV{erank-n-ream50-brand}{1{,}732}
\defV{nrank-n-ream50-brand}{2{,}687}
\defV{erank-n-ream50-categ}{1{,}815}
\defV{nrank-n-ream50-categ}{2{,}687}
\defV{erank-n-ream50-domains}{1{,}116}
\defV{nrank-n-ream50-domains}{2{,}295}
\defV{erank-n-ream50-reach}{1{,}910}
\defV{nrank-n-ream50-reach}{2{,}687}
\defV{erank-maxchange}{8.7}
\defV{dim-q}{2{,}048}
\defV{dim-n}{2{,}688}
\defV{centre-overlap-mean-min}{95.6}
\defV{centre-overlap-mean-max}{98.2}
\defV{centre-overlap-layer-min}{89.1}
\defV{centre-first-layer-identical}{all four}
\defV{ciw-usage-min}{0.6}
\defV{ciw-usage-max}{2.0}
\defV{ciw-other-min}{2.4}
\defV{ciw-other-max}{5.2}
\defV{calib-en-usage}{78}
\defV{calib-rows-usage}{1{,}103}
\defV{calib-en-brand}{19}
\defV{calib-rows-brand}{272}
\defV{calib-en-categ}{20}
\defV{calib-rows-categ}{1{,}067}
\defV{calib-en-domains}{71}
\defV{calib-rows-domains}{3{,}174}
\defV{calib-en-reach}{18}
\defV{calib-rows-reach}{50}
\defV{calib-en-all}{60}
\defV{rmpg-answer-q-reap25-usage-en}{19.8}
\defV{rmpg-answer-q-reap25-usage-pt}{19.3}
\defV{rmpg-answer-q-reap25-usage-es}{20.0}
\defV{rmpg-answer-q-reap25-usage-fr}{19.4}
\defV{rmpg-answer-q-reap25-usage-id}{19.9}
\defV{rmpg-answer-q-reap25-usage-other}{20.1}
\defV{rm-answer-q-reap25-usage}{19.7}
\defV{rmpg-answer-q-reap25-brand-en}{8.3}
\defV{rmpg-answer-q-reap25-brand-pt}{7.7}
\defV{rmpg-answer-q-reap25-brand-es}{7.6}
\defV{rmpg-answer-q-reap25-brand-fr}{7.7}
\defV{rmpg-answer-q-reap25-brand-id}{7.8}
\defV{rmpg-answer-q-reap25-brand-other}{8.2}
\defV{rm-answer-q-reap25-brand}{7.9}
\defV{rmpg-answer-q-reap25-categ-en}{8.9}
\defV{rmpg-answer-q-reap25-categ-pt}{7.4}
\defV{rmpg-answer-q-reap25-categ-es}{7.0}
\defV{rmpg-answer-q-reap25-categ-fr}{7.0}
\defV{rmpg-answer-q-reap25-categ-id}{7.4}
\defV{rmpg-answer-q-reap25-categ-other}{7.0}
\defV{rm-answer-q-reap25-categ}{7.3}
\defV{rmpg-answer-q-reap25-domains-en}{7.6}
\defV{rmpg-answer-q-reap25-domains-pt}{5.6}
\defV{rmpg-answer-q-reap25-domains-es}{5.5}
\defV{rmpg-answer-q-reap25-domains-fr}{5.4}
\defV{rmpg-answer-q-reap25-domains-id}{6.0}
\defV{rmpg-answer-q-reap25-domains-other}{7.5}
\defV{rm-answer-q-reap25-domains}{6.5}
\defV{rmpg-answer-q-reap25-reach-en}{18.5}
\defV{rmpg-answer-q-reap25-reach-pt}{17.3}
\defV{rmpg-answer-q-reap25-reach-es}{17.0}
\defV{rmpg-answer-q-reap25-reach-fr}{17.5}
\defV{rmpg-answer-q-reap25-reach-id}{17.3}
\defV{rmpg-answer-q-reap25-reach-other}{17.6}
\defV{rm-answer-q-reap25-reach}{17.6}
\defV{rmlang-answer-q-reap25-en}{18.4}
\defV{rmlang-answer-q-reap25-pt}{17.0}
\defV{rmlang-answer-q-reap25-es}{16.9}
\defV{rmlang-answer-q-reap25-fr}{17.4}
\defV{rmlang-answer-q-reap25-id}{17.0}
\defV{rmlang-answer-q-reap25-other}{17.5}
\defV{rm-answer-q-reap25-all}{17.4}
\defV{rmpg-answer-q-reap50-usage-en}{38.8}
\defV{rmpg-answer-q-reap50-usage-pt}{38.2}
\defV{rmpg-answer-q-reap50-usage-es}{39.1}
\defV{rmpg-answer-q-reap50-usage-fr}{38.4}
\defV{rmpg-answer-q-reap50-usage-id}{39.0}
\defV{rmpg-answer-q-reap50-usage-other}{39.4}
\defV{rm-answer-q-reap50-usage}{38.8}
\defV{rmpg-answer-q-reap50-brand-en}{25.7}
\defV{rmpg-answer-q-reap50-brand-pt}{24.3}
\defV{rmpg-answer-q-reap50-brand-es}{24.1}
\defV{rmpg-answer-q-reap50-brand-fr}{24.4}
\defV{rmpg-answer-q-reap50-brand-id}{25.0}
\defV{rmpg-answer-q-reap50-brand-other}{25.2}
\defV{rm-answer-q-reap50-brand}{24.8}
\defV{rmpg-answer-q-reap50-categ-en}{26.9}
\defV{rmpg-answer-q-reap50-categ-pt}{24.3}
\defV{rmpg-answer-q-reap50-categ-es}{23.6}
\defV{rmpg-answer-q-reap50-categ-fr}{23.7}
\defV{rmpg-answer-q-reap50-categ-id}{24.9}
\defV{rmpg-answer-q-reap50-categ-other}{24.0}
\defV{rm-answer-q-reap50-categ}{24.3}
\defV{rmpg-answer-q-reap50-domains-en}{27.5}
\defV{rmpg-answer-q-reap50-domains-pt}{23.9}
\defV{rmpg-answer-q-reap50-domains-es}{23.5}
\defV{rmpg-answer-q-reap50-domains-fr}{24.1}
\defV{rmpg-answer-q-reap50-domains-id}{25.2}
\defV{rmpg-answer-q-reap50-domains-other}{28.3}
\defV{rm-answer-q-reap50-domains}{26.0}
\defV{rmpg-answer-q-reap50-reach-en}{36.5}
\defV{rmpg-answer-q-reap50-reach-pt}{35.1}
\defV{rmpg-answer-q-reap50-reach-es}{34.7}
\defV{rmpg-answer-q-reap50-reach-fr}{35.2}
\defV{rmpg-answer-q-reap50-reach-id}{35.1}
\defV{rmpg-answer-q-reap50-reach-other}{35.6}
\defV{rm-answer-q-reap50-reach}{35.4}
\defV{rmlang-answer-q-reap50-en}{36.4}
\defV{rmlang-answer-q-reap50-pt}{34.9}
\defV{rmlang-answer-q-reap50-es}{34.6}
\defV{rmlang-answer-q-reap50-fr}{35.1}
\defV{rmlang-answer-q-reap50-id}{34.8}
\defV{rmlang-answer-q-reap50-other}{35.5}
\defV{rm-answer-q-reap50-all}{35.3}
\defV{rmpg-answer-q-ream25-usage-en}{19.7}
\defV{rmpg-answer-q-ream25-usage-pt}{19.3}
\defV{rmpg-answer-q-ream25-usage-es}{20.0}
\defV{rmpg-answer-q-ream25-usage-fr}{19.4}
\defV{rmpg-answer-q-ream25-usage-id}{20.0}
\defV{rmpg-answer-q-ream25-usage-other}{20.1}
\defV{rm-answer-q-ream25-usage}{19.7}
\defV{rmpg-answer-q-ream25-brand-en}{8.0}
\defV{rmpg-answer-q-ream25-brand-pt}{7.4}
\defV{rmpg-answer-q-ream25-brand-es}{7.3}
\defV{rmpg-answer-q-ream25-brand-fr}{7.4}
\defV{rmpg-answer-q-ream25-brand-id}{7.6}
\defV{rmpg-answer-q-ream25-brand-other}{7.9}
\defV{rm-answer-q-ream25-brand}{7.6}
\defV{rmpg-answer-q-ream25-categ-en}{8.7}
\defV{rmpg-answer-q-ream25-categ-pt}{7.4}
\defV{rmpg-answer-q-ream25-categ-es}{7.1}
\defV{rmpg-answer-q-ream25-categ-fr}{7.1}
\defV{rmpg-answer-q-ream25-categ-id}{7.3}
\defV{rmpg-answer-q-ream25-categ-other}{7.0}
\defV{rm-answer-q-ream25-categ}{7.3}
\defV{rmpg-answer-q-ream25-domains-en}{6.6}
\defV{rmpg-answer-q-ream25-domains-pt}{4.8}
\defV{rmpg-answer-q-ream25-domains-es}{4.6}
\defV{rmpg-answer-q-ream25-domains-fr}{4.6}
\defV{rmpg-answer-q-ream25-domains-id}{5.3}
\defV{rmpg-answer-q-ream25-domains-other}{6.5}
\defV{rm-answer-q-ream25-domains}{5.6}
\defV{rmpg-answer-q-ream25-reach-en}{19.6}
\defV{rmpg-answer-q-ream25-reach-pt}{18.2}
\defV{rmpg-answer-q-ream25-reach-es}{18.0}
\defV{rmpg-answer-q-ream25-reach-fr}{18.4}
\defV{rmpg-answer-q-ream25-reach-id}{18.3}
\defV{rmpg-answer-q-ream25-reach-other}{18.6}
\defV{rm-answer-q-ream25-reach}{18.5}
\defV{rmlang-answer-q-ream25-en}{19.5}
\defV{rmlang-answer-q-ream25-pt}{17.9}
\defV{rmlang-answer-q-ream25-es}{17.9}
\defV{rmlang-answer-q-ream25-fr}{18.3}
\defV{rmlang-answer-q-ream25-id}{18.0}
\defV{rmlang-answer-q-ream25-other}{18.4}
\defV{rm-answer-q-ream25-all}{18.4}
\defV{rmpg-answer-q-ream50-usage-en}{39.0}
\defV{rmpg-answer-q-ream50-usage-pt}{38.4}
\defV{rmpg-answer-q-ream50-usage-es}{39.2}
\defV{rmpg-answer-q-ream50-usage-fr}{38.6}
\defV{rmpg-answer-q-ream50-usage-id}{39.1}
\defV{rmpg-answer-q-ream50-usage-other}{39.5}
\defV{rm-answer-q-ream50-usage}{39.0}
\defV{rmpg-answer-q-ream50-brand-en}{25.7}
\defV{rmpg-answer-q-ream50-brand-pt}{24.3}
\defV{rmpg-answer-q-ream50-brand-es}{24.1}
\defV{rmpg-answer-q-ream50-brand-fr}{24.4}
\defV{rmpg-answer-q-ream50-brand-id}{24.9}
\defV{rmpg-answer-q-ream50-brand-other}{25.1}
\defV{rm-answer-q-ream50-brand}{24.8}
\defV{rmpg-answer-q-ream50-categ-en}{27.0}
\defV{rmpg-answer-q-ream50-categ-pt}{24.1}
\defV{rmpg-answer-q-ream50-categ-es}{23.7}
\defV{rmpg-answer-q-ream50-categ-fr}{23.8}
\defV{rmpg-answer-q-ream50-categ-id}{24.7}
\defV{rmpg-answer-q-ream50-categ-other}{23.9}
\defV{rm-answer-q-ream50-categ}{24.2}
\defV{rmpg-answer-q-ream50-domains-en}{27.1}
\defV{rmpg-answer-q-ream50-domains-pt}{23.3}
\defV{rmpg-answer-q-ream50-domains-es}{23.1}
\defV{rmpg-answer-q-ream50-domains-fr}{23.8}
\defV{rmpg-answer-q-ream50-domains-id}{24.5}
\defV{rmpg-answer-q-ream50-domains-other}{27.7}
\defV{rm-answer-q-ream50-domains}{25.4}
\defV{rmpg-answer-q-ream50-reach-en}{38.6}
\defV{rmpg-answer-q-ream50-reach-pt}{37.0}
\defV{rmpg-answer-q-ream50-reach-es}{36.7}
\defV{rmpg-answer-q-ream50-reach-fr}{37.2}
\defV{rmpg-answer-q-ream50-reach-id}{37.1}
\defV{rmpg-answer-q-ream50-reach-other}{37.5}
\defV{rm-answer-q-ream50-reach}{37.4}
\defV{rmlang-answer-q-ream50-en}{38.5}
\defV{rmlang-answer-q-ream50-pt}{36.7}
\defV{rmlang-answer-q-ream50-es}{36.6}
\defV{rmlang-answer-q-ream50-fr}{37.1}
\defV{rmlang-answer-q-ream50-id}{36.7}
\defV{rmlang-answer-q-ream50-other}{37.4}
\defV{rm-answer-q-ream50-all}{37.2}
\defV{rmpg-answer-n-reap25-usage-en}{18.9}
\defV{rmpg-answer-n-reap25-usage-pt}{18.8}
\defV{rmpg-answer-n-reap25-usage-es}{18.9}
\defV{rmpg-answer-n-reap25-usage-fr}{18.9}
\defV{rmpg-answer-n-reap25-usage-id}{19.0}
\defV{rmpg-answer-n-reap25-usage-other}{18.8}
\defV{rm-answer-n-reap25-usage}{18.9}
\defV{rmpg-answer-n-reap25-brand-en}{16.5}
\defV{rmpg-answer-n-reap25-brand-pt}{15.8}
\defV{rmpg-answer-n-reap25-brand-es}{15.7}
\defV{rmpg-answer-n-reap25-brand-fr}{15.9}
\defV{rmpg-answer-n-reap25-brand-id}{16.3}
\defV{rmpg-answer-n-reap25-brand-other}{16.3}
\defV{rm-answer-n-reap25-brand}{16.1}
\defV{rmpg-answer-n-reap25-categ-en}{16.5}
\defV{rmpg-answer-n-reap25-categ-pt}{16.2}
\defV{rmpg-answer-n-reap25-categ-es}{16.2}
\defV{rmpg-answer-n-reap25-categ-fr}{17.0}
\defV{rmpg-answer-n-reap25-categ-id}{17.3}
\defV{rmpg-answer-n-reap25-categ-other}{17.0}
\defV{rm-answer-n-reap25-categ}{16.8}
\defV{rmpg-answer-n-reap25-domains-en}{16.3}
\defV{rmpg-answer-n-reap25-domains-pt}{15.7}
\defV{rmpg-answer-n-reap25-domains-es}{15.5}
\defV{rmpg-answer-n-reap25-domains-fr}{16.1}
\defV{rmpg-answer-n-reap25-domains-id}{16.8}
\defV{rmpg-answer-n-reap25-domains-other}{18.1}
\defV{rm-answer-n-reap25-domains}{16.8}
\defV{rmpg-answer-n-reap25-reach-en}{22.1}
\defV{rmpg-answer-n-reap25-reach-pt}{21.7}
\defV{rmpg-answer-n-reap25-reach-es}{21.5}
\defV{rmpg-answer-n-reap25-reach-fr}{21.7}
\defV{rmpg-answer-n-reap25-reach-id}{21.7}
\defV{rmpg-answer-n-reap25-reach-other}{22.0}
\defV{rm-answer-n-reap25-reach}{21.8}
\defV{rmlang-answer-n-reap25-en}{22.0}
\defV{rmlang-answer-n-reap25-pt}{21.5}
\defV{rmlang-answer-n-reap25-es}{21.5}
\defV{rmlang-answer-n-reap25-fr}{21.6}
\defV{rmlang-answer-n-reap25-id}{21.6}
\defV{rmlang-answer-n-reap25-other}{21.9}
\defV{rm-answer-n-reap25-all}{21.7}
\defV{rmpg-answer-n-reap50-usage-en}{39.3}
\defV{rmpg-answer-n-reap50-usage-pt}{39.1}
\defV{rmpg-answer-n-reap50-usage-es}{39.2}
\defV{rmpg-answer-n-reap50-usage-fr}{39.2}
\defV{rmpg-answer-n-reap50-usage-id}{39.2}
\defV{rmpg-answer-n-reap50-usage-other}{39.2}
\defV{rm-answer-n-reap50-usage}{39.2}
\defV{rmpg-answer-n-reap50-brand-en}{37.1}
\defV{rmpg-answer-n-reap50-brand-pt}{35.3}
\defV{rmpg-answer-n-reap50-brand-es}{35.1}
\defV{rmpg-answer-n-reap50-brand-fr}{35.7}
\defV{rmpg-answer-n-reap50-brand-id}{36.2}
\defV{rmpg-answer-n-reap50-brand-other}{37.4}
\defV{rm-answer-n-reap50-brand}{36.1}
\defV{rmpg-answer-n-reap50-categ-en}{37.3}
\defV{rmpg-answer-n-reap50-categ-pt}{36.3}
\defV{rmpg-answer-n-reap50-categ-es}{36.7}
\defV{rmpg-answer-n-reap50-categ-fr}{38.0}
\defV{rmpg-answer-n-reap50-categ-id}{38.5}
\defV{rmpg-answer-n-reap50-categ-other}{39.3}
\defV{rm-answer-n-reap50-categ}{38.1}
\defV{rmpg-answer-n-reap50-domains-en}{39.3}
\defV{rmpg-answer-n-reap50-domains-pt}{37.6}
\defV{rmpg-answer-n-reap50-domains-es}{36.2}
\defV{rmpg-answer-n-reap50-domains-fr}{37.5}
\defV{rmpg-answer-n-reap50-domains-id}{39.4}
\defV{rmpg-answer-n-reap50-domains-other}{42.8}
\defV{rm-answer-n-reap50-domains}{39.6}
\defV{rmpg-answer-n-reap50-reach-en}{45.2}
\defV{rmpg-answer-n-reap50-reach-pt}{44.4}
\defV{rmpg-answer-n-reap50-reach-es}{44.2}
\defV{rmpg-answer-n-reap50-reach-fr}{44.4}
\defV{rmpg-answer-n-reap50-reach-id}{44.5}
\defV{rmpg-answer-n-reap50-reach-other}{45.0}
\defV{rm-answer-n-reap50-reach}{44.7}
\defV{rmlang-answer-n-reap50-en}{45.0}
\defV{rmlang-answer-n-reap50-pt}{44.1}
\defV{rmlang-answer-n-reap50-es}{44.1}
\defV{rmlang-answer-n-reap50-fr}{44.3}
\defV{rmlang-answer-n-reap50-id}{44.3}
\defV{rmlang-answer-n-reap50-other}{45.0}
\defV{rm-answer-n-reap50-all}{44.5}
\defV{rmpg-answer-n-ream25-usage-en}{18.8}
\defV{rmpg-answer-n-ream25-usage-pt}{18.7}
\defV{rmpg-answer-n-ream25-usage-es}{18.8}
\defV{rmpg-answer-n-ream25-usage-fr}{18.7}
\defV{rmpg-answer-n-ream25-usage-id}{18.8}
\defV{rmpg-answer-n-ream25-usage-other}{18.7}
\defV{rm-answer-n-ream25-usage}{18.8}
\defV{rmpg-answer-n-ream25-brand-en}{16.4}
\defV{rmpg-answer-n-ream25-brand-pt}{15.7}
\defV{rmpg-answer-n-ream25-brand-es}{15.6}
\defV{rmpg-answer-n-ream25-brand-fr}{15.8}
\defV{rmpg-answer-n-ream25-brand-id}{16.2}
\defV{rmpg-answer-n-ream25-brand-other}{16.2}
\defV{rm-answer-n-ream25-brand}{16.0}
\defV{rmpg-answer-n-ream25-categ-en}{15.9}
\defV{rmpg-answer-n-ream25-categ-pt}{15.9}
\defV{rmpg-answer-n-ream25-categ-es}{15.7}
\defV{rmpg-answer-n-ream25-categ-fr}{16.5}
\defV{rmpg-answer-n-ream25-categ-id}{17.0}
\defV{rmpg-answer-n-ream25-categ-other}{16.8}
\defV{rm-answer-n-ream25-categ}{16.4}
\defV{rmpg-answer-n-ream25-domains-en}{16.1}
\defV{rmpg-answer-n-ream25-domains-pt}{15.6}
\defV{rmpg-answer-n-ream25-domains-es}{15.4}
\defV{rmpg-answer-n-ream25-domains-fr}{15.8}
\defV{rmpg-answer-n-ream25-domains-id}{17.2}
\defV{rmpg-answer-n-ream25-domains-other}{18.5}
\defV{rm-answer-n-ream25-domains}{16.9}
\defV{rmpg-answer-n-ream25-reach-en}{21.9}
\defV{rmpg-answer-n-ream25-reach-pt}{21.4}
\defV{rmpg-answer-n-ream25-reach-es}{21.3}
\defV{rmpg-answer-n-ream25-reach-fr}{21.5}
\defV{rmpg-answer-n-ream25-reach-id}{21.5}
\defV{rmpg-answer-n-ream25-reach-other}{21.8}
\defV{rm-answer-n-ream25-reach}{21.6}
\defV{rmlang-answer-n-ream25-en}{21.7}
\defV{rmlang-answer-n-ream25-pt}{21.3}
\defV{rmlang-answer-n-ream25-es}{21.3}
\defV{rmlang-answer-n-ream25-fr}{21.4}
\defV{rmlang-answer-n-ream25-id}{21.4}
\defV{rmlang-answer-n-ream25-other}{21.7}
\defV{rm-answer-n-ream25-all}{21.5}
\defV{rmpg-answer-n-ream50-usage-en}{39.2}
\defV{rmpg-answer-n-ream50-usage-pt}{38.9}
\defV{rmpg-answer-n-ream50-usage-es}{39.1}
\defV{rmpg-answer-n-ream50-usage-fr}{39.0}
\defV{rmpg-answer-n-ream50-usage-id}{39.0}
\defV{rmpg-answer-n-ream50-usage-other}{39.1}
\defV{rm-answer-n-ream50-usage}{39.1}
\defV{rmpg-answer-n-ream50-brand-en}{37.1}
\defV{rmpg-answer-n-ream50-brand-pt}{35.4}
\defV{rmpg-answer-n-ream50-brand-es}{35.2}
\defV{rmpg-answer-n-ream50-brand-fr}{35.7}
\defV{rmpg-answer-n-ream50-brand-id}{36.4}
\defV{rmpg-answer-n-ream50-brand-other}{37.3}
\defV{rm-answer-n-ream50-brand}{36.2}
\defV{rmpg-answer-n-ream50-categ-en}{38.2}
\defV{rmpg-answer-n-ream50-categ-pt}{37.4}
\defV{rmpg-answer-n-ream50-categ-es}{37.6}
\defV{rmpg-answer-n-ream50-categ-fr}{38.7}
\defV{rmpg-answer-n-ream50-categ-id}{39.5}
\defV{rmpg-answer-n-ream50-categ-other}{40.3}
\defV{rm-answer-n-ream50-categ}{39.0}
\defV{rmpg-answer-n-ream50-domains-en}{39.7}
\defV{rmpg-answer-n-ream50-domains-pt}{38.2}
\defV{rmpg-answer-n-ream50-domains-es}{36.9}
\defV{rmpg-answer-n-ream50-domains-fr}{38.4}
\defV{rmpg-answer-n-ream50-domains-id}{40.2}
\defV{rmpg-answer-n-ream50-domains-other}{43.1}
\defV{rm-answer-n-ream50-domains}{40.2}
\defV{rmpg-answer-n-ream50-reach-en}{44.9}
\defV{rmpg-answer-n-ream50-reach-pt}{44.1}
\defV{rmpg-answer-n-ream50-reach-es}{43.9}
\defV{rmpg-answer-n-ream50-reach-fr}{44.1}
\defV{rmpg-answer-n-ream50-reach-id}{44.3}
\defV{rmpg-answer-n-ream50-reach-other}{44.7}
\defV{rm-answer-n-ream50-reach}{44.4}
\defV{rmlang-answer-n-ream50-en}{44.7}
\defV{rmlang-answer-n-ream50-pt}{43.8}
\defV{rmlang-answer-n-ream50-es}{43.8}
\defV{rmlang-answer-n-ream50-fr}{44.1}
\defV{rmlang-answer-n-ream50-id}{44.1}
\defV{rmlang-answer-n-ream50-other}{44.7}
\defV{rm-answer-n-ream50-all}{44.3}
\defV{rmpg-prompt-q-reap25-usage-en}{9.6}
\defV{rmpg-prompt-q-reap25-usage-pt}{6.2}
\defV{rmpg-prompt-q-reap25-usage-es}{13.5}
\defV{rmpg-prompt-q-reap25-usage-fr}{7.1}
\defV{rmpg-prompt-q-reap25-usage-id}{6.8}
\defV{rmpg-prompt-q-reap25-usage-other}{9.1}
\defV{rm-prompt-q-reap25-usage}{9.0}
\defV{rmpg-prompt-q-reap25-brand-en}{7.3}
\defV{rmpg-prompt-q-reap25-brand-pt}{7.1}
\defV{rmpg-prompt-q-reap25-brand-es}{7.5}
\defV{rmpg-prompt-q-reap25-brand-fr}{6.5}
\defV{rmpg-prompt-q-reap25-brand-id}{6.7}
\defV{rmpg-prompt-q-reap25-brand-other}{7.0}
\defV{rm-prompt-q-reap25-brand}{7.1}
\defV{rmpg-prompt-q-reap25-categ-en}{8.1}
\defV{rmpg-prompt-q-reap25-categ-pt}{7.2}
\defV{rmpg-prompt-q-reap25-categ-es}{6.7}
\defV{rmpg-prompt-q-reap25-categ-fr}{6.2}
\defV{rmpg-prompt-q-reap25-categ-id}{9.1}
\defV{rmpg-prompt-q-reap25-categ-other}{7.3}
\defV{rm-prompt-q-reap25-categ}{7.2}
\defV{rmpg-prompt-q-reap25-domains-en}{6.6}
\defV{rmpg-prompt-q-reap25-domains-pt}{6.5}
\defV{rmpg-prompt-q-reap25-domains-es}{6.5}
\defV{rmpg-prompt-q-reap25-domains-fr}{6.5}
\defV{rmpg-prompt-q-reap25-domains-id}{6.4}
\defV{rmpg-prompt-q-reap25-domains-other}{6.5}
\defV{rm-prompt-q-reap25-domains}{6.5}
\defV{rmpg-prompt-q-reap25-reach-en}{8.5}
\defV{rmpg-prompt-q-reap25-reach-pt}{8.5}
\defV{rmpg-prompt-q-reap25-reach-es}{8.4}
\defV{rmpg-prompt-q-reap25-reach-fr}{8.4}
\defV{rmpg-prompt-q-reap25-reach-id}{8.5}
\defV{rmpg-prompt-q-reap25-reach-other}{8.5}
\defV{rm-prompt-q-reap25-reach}{8.5}
\defV{rmlang-prompt-q-reap25-en}{8.4}
\defV{rmlang-prompt-q-reap25-pt}{7.7}
\defV{rmlang-prompt-q-reap25-es}{8.3}
\defV{rmlang-prompt-q-reap25-fr}{8.0}
\defV{rmlang-prompt-q-reap25-id}{8.0}
\defV{rmlang-prompt-q-reap25-other}{8.2}
\defV{rm-prompt-q-reap25-all}{8.2}
\defV{rmpg-prompt-q-reap50-usage-en}{29.4}
\defV{rmpg-prompt-q-reap50-usage-pt}{24.1}
\defV{rmpg-prompt-q-reap50-usage-es}{31.8}
\defV{rmpg-prompt-q-reap50-usage-fr}{26.4}
\defV{rmpg-prompt-q-reap50-usage-id}{24.5}
\defV{rmpg-prompt-q-reap50-usage-other}{30.3}
\defV{rm-prompt-q-reap50-usage}{28.6}
\defV{rmpg-prompt-q-reap50-brand-en}{24.9}
\defV{rmpg-prompt-q-reap50-brand-pt}{24.6}
\defV{rmpg-prompt-q-reap50-brand-es}{25.0}
\defV{rmpg-prompt-q-reap50-brand-fr}{23.9}
\defV{rmpg-prompt-q-reap50-brand-id}{24.1}
\defV{rmpg-prompt-q-reap50-brand-other}{24.9}
\defV{rm-prompt-q-reap50-brand}{24.7}
\defV{rmpg-prompt-q-reap50-categ-en}{27.9}
\defV{rmpg-prompt-q-reap50-categ-pt}{25.9}
\defV{rmpg-prompt-q-reap50-categ-es}{25.8}
\defV{rmpg-prompt-q-reap50-categ-fr}{24.8}
\defV{rmpg-prompt-q-reap50-categ-id}{29.1}
\defV{rmpg-prompt-q-reap50-categ-other}{26.4}
\defV{rm-prompt-q-reap50-categ}{26.3}
\defV{rmpg-prompt-q-reap50-domains-en}{24.0}
\defV{rmpg-prompt-q-reap50-domains-pt}{23.9}
\defV{rmpg-prompt-q-reap50-domains-es}{23.8}
\defV{rmpg-prompt-q-reap50-domains-fr}{23.7}
\defV{rmpg-prompt-q-reap50-domains-id}{23.6}
\defV{rmpg-prompt-q-reap50-domains-other}{23.8}
\defV{rm-prompt-q-reap50-domains}{23.8}
\defV{rmpg-prompt-q-reap50-reach-en}{24.8}
\defV{rmpg-prompt-q-reap50-reach-pt}{24.8}
\defV{rmpg-prompt-q-reap50-reach-es}{24.7}
\defV{rmpg-prompt-q-reap50-reach-fr}{24.7}
\defV{rmpg-prompt-q-reap50-reach-id}{24.8}
\defV{rmpg-prompt-q-reap50-reach-other}{24.8}
\defV{rm-prompt-q-reap50-reach}{24.8}
\defV{rmlang-prompt-q-reap50-en}{25.5}
\defV{rmlang-prompt-q-reap50-pt}{24.7}
\defV{rmlang-prompt-q-reap50-es}{24.9}
\defV{rmlang-prompt-q-reap50-fr}{24.6}
\defV{rmlang-prompt-q-reap50-id}{24.7}
\defV{rmlang-prompt-q-reap50-other}{25.1}
\defV{rm-prompt-q-reap50-all}{25.0}
\defV{rmpg-prompt-q-ream25-usage-en}{9.3}
\defV{rmpg-prompt-q-ream25-usage-pt}{6.0}
\defV{rmpg-prompt-q-ream25-usage-es}{14.0}
\defV{rmpg-prompt-q-ream25-usage-fr}{6.7}
\defV{rmpg-prompt-q-ream25-usage-id}{6.5}
\defV{rmpg-prompt-q-ream25-usage-other}{9.0}
\defV{rm-prompt-q-ream25-usage}{8.9}
\defV{rmpg-prompt-q-ream25-brand-en}{6.8}
\defV{rmpg-prompt-q-ream25-brand-pt}{6.7}
\defV{rmpg-prompt-q-ream25-brand-es}{7.1}
\defV{rmpg-prompt-q-ream25-brand-fr}{6.0}
\defV{rmpg-prompt-q-ream25-brand-id}{6.2}
\defV{rmpg-prompt-q-ream25-brand-other}{6.5}
\defV{rm-prompt-q-ream25-brand}{6.7}
\defV{rmpg-prompt-q-ream25-categ-en}{7.9}
\defV{rmpg-prompt-q-ream25-categ-pt}{7.1}
\defV{rmpg-prompt-q-ream25-categ-es}{6.7}
\defV{rmpg-prompt-q-ream25-categ-fr}{6.2}
\defV{rmpg-prompt-q-ream25-categ-id}{9.1}
\defV{rmpg-prompt-q-ream25-categ-other}{7.2}
\defV{rm-prompt-q-ream25-categ}{7.1}
\defV{rmpg-prompt-q-ream25-domains-en}{6.2}
\defV{rmpg-prompt-q-ream25-domains-pt}{6.1}
\defV{rmpg-prompt-q-ream25-domains-es}{6.1}
\defV{rmpg-prompt-q-ream25-domains-fr}{6.1}
\defV{rmpg-prompt-q-ream25-domains-id}{6.1}
\defV{rmpg-prompt-q-ream25-domains-other}{6.2}
\defV{rm-prompt-q-ream25-domains}{6.1}
\defV{rmpg-prompt-q-ream25-reach-en}{8.6}
\defV{rmpg-prompt-q-ream25-reach-pt}{8.6}
\defV{rmpg-prompt-q-ream25-reach-es}{8.5}
\defV{rmpg-prompt-q-ream25-reach-fr}{8.5}
\defV{rmpg-prompt-q-ream25-reach-id}{8.6}
\defV{rmpg-prompt-q-ream25-reach-other}{8.6}
\defV{rm-prompt-q-ream25-reach}{8.6}
\defV{rmlang-prompt-q-ream25-en}{8.3}
\defV{rmlang-prompt-q-ream25-pt}{7.6}
\defV{rmlang-prompt-q-ream25-es}{8.4}
\defV{rmlang-prompt-q-ream25-fr}{8.0}
\defV{rmlang-prompt-q-ream25-id}{8.0}
\defV{rmlang-prompt-q-ream25-other}{8.2}
\defV{rm-prompt-q-ream25-all}{8.1}
\defV{rmpg-prompt-q-ream50-usage-en}{29.5}
\defV{rmpg-prompt-q-ream50-usage-pt}{23.6}
\defV{rmpg-prompt-q-ream50-usage-es}{32.8}
\defV{rmpg-prompt-q-ream50-usage-fr}{26.0}
\defV{rmpg-prompt-q-ream50-usage-id}{24.1}
\defV{rmpg-prompt-q-ream50-usage-other}{30.0}
\defV{rm-prompt-q-ream50-usage}{28.6}
\defV{rmpg-prompt-q-ream50-brand-en}{24.8}
\defV{rmpg-prompt-q-ream50-brand-pt}{24.4}
\defV{rmpg-prompt-q-ream50-brand-es}{24.9}
\defV{rmpg-prompt-q-ream50-brand-fr}{23.7}
\defV{rmpg-prompt-q-ream50-brand-id}{23.9}
\defV{rmpg-prompt-q-ream50-brand-other}{24.6}
\defV{rm-prompt-q-ream50-brand}{24.5}
\defV{rmpg-prompt-q-ream50-categ-en}{27.9}
\defV{rmpg-prompt-q-ream50-categ-pt}{25.8}
\defV{rmpg-prompt-q-ream50-categ-es}{25.8}
\defV{rmpg-prompt-q-ream50-categ-fr}{24.8}
\defV{rmpg-prompt-q-ream50-categ-id}{28.8}
\defV{rmpg-prompt-q-ream50-categ-other}{26.1}
\defV{rm-prompt-q-ream50-categ}{26.2}
\defV{rmpg-prompt-q-ream50-domains-en}{23.9}
\defV{rmpg-prompt-q-ream50-domains-pt}{23.8}
\defV{rmpg-prompt-q-ream50-domains-es}{23.7}
\defV{rmpg-prompt-q-ream50-domains-fr}{23.6}
\defV{rmpg-prompt-q-ream50-domains-id}{23.6}
\defV{rmpg-prompt-q-ream50-domains-other}{23.7}
\defV{rm-prompt-q-ream50-domains}{23.7}
\defV{rmpg-prompt-q-ream50-reach-en}{25.4}
\defV{rmpg-prompt-q-ream50-reach-pt}{25.3}
\defV{rmpg-prompt-q-ream50-reach-es}{25.2}
\defV{rmpg-prompt-q-ream50-reach-fr}{25.2}
\defV{rmpg-prompt-q-ream50-reach-id}{25.3}
\defV{rmpg-prompt-q-ream50-reach-other}{25.3}
\defV{rm-prompt-q-ream50-reach}{25.3}
\defV{rmlang-prompt-q-ream50-en}{25.8}
\defV{rmlang-prompt-q-ream50-pt}{24.8}
\defV{rmlang-prompt-q-ream50-es}{25.3}
\defV{rmlang-prompt-q-ream50-fr}{25.0}
\defV{rmlang-prompt-q-ream50-id}{25.0}
\defV{rmlang-prompt-q-ream50-other}{25.4}
\defV{rm-prompt-q-ream50-all}{25.3}
\defV{rmpg-prompt-n-reap25-usage-en}{17.5}
\defV{rmpg-prompt-n-reap25-usage-pt}{15.9}
\defV{rmpg-prompt-n-reap25-usage-es}{19.0}
\defV{rmpg-prompt-n-reap25-usage-fr}{17.5}
\defV{rmpg-prompt-n-reap25-usage-id}{15.2}
\defV{rmpg-prompt-n-reap25-usage-other}{19.3}
\defV{rm-prompt-n-reap25-usage}{17.6}
\defV{rmpg-prompt-n-reap25-brand-en}{16.7}
\defV{rmpg-prompt-n-reap25-brand-pt}{17.3}
\defV{rmpg-prompt-n-reap25-brand-es}{17.2}
\defV{rmpg-prompt-n-reap25-brand-fr}{16.6}
\defV{rmpg-prompt-n-reap25-brand-id}{16.7}
\defV{rmpg-prompt-n-reap25-brand-other}{17.4}
\defV{rm-prompt-n-reap25-brand}{17.1}
\defV{rmpg-prompt-n-reap25-categ-en}{17.2}
\defV{rmpg-prompt-n-reap25-categ-pt}{17.3}
\defV{rmpg-prompt-n-reap25-categ-es}{18.3}
\defV{rmpg-prompt-n-reap25-categ-fr}{18.0}
\defV{rmpg-prompt-n-reap25-categ-id}{18.9}
\defV{rmpg-prompt-n-reap25-categ-other}{18.4}
\defV{rm-prompt-n-reap25-categ}{18.2}
\defV{rmpg-prompt-n-reap25-domains-en}{13.2}
\defV{rmpg-prompt-n-reap25-domains-pt}{13.0}
\defV{rmpg-prompt-n-reap25-domains-es}{13.0}
\defV{rmpg-prompt-n-reap25-domains-fr}{13.0}
\defV{rmpg-prompt-n-reap25-domains-id}{13.2}
\defV{rmpg-prompt-n-reap25-domains-other}{13.3}
\defV{rm-prompt-n-reap25-domains}{13.1}
\defV{rmpg-prompt-n-reap25-reach-en}{17.3}
\defV{rmpg-prompt-n-reap25-reach-pt}{17.3}
\defV{rmpg-prompt-n-reap25-reach-es}{17.3}
\defV{rmpg-prompt-n-reap25-reach-fr}{17.3}
\defV{rmpg-prompt-n-reap25-reach-id}{17.3}
\defV{rmpg-prompt-n-reap25-reach-other}{17.4}
\defV{rm-prompt-n-reap25-reach}{17.3}
\defV{rmlang-prompt-n-reap25-en}{17.2}
\defV{rmlang-prompt-n-reap25-pt}{17.2}
\defV{rmlang-prompt-n-reap25-es}{17.3}
\defV{rmlang-prompt-n-reap25-fr}{17.2}
\defV{rmlang-prompt-n-reap25-id}{17.0}
\defV{rmlang-prompt-n-reap25-other}{17.4}
\defV{rm-prompt-n-reap25-all}{17.3}
\defV{rmpg-prompt-n-reap50-usage-en}{39.5}
\defV{rmpg-prompt-n-reap50-usage-pt}{36.5}
\defV{rmpg-prompt-n-reap50-usage-es}{40.8}
\defV{rmpg-prompt-n-reap50-usage-fr}{39.2}
\defV{rmpg-prompt-n-reap50-usage-id}{35.4}
\defV{rmpg-prompt-n-reap50-usage-other}{41.8}
\defV{rm-prompt-n-reap50-usage}{39.5}
\defV{rmpg-prompt-n-reap50-brand-en}{38.1}
\defV{rmpg-prompt-n-reap50-brand-pt}{38.5}
\defV{rmpg-prompt-n-reap50-brand-es}{38.3}
\defV{rmpg-prompt-n-reap50-brand-fr}{37.7}
\defV{rmpg-prompt-n-reap50-brand-id}{37.9}
\defV{rmpg-prompt-n-reap50-brand-other}{39.5}
\defV{rm-prompt-n-reap50-brand}{38.4}
\defV{rmpg-prompt-n-reap50-categ-en}{40.1}
\defV{rmpg-prompt-n-reap50-categ-pt}{39.6}
\defV{rmpg-prompt-n-reap50-categ-es}{41.5}
\defV{rmpg-prompt-n-reap50-categ-fr}{41.2}
\defV{rmpg-prompt-n-reap50-categ-id}{41.9}
\defV{rmpg-prompt-n-reap50-categ-other}{43.1}
\defV{rm-prompt-n-reap50-categ}{42.0}
\defV{rmpg-prompt-n-reap50-domains-en}{31.2}
\defV{rmpg-prompt-n-reap50-domains-pt}{30.8}
\defV{rmpg-prompt-n-reap50-domains-es}{30.8}
\defV{rmpg-prompt-n-reap50-domains-fr}{30.8}
\defV{rmpg-prompt-n-reap50-domains-id}{30.9}
\defV{rmpg-prompt-n-reap50-domains-other}{31.1}
\defV{rm-prompt-n-reap50-domains}{31.0}
\defV{rmpg-prompt-n-reap50-reach-en}{39.4}
\defV{rmpg-prompt-n-reap50-reach-pt}{39.3}
\defV{rmpg-prompt-n-reap50-reach-es}{39.2}
\defV{rmpg-prompt-n-reap50-reach-fr}{39.2}
\defV{rmpg-prompt-n-reap50-reach-id}{39.3}
\defV{rmpg-prompt-n-reap50-reach-other}{39.4}
\defV{rm-prompt-n-reap50-reach}{39.3}
\defV{rmlang-prompt-n-reap50-en}{39.1}
\defV{rmlang-prompt-n-reap50-pt}{38.7}
\defV{rmlang-prompt-n-reap50-es}{39.1}
\defV{rmlang-prompt-n-reap50-fr}{39.1}
\defV{rmlang-prompt-n-reap50-id}{38.6}
\defV{rmlang-prompt-n-reap50-other}{39.7}
\defV{rm-prompt-n-reap50-all}{39.2}
\defV{rmpg-prompt-n-ream25-usage-en}{17.4}
\defV{rmpg-prompt-n-ream25-usage-pt}{15.9}
\defV{rmpg-prompt-n-ream25-usage-es}{19.1}
\defV{rmpg-prompt-n-ream25-usage-fr}{17.4}
\defV{rmpg-prompt-n-ream25-usage-id}{15.2}
\defV{rmpg-prompt-n-ream25-usage-other}{19.4}
\defV{rm-prompt-n-ream25-usage}{17.6}
\defV{rmpg-prompt-n-ream25-brand-en}{16.7}
\defV{rmpg-prompt-n-ream25-brand-pt}{17.3}
\defV{rmpg-prompt-n-ream25-brand-es}{17.2}
\defV{rmpg-prompt-n-ream25-brand-fr}{16.6}
\defV{rmpg-prompt-n-ream25-brand-id}{16.7}
\defV{rmpg-prompt-n-ream25-brand-other}{17.5}
\defV{rm-prompt-n-ream25-brand}{17.1}
\defV{rmpg-prompt-n-ream25-categ-en}{17.1}
\defV{rmpg-prompt-n-ream25-categ-pt}{17.2}
\defV{rmpg-prompt-n-ream25-categ-es}{18.1}
\defV{rmpg-prompt-n-ream25-categ-fr}{17.8}
\defV{rmpg-prompt-n-ream25-categ-id}{18.9}
\defV{rmpg-prompt-n-ream25-categ-other}{18.4}
\defV{rm-prompt-n-ream25-categ}{18.1}
\defV{rmpg-prompt-n-ream25-domains-en}{13.0}
\defV{rmpg-prompt-n-ream25-domains-pt}{12.9}
\defV{rmpg-prompt-n-ream25-domains-es}{12.9}
\defV{rmpg-prompt-n-ream25-domains-fr}{12.9}
\defV{rmpg-prompt-n-ream25-domains-id}{13.0}
\defV{rmpg-prompt-n-ream25-domains-other}{13.1}
\defV{rm-prompt-n-ream25-domains}{13.0}
\defV{rmpg-prompt-n-ream25-reach-en}{17.2}
\defV{rmpg-prompt-n-ream25-reach-pt}{17.2}
\defV{rmpg-prompt-n-ream25-reach-es}{17.1}
\defV{rmpg-prompt-n-ream25-reach-fr}{17.1}
\defV{rmpg-prompt-n-ream25-reach-id}{17.1}
\defV{rmpg-prompt-n-ream25-reach-other}{17.2}
\defV{rm-prompt-n-ream25-reach}{17.1}
\defV{rmlang-prompt-n-ream25-en}{17.0}
\defV{rmlang-prompt-n-ream25-pt}{17.1}
\defV{rmlang-prompt-n-ream25-es}{17.2}
\defV{rmlang-prompt-n-ream25-fr}{17.0}
\defV{rmlang-prompt-n-ream25-id}{16.9}
\defV{rmlang-prompt-n-ream25-other}{17.3}
\defV{rm-prompt-n-ream25-all}{17.1}
\defV{rmpg-prompt-n-ream50-usage-en}{39.6}
\defV{rmpg-prompt-n-ream50-usage-pt}{36.9}
\defV{rmpg-prompt-n-ream50-usage-es}{40.7}
\defV{rmpg-prompt-n-ream50-usage-fr}{39.2}
\defV{rmpg-prompt-n-ream50-usage-id}{35.8}
\defV{rmpg-prompt-n-ream50-usage-other}{42.2}
\defV{rm-prompt-n-ream50-usage}{39.6}
\defV{rmpg-prompt-n-ream50-brand-en}{38.3}
\defV{rmpg-prompt-n-ream50-brand-pt}{38.8}
\defV{rmpg-prompt-n-ream50-brand-es}{38.5}
\defV{rmpg-prompt-n-ream50-brand-fr}{38.0}
\defV{rmpg-prompt-n-ream50-brand-id}{38.4}
\defV{rmpg-prompt-n-ream50-brand-other}{39.8}
\defV{rm-prompt-n-ream50-brand}{38.7}
\defV{rmpg-prompt-n-ream50-categ-en}{40.4}
\defV{rmpg-prompt-n-ream50-categ-pt}{40.0}
\defV{rmpg-prompt-n-ream50-categ-es}{41.8}
\defV{rmpg-prompt-n-ream50-categ-fr}{41.4}
\defV{rmpg-prompt-n-ream50-categ-id}{42.4}
\defV{rmpg-prompt-n-ream50-categ-other}{43.4}
\defV{rm-prompt-n-ream50-categ}{42.3}
\defV{rmpg-prompt-n-ream50-domains-en}{31.5}
\defV{rmpg-prompt-n-ream50-domains-pt}{31.1}
\defV{rmpg-prompt-n-ream50-domains-es}{31.0}
\defV{rmpg-prompt-n-ream50-domains-fr}{31.1}
\defV{rmpg-prompt-n-ream50-domains-id}{31.2}
\defV{rmpg-prompt-n-ream50-domains-other}{31.4}
\defV{rm-prompt-n-ream50-domains}{31.2}
\defV{rmpg-prompt-n-ream50-reach-en}{39.3}
\defV{rmpg-prompt-n-ream50-reach-pt}{39.2}
\defV{rmpg-prompt-n-ream50-reach-es}{39.1}
\defV{rmpg-prompt-n-ream50-reach-fr}{39.1}
\defV{rmpg-prompt-n-ream50-reach-id}{39.2}
\defV{rmpg-prompt-n-ream50-reach-other}{39.3}
\defV{rm-prompt-n-ream50-reach}{39.2}
\defV{rmlang-prompt-n-ream50-en}{39.1}
\defV{rmlang-prompt-n-ream50-pt}{38.9}
\defV{rmlang-prompt-n-ream50-es}{39.1}
\defV{rmlang-prompt-n-ream50-fr}{39.0}
\defV{rmlang-prompt-n-ream50-id}{38.7}
\defV{rmlang-prompt-n-ream50-other}{39.6}
\defV{rm-prompt-n-ream50-all}{39.2}
\defV{rmlang-q-reap25-min}{16.9}
\defV{rmlang-q-reap25-max}{18.4}
\defV{rm-reap-ream-maxdiff}{2.0}
\defV{params-q-base}{30.5}
\defV{ckpt-q-base}{61.1}
\defV{gpupeak-q-base}{60.5}
\defV{gpuw-q-base}{56.9}
\defV{params-q-reap25}{23.3}
\defV{ckpt-q-reap25}{46.6}
\defV{gpupeak-q-reap25}{47.0}
\defV{gpuw-q-reap25}{43.4}
\defV{params-q-reap50}{16.0}
\defV{ckpt-q-reap50}{32.1}
\defV{gpupeak-q-reap50}{33.5}
\defV{gpuw-q-reap50}{29.9}
\defV{params-q-ream25}{23.3}
\defV{ckpt-q-ream25}{46.6}
\defV{gpupeak-q-ream25}{47.0}
\defV{gpuw-q-ream25}{43.4}
\defV{params-q-ream50}{16.0}
\defV{ckpt-q-ream50}{32.1}
\defV{gpupeak-q-ream50}{33.5}
\defV{gpuw-q-ream50}{29.9}
\defV{params-n-base}{31.6}
\defV{ckpt-n-base}{65.8}
\defV{gpupeak-n-base}{64.6}
\defV{gpuw-n-base}{58.8}
\defV{params-n-reap25}{24.2}
\defV{ckpt-n-reap25}{51.1}
\defV{gpupeak-n-reap25}{50.9}
\defV{gpuw-n-reap25}{45.1}
\defV{params-n-reap50}{16.9}
\defV{ckpt-n-reap50}{36.4}
\defV{gpupeak-n-reap50}{37.2}
\defV{gpuw-n-reap50}{31.5}
\defV{params-n-ream25}{24.2}
\defV{ckpt-n-ream25}{51.1}
\defV{gpupeak-n-ream25}{50.9}
\defV{gpuw-n-ream25}{45.1}
\defV{params-n-ream50}{16.9}
\defV{ckpt-n-ream50}{36.4}
\defV{gpupeak-n-ream50}{37.2}
\defV{gpuw-n-ream50}{31.5}
\defV{memred-25-min}{21}
\defV{memred-25-max}{24}
\defV{memred-50-min}{42}
\defV{memred-50-max}{48}
\defV{chours-n-ream25}{2.4}
\defV{cpeak-n-ream25}{58.3}
\defV{chours-n-ream50}{2.2}
\defV{cpeak-n-ream50}{58.3}
\defV{chours-n-reap25}{2.1}
\defV{cpeak-n-reap25}{42.7}
\defV{chours-n-reap50}{1.9}
\defV{cpeak-n-reap50}{42.7}
\defV{chours-q-ream25}{4.4}
\defV{cpeak-q-ream25}{33.6}
\defV{chours-q-ream50}{4.0}
\defV{cpeak-q-ream50}{33.6}
\defV{chours-q-reap25}{4.1}
\defV{cpeak-q-reap25}{26.0}
\defV{chours-q-reap50}{3.7}
\defV{cpeak-q-reap50}{26.0}
\defV{ream-time-min}{7}
\defV{ream-time-max}{16}
\defV{ream-mem-min}{29}
\defV{ream-mem-max}{37}
\defV{domsec-q-base}{11}
\defV{domsec-q-reap25}{11}
\defV{domsec-q-reap50}{367}
\defV{domsec-q-ream25}{10}
\defV{domsec-q-ream50}{699}
\defV{domsec-n-base}{28}
\defV{domsec-n-reap25}{216}
\defV{domsec-n-reap50}{1{,}256}
\defV{domsec-n-ream25}{592}
\defV{domsec-n-ream50}{1{,}424}
\defV{domslow-min}{7}
\defV{domslow-max}{66}

\providecommand{\defV}[2]{\expandafter\def\csname V@#1\endcsname{#2}}
\defV{r-opener-q-base}{3.3}
\defV{r-wronglang-q-base}{0.2}
\defV{r-undet-q-base}{1.7}
\defV{r-notok-q-base}{2.0}
\defV{r-wrongen-q-base}{0.0}
\defV{r-deten-q-base}{0.0}
\defV{r-pass-opener-q-base}{0.0}
\defV{r-pass-wrong-q-base}{25.0}
\defV{r-pass-undet-q-base}{100.0}
\defV{r-pass-neither-q-base}{95.8}
\defV{r-passen-q-base}{98.2}
\defV{r-passnonen-q-base}{92.6}
\defV{r-opener-q-reap25}{11.5}
\defV{r-wronglang-q-reap25}{0.2}
\defV{r-undet-q-reap25}{1.8}
\defV{r-notok-q-reap25}{2.1}
\defV{r-wrongen-q-reap25}{0.0}
\defV{r-deten-q-reap25}{0.0}
\defV{r-pass-opener-q-reap25}{1.0}
\defV{r-pass-wrong-q-reap25}{50.0}
\defV{r-pass-undet-q-reap25}{96.7}
\defV{r-pass-neither-q-reap25}{86.3}
\defV{r-passen-q-reap25}{99.4}
\defV{r-passnonen-q-reap25}{76.6}
\defV{r-nonen-loss-q-reap25}{16.0}
\defV{r-cf-pass-q-reap25}{82.6}
\defV{r-drift-explains-q-reap25}{6.0}
\defV{r-drift-explains-share-q-reap25}{37.3}
\defV{r-opener-q-reap50}{17.0}
\defV{r-wronglang-q-reap50}{6.7}
\defV{r-undet-q-reap50}{2.2}
\defV{r-notok-q-reap50}{8.9}
\defV{r-wrongen-q-reap50}{97.3}
\defV{r-deten-q-reap50}{6.5}
\defV{r-pass-opener-q-reap50}{0.0}
\defV{r-pass-wrong-q-reap50}{0.0}
\defV{r-pass-undet-q-reap50}{81.1}
\defV{r-pass-neither-q-reap50}{86.7}
\defV{r-passen-q-reap50}{96.8}
\defV{r-passnonen-q-reap50}{66.0}
\defV{r-nonen-loss-q-reap50}{26.5}
\defV{r-cf-pass-q-reap50}{82.5}
\defV{r-drift-explains-q-reap50}{16.5}
\defV{r-drift-explains-share-q-reap50}{62.2}
\defV{r-opener-q-ream25}{39.6}
\defV{r-wronglang-q-ream25}{2.4}
\defV{r-undet-q-ream25}{1.8}
\defV{r-notok-q-ream25}{4.2}
\defV{r-wrongen-q-ream25}{53.8}
\defV{r-deten-q-ream25}{1.3}
\defV{r-pass-opener-q-ream25}{0.9}
\defV{r-pass-wrong-q-ream25}{10.3}
\defV{r-pass-undet-q-ream25}{80.0}
\defV{r-pass-neither-q-ream25}{94.4}
\defV{r-passen-q-ream25}{97.7}
\defV{r-passnonen-q-ream25}{55.1}
\defV{r-nonen-loss-q-ream25}{37.5}
\defV{r-cf-pass-q-ream25}{89.7}
\defV{r-drift-explains-q-ream25}{34.6}
\defV{r-drift-explains-share-q-ream25}{92.3}
\defV{r-opener-q-ream50}{23.2}
\defV{r-wronglang-q-ream50}{10.0}
\defV{r-undet-q-ream50}{0.7}
\defV{r-notok-q-ream50}{10.7}
\defV{r-wrongen-q-ream50}{89.8}
\defV{r-deten-q-ream50}{9.0}
\defV{r-pass-opener-q-ream50}{2.1}
\defV{r-pass-wrong-q-ream50}{1.8}
\defV{r-pass-undet-q-ream50}{45.5}
\defV{r-pass-neither-q-ream50}{84.5}
\defV{r-passen-q-ream50}{95.6}
\defV{r-passnonen-q-ream50}{56.8}
\defV{r-nonen-loss-q-ream50}{35.8}
\defV{r-cf-pass-q-ream50}{80.2}
\defV{r-drift-explains-q-ream50}{23.4}
\defV{r-drift-explains-share-q-ream50}{65.4}
\defV{r-opener-n-base}{19.4}
\defV{r-wronglang-n-base}{14.0}
\defV{r-undet-n-base}{1.4}
\defV{r-notok-n-base}{15.4}
\defV{r-wrongen-n-base}{84.5}
\defV{r-deten-n-base}{11.8}
\defV{r-pass-opener-n-base}{24.5}
\defV{r-pass-wrong-n-base}{3.9}
\defV{r-pass-undet-n-base}{83.3}
\defV{r-pass-neither-n-base}{69.4}
\defV{r-passen-n-base}{92.7}
\defV{r-passnonen-n-base}{51.7}
\defV{r-opener-n-reap25}{25.8}
\defV{r-wronglang-n-reap25}{15.0}
\defV{r-undet-n-reap25}{0.7}
\defV{r-notok-n-reap25}{15.7}
\defV{r-wrongen-n-reap25}{90.0}
\defV{r-deten-n-reap25}{13.5}
\defV{r-pass-opener-n-reap25}{29.0}
\defV{r-pass-wrong-n-reap25}{1.6}
\defV{r-pass-undet-n-reap25}{81.8}
\defV{r-pass-neither-n-reap25}{61.9}
\defV{r-passen-n-reap25}{93.3}
\defV{r-passnonen-n-reap25}{44.5}
\defV{r-nonen-loss-n-reap25}{7.2}
\defV{r-cf-pass-n-reap25}{47.3}
\defV{r-drift-explains-n-reap25}{2.8}
\defV{r-drift-explains-share-n-reap25}{38.1}
\defV{r-opener-n-reap50}{36.9}
\defV{r-wronglang-n-reap50}{25.4}
\defV{r-undet-n-reap50}{1.3}
\defV{r-notok-n-reap50}{26.7}
\defV{r-wrongen-n-reap50}{90.3}
\defV{r-deten-n-reap50}{22.9}
\defV{r-pass-opener-n-reap50}{20.6}
\defV{r-pass-wrong-n-reap50}{1.7}
\defV{r-pass-undet-n-reap50}{4.5}
\defV{r-pass-neither-n-reap50}{45.1}
\defV{r-passen-n-reap50}{96.2}
\defV{r-passnonen-n-reap50}{24.5}
\defV{r-nonen-loss-n-reap50}{27.3}
\defV{r-cf-pass-n-reap50}{33.8}
\defV{r-drift-explains-n-reap50}{9.3}
\defV{r-drift-explains-share-n-reap50}{34.2}
\defV{r-opener-n-ream25}{26.2}
\defV{r-wronglang-n-ream25}{12.7}
\defV{r-undet-n-ream25}{11.2}
\defV{r-notok-n-ream25}{23.9}
\defV{r-wrongen-n-ream25}{53.3}
\defV{r-deten-n-ream25}{6.8}
\defV{r-pass-opener-n-ream25}{15.9}
\defV{r-pass-wrong-n-ream25}{1.4}
\defV{r-pass-undet-n-ream25}{31.7}
\defV{r-pass-neither-n-ream25}{45.0}
\defV{r-passen-n-ream25}{86.3}
\defV{r-passnonen-n-ream25}{30.4}
\defV{r-nonen-loss-n-ream25}{21.4}
\defV{r-cf-pass-n-ream25}{34.8}
\defV{r-drift-explains-n-ream25}{4.4}
\defV{r-drift-explains-share-n-ream25}{20.8}
\defV{r-opener-n-ream50}{18.8}
\defV{r-wronglang-n-ream50}{16.0}
\defV{r-undet-n-ream50}{5.7}
\defV{r-notok-n-ream50}{21.7}
\defV{r-wrongen-n-ream50}{62.4}
\defV{r-deten-n-ream50}{10.0}
\defV{r-pass-opener-n-ream50}{11.5}
\defV{r-pass-wrong-n-ream50}{1.5}
\defV{r-pass-undet-n-ream50}{6.4}
\defV{r-pass-neither-n-ream50}{32.3}
\defV{r-passen-n-ream50}{78.9}
\defV{r-passnonen-n-ream50}{22.0}
\defV{r-nonen-loss-n-ream50}{29.8}
\defV{r-cf-pass-n-ream50}{23.4}
\defV{r-drift-explains-n-ream50}{1.4}
\defV{r-drift-explains-share-n-ream50}{4.8}
\defV{r-ream-groups-25}{16, 16 and 3}
\defV{r-ream-absorbed-25}{15}
\defV{r-ream-groups-50}{32, 32 and 3}
\defV{r-ream-absorbed-50}{31}
\defV{r-ream-layers-q}{48}
\defV{r-ream-layers-n}{23}
\defV{r-ream-groups-q25}{16, 16 and 3}
\defV{r-ream-grouplayers-q25}{48}
\defV{r-ream-top3own-q25}{48}
\defV{r-ream-top3reap-q25}{22}
\defV{r-ream-reaprank3-q25}{4.1}
\defV{r-ream-centrew-q25-big}{0.268}
\defV{r-ream-centrew-q25-big-min}{0.140}
\defV{r-ream-centrew-q25-big-max}{0.992}
\defV{r-ream-centrew-q25-third}{0.629}
\defV{r-ream-centre-route-q25}{6.2}
\defV{r-ream-centre-route-tm-q25}{4.6}
\defV{r-ream-centre-route-tmin-q25}{2.6}
\defV{r-ream-centre-route-tmax-q25}{6.2}
\defV{r-ream-centre-route-prompt-q25}{4.6}
\defV{r-ream-centre-route-prompt-tm-q25}{4.0}
\defV{r-ream-overlap-q25}{97.6}
\defV{r-ream-overlap-min-q25}{95.8}
\defV{r-ream-groups-q50}{32, 32 and 3}
\defV{r-ream-grouplayers-q50}{48}
\defV{r-ream-top3own-q50}{48}
\defV{r-ream-top3reap-q50}{18}
\defV{r-ream-reaprank3-q50}{4.6}
\defV{r-ream-centrew-q50-big}{0.130}
\defV{r-ream-centrew-q50-big-min}{0.066}
\defV{r-ream-centrew-q50-big-max}{0.983}
\defV{r-ream-centrew-q50-third}{0.556}
\defV{r-ream-centre-route-q50}{5.8}
\defV{r-ream-centre-route-tm-q50}{4.6}
\defV{r-ream-centre-route-tmin-q50}{2.9}
\defV{r-ream-centre-route-tmax-q50}{5.9}
\defV{r-ream-centre-route-prompt-q50}{4.7}
\defV{r-ream-centre-route-prompt-tm-q50}{4.1}
\defV{r-ream-overlap-q50}{95.6}
\defV{r-ream-overlap-min-q50}{89.1}
\defV{r-ream-groups-n25}{16, 16 and 3}
\defV{r-ream-grouplayers-n25}{23}
\defV{r-ream-top3own-n25}{23}
\defV{r-ream-top3reap-n25}{12}
\defV{r-ream-reaprank3-n25}{4.5}
\defV{r-ream-centrew-n25-big}{0.282}
\defV{r-ream-centrew-n25-big-min}{0.141}
\defV{r-ream-centrew-n25-big-max}{0.627}
\defV{r-ream-centrew-n25-third}{0.679}
\defV{r-ream-centre-route-n25}{3.8}
\defV{r-ream-centre-route-tm-n25}{3.1}
\defV{r-ream-centre-route-tmin-n25}{2.5}
\defV{r-ream-centre-route-tmax-n25}{3.8}
\defV{r-ream-centre-route-prompt-n25}{3.8}
\defV{r-ream-centre-route-prompt-tm-n25}{4.0}
\defV{r-ream-overlap-n25}{98.2}
\defV{r-ream-overlap-min-n25}{94.8}
\defV{r-ream-groups-n50}{32, 32 and 3}
\defV{r-ream-grouplayers-n50}{23}
\defV{r-ream-top3own-n50}{23}
\defV{r-ream-top3reap-n50}{12}
\defV{r-ream-reaprank3-n50}{4.7}
\defV{r-ream-centrew-n50-big}{0.129}
\defV{r-ream-centrew-n50-big-min}{0.062}
\defV{r-ream-centrew-n50-big-max}{0.418}
\defV{r-ream-centrew-n50-third}{0.605}
\defV{r-ream-centre-route-n50}{2.9}
\defV{r-ream-centre-route-tm-n50}{3.0}
\defV{r-ream-centre-route-tmin-n50}{2.8}
\defV{r-ream-centre-route-tmax-n50}{3.1}
\defV{r-ream-centre-route-prompt-n50}{3.4}
\defV{r-ream-centre-route-prompt-tm-n50}{3.9}
\defV{r-ream-overlap-n50}{96.8}
\defV{r-ream-overlap-min-n50}{92.2}
\defV{r-ream-uniform}{2.3}
\defV{r-h2hprod-q25}{+51.0}
\defV{r-h2hprodlo-q25}{+49.3}
\defV{r-h2hprodhi-q25}{+52.6}
\defV{r-h2hprodp-q25}{p < 0.001}
\defV{r-h2hprod-reap-q25}{51.3}
\defV{r-h2hprod-ream-q25}{0.4}
\defV{r-h2hprod-q50}{+35.8}
\defV{r-h2hprodlo-q50}{+34.0}
\defV{r-h2hprodhi-q50}{+37.5}
\defV{r-h2hprodp-q50}{p < 0.001}
\defV{r-h2hprod-reap-q50}{37.7}
\defV{r-h2hprod-ream-q50}{1.9}
\defV{r-h2hprod-n25}{+9.2}
\defV{r-h2hprodlo-n25}{+7.9}
\defV{r-h2hprodhi-n25}{+10.4}
\defV{r-h2hprodp-n25}{p < 0.001}
\defV{r-h2hprod-reap-n25}{12.1}
\defV{r-h2hprod-ream-n25}{2.9}
\defV{r-h2hprod-n50}{+2.5}
\defV{r-h2hprodlo-n50}{+1.9}
\defV{r-h2hprodhi-n50}{+3.1}
\defV{r-h2hprodp-n50}{p < 0.001}
\defV{r-h2hprod-reap-n50}{2.6}
\defV{r-h2hprod-ream-n50}{0.1}
\defV{r-h2hprod-nsig}{4}
\defV{r-loss-q-reap25}{\ensuremath{-}0.9}
\defV{r-gateloss-sym-q-reap25}{\ensuremath{-}0.9}
\defV{r-qualloss-sym-q-reap25}{+0.1}
\defV{r-gateshare-sym-q-reap25}{n/a}
\defV{r-gateshare-bfirst-q-reap25}{n/a}
\defV{r-gateshare-cfirst-q-reap25}{n/a}
\defV{r-loss-q-reap50}{+25.0}
\defV{r-gateloss-sym-q-reap50}{+23.0}
\defV{r-qualloss-sym-q-reap50}{+2.0}
\defV{r-gateshare-sym-q-reap50}{92.1}
\defV{r-gateshare-bfirst-q-reap50}{93.7}
\defV{r-gateshare-cfirst-q-reap50}{90.5}
\defV{r-loss-q-ream25}{+29.3}
\defV{r-gateloss-sym-q-ream25}{+26.3}
\defV{r-qualloss-sym-q-ream25}{+3.0}
\defV{r-gateshare-sym-q-ream25}{89.8}
\defV{r-gateshare-bfirst-q-ream25}{92.2}
\defV{r-gateshare-cfirst-q-ream25}{87.4}
\defV{r-loss-q-ream50}{+45.1}
\defV{r-gateloss-sym-q-ream50}{+41.3}
\defV{r-qualloss-sym-q-ream50}{+3.8}
\defV{r-gateshare-sym-q-ream50}{91.7}
\defV{r-gateshare-bfirst-q-ream50}{95.3}
\defV{r-gateshare-cfirst-q-ream50}{88.0}
\defV{r-loss-n-reap25}{\ensuremath{-}0.2}
\defV{r-gateloss-sym-n-reap25}{\ensuremath{-}0.1}
\defV{r-qualloss-sym-n-reap25}{\ensuremath{-}0.1}
\defV{r-gateshare-sym-n-reap25}{n/a}
\defV{r-gateshare-bfirst-n-reap25}{n/a}
\defV{r-gateshare-cfirst-n-reap25}{n/a}
\defV{r-loss-n-reap50}{+9.8}
\defV{r-gateloss-sym-n-reap50}{+9.4}
\defV{r-qualloss-sym-n-reap50}{+0.5}
\defV{r-gateshare-sym-n-reap50}{95.2}
\defV{r-gateshare-bfirst-n-reap50}{98.1}
\defV{r-gateshare-cfirst-n-reap50}{92.3}
\defV{r-loss-n-ream25}{+8.3}
\defV{r-gateloss-sym-n-ream25}{+7.8}
\defV{r-qualloss-sym-n-ream25}{+0.5}
\defV{r-gateshare-sym-n-ream25}{93.7}
\defV{r-gateshare-bfirst-n-ream25}{96.5}
\defV{r-gateshare-cfirst-n-ream25}{90.8}
\defV{r-loss-n-ream50}{+12.4}
\defV{r-gateloss-sym-n-ream50}{+11.6}
\defV{r-qualloss-sym-n-ream50}{+0.8}
\defV{r-gateshare-sym-n-ream50}{93.4}
\defV{r-gateshare-bfirst-n-ream50}{99.6}
\defV{r-gateshare-cfirst-n-ream50}{87.2}
\defV{r-gateshare-sym-min}{89.8}
\defV{r-gateshare-sym-max}{95.2}
\defV{r-gateshare-bfirst-min}{92.2}
\defV{r-gateshare-bfirst-max}{99.6}
\defV{r-gateshare-cfirst-min}{87.2}
\defV{r-gateshare-cfirst-max}{92.3}
\defV{r-gateshare-sym-q-min}{89.8}
\defV{r-gateshare-sym-q-max}{92.1}
\defV{r-gateshare-sym-n-min}{93.4}
\defV{r-gateshare-sym-n-max}{95.2}
\defV{r-brand-exact-q-base}{64.8}
\defV{r-brand-det-q-base}{87.2}
\defV{r-brand-detlo-q-base}{85.8}
\defV{r-brand-dethi-q-base}{88.6}
\defV{r-brand-prec-q-base}{74.3}
\defV{r-brand-preclo-q-base}{72.3}
\defV{r-brand-prechi-q-base}{76.3}
\defV{r-brand-exact-q-reap25}{68.7}
\defV{r-brand-det-q-reap25}{91.0}
\defV{r-brand-detlo-q-reap25}{89.7}
\defV{r-brand-dethi-q-reap25}{92.2}
\defV{r-brand-prec-q-reap25}{75.5}
\defV{r-brand-preclo-q-reap25}{73.5}
\defV{r-brand-prechi-q-reap25}{77.4}
\defV{r-brand-d-q-reap25}{+3.9}
\defV{r-brand-detpart-q-reap25}{+2.8}
\defV{r-brand-precpart-q-reap25}{+1.1}
\defV{r-brand-exact-q-reap50}{69.0}
\defV{r-brand-det-q-reap50}{91.5}
\defV{r-brand-detlo-q-reap50}{90.2}
\defV{r-brand-dethi-q-reap50}{92.7}
\defV{r-brand-prec-q-reap50}{75.5}
\defV{r-brand-preclo-q-reap50}{73.5}
\defV{r-brand-prechi-q-reap50}{77.4}
\defV{r-brand-d-q-reap50}{+4.2}
\defV{r-brand-detpart-q-reap50}{+3.2}
\defV{r-brand-precpart-q-reap50}{+1.0}
\defV{r-brand-exact-q-ream25}{64.5}
\defV{r-brand-det-q-ream25}{88.2}
\defV{r-brand-detlo-q-ream25}{86.8}
\defV{r-brand-dethi-q-ream25}{89.6}
\defV{r-brand-prec-q-ream25}{73.1}
\defV{r-brand-preclo-q-ream25}{71.1}
\defV{r-brand-prechi-q-ream25}{75.2}
\defV{r-brand-d-q-ream25}{\ensuremath{-}0.3}
\defV{r-brand-detpart-q-ream25}{+0.7}
\defV{r-brand-precpart-q-ream25}{\ensuremath{-}1.0}
\defV{r-brand-exact-q-ream50}{70.7}
\defV{r-brand-det-q-ream50}{95.2}
\defV{r-brand-detlo-q-ream50}{94.2}
\defV{r-brand-dethi-q-ream50}{96.0}
\defV{r-brand-prec-q-ream50}{74.3}
\defV{r-brand-preclo-q-ream50}{72.2}
\defV{r-brand-prechi-q-ream50}{76.2}
\defV{r-brand-d-q-ream50}{+5.9}
\defV{r-brand-detpart-q-ream50}{+5.9}
\defV{r-brand-precpart-q-ream50}{\ensuremath{-}0.1}
\defV{r-brand-exact-n-base}{49.8}
\defV{r-brand-det-n-base}{78.5}
\defV{r-brand-detlo-n-base}{76.6}
\defV{r-brand-dethi-n-base}{80.2}
\defV{r-brand-prec-n-base}{63.4}
\defV{r-brand-preclo-n-base}{61.1}
\defV{r-brand-prechi-n-base}{65.8}
\defV{r-brand-exact-n-reap25}{45.8}
\defV{r-brand-det-n-reap25}{72.0}
\defV{r-brand-detlo-n-reap25}{70.0}
\defV{r-brand-dethi-n-reap25}{74.1}
\defV{r-brand-prec-n-reap25}{63.5}
\defV{r-brand-preclo-n-reap25}{61.0}
\defV{r-brand-prechi-n-reap25}{66.0}
\defV{r-brand-d-n-reap25}{\ensuremath{-}4.0}
\defV{r-brand-detpart-n-reap25}{\ensuremath{-}4.1}
\defV{r-brand-precpart-n-reap25}{+0.1}
\defV{r-brand-exact-n-reap50}{28.6}
\defV{r-brand-det-n-reap50}{54.5}
\defV{r-brand-detlo-n-reap50}{52.3}
\defV{r-brand-dethi-n-reap50}{56.7}
\defV{r-brand-prec-n-reap50}{52.6}
\defV{r-brand-preclo-n-reap50}{49.7}
\defV{r-brand-prechi-n-reap50}{55.5}
\defV{r-brand-d-n-reap50}{\ensuremath{-}21.1}
\defV{r-brand-detpart-n-reap50}{\ensuremath{-}13.9}
\defV{r-brand-precpart-n-reap50}{\ensuremath{-}7.2}
\defV{r-brand-exact-n-ream25}{39.8}
\defV{r-brand-det-n-ream25}{61.6}
\defV{r-brand-detlo-n-ream25}{59.5}
\defV{r-brand-dethi-n-ream25}{63.7}
\defV{r-brand-prec-n-ream25}{64.6}
\defV{r-brand-preclo-n-ream25}{61.9}
\defV{r-brand-prechi-n-ream25}{67.3}
\defV{r-brand-d-n-ream25}{\ensuremath{-}9.9}
\defV{r-brand-detpart-n-ream25}{\ensuremath{-}10.8}
\defV{r-brand-precpart-n-ream25}{+0.8}
\defV{r-brand-exact-n-ream50}{34.1}
\defV{r-brand-det-n-ream50}{53.5}
\defV{r-brand-detlo-n-ream50}{51.3}
\defV{r-brand-dethi-n-ream50}{55.7}
\defV{r-brand-prec-n-ream50}{63.7}
\defV{r-brand-preclo-n-ream50}{60.8}
\defV{r-brand-prechi-n-ream50}{66.6}
\defV{r-brand-d-n-ream50}{\ensuremath{-}15.6}
\defV{r-brand-detpart-n-ream50}{\ensuremath{-}15.9}
\defV{r-brand-precpart-n-ream50}{+0.2}
\defV{r-brand-hdet-q25}{+2.8}
\defV{r-brand-hprec-q25}{+2.4}
\defV{r-brand-hdet-q50}{\ensuremath{-}3.7}
\defV{r-brand-hprec-q50}{+1.2}
\defV{r-brand-hdet-n25}{+10.4}
\defV{r-brand-hprec-n25}{\ensuremath{-}1.1}
\defV{r-brand-hdet-n50}{+1.0}
\defV{r-brand-hprec-n50}{\ensuremath{-}11.2}
\defV{r-langen-q-reap25-usage}{\ensuremath{-}0.1}
\defV{r-langnonen-q-reap25-usage}{\ensuremath{-}0.4}
\defV{r-langgap-q-reap25-usage}{\ensuremath{-}0.3}
\defV{r-langgaplo-q-reap25-usage}{\ensuremath{-}1.0}
\defV{r-langgaphi-q-reap25-usage}{+0.3}
\defV{r-langgapp-q-reap25-usage}{p = 1}
\defV{r-langen-q-reap50-usage}{\ensuremath{-}0.5}
\defV{r-langnonen-q-reap50-usage}{\ensuremath{-}1.5}
\defV{r-langgap-q-reap50-usage}{\ensuremath{-}1.0}
\defV{r-langgaplo-q-reap50-usage}{\ensuremath{-}2.1}
\defV{r-langgaphi-q-reap50-usage}{+0.1}
\defV{r-langgapp-q-reap50-usage}{p = 1}
\defV{r-langen-q-ream25-usage}{0.0}
\defV{r-langnonen-q-ream25-usage}{\ensuremath{-}1.1}
\defV{r-langgap-q-ream25-usage}{\ensuremath{-}1.1}
\defV{r-langgaplo-q-ream25-usage}{\ensuremath{-}2.1}
\defV{r-langgaphi-q-ream25-usage}{\ensuremath{-}0.1}
\defV{r-langgapp-q-ream25-usage}{p = 0.78}
\defV{r-langen-q-ream50-usage}{\ensuremath{-}0.7}
\defV{r-langnonen-q-ream50-usage}{\ensuremath{-}2.7}
\defV{r-langgap-q-ream50-usage}{\ensuremath{-}2.1}
\defV{r-langgaplo-q-ream50-usage}{\ensuremath{-}3.5}
\defV{r-langgaphi-q-ream50-usage}{\ensuremath{-}0.7}
\defV{r-langgapp-q-ream50-usage}{p = 0.068}
\defV{r-langen-n-reap25-usage}{+0.3}
\defV{r-langnonen-n-reap25-usage}{+0.8}
\defV{r-langgap-n-reap25-usage}{+0.6}
\defV{r-langgaplo-n-reap25-usage}{\ensuremath{-}0.5}
\defV{r-langgaphi-n-reap25-usage}{+1.6}
\defV{r-langgapp-n-reap25-usage}{p = 1}
\defV{r-langen-n-reap50-usage}{\ensuremath{-}1.0}
\defV{r-langnonen-n-reap50-usage}{\ensuremath{-}1.6}
\defV{r-langgap-n-reap50-usage}{\ensuremath{-}0.5}
\defV{r-langgaplo-n-reap50-usage}{\ensuremath{-}2.2}
\defV{r-langgaphi-n-reap50-usage}{+1.1}
\defV{r-langgapp-n-reap50-usage}{p = 1}
\defV{r-langen-n-ream25-usage}{\ensuremath{-}0.7}
\defV{r-langnonen-n-ream25-usage}{\ensuremath{-}0.2}
\defV{r-langgap-n-ream25-usage}{+0.5}
\defV{r-langgaplo-n-ream25-usage}{\ensuremath{-}1.0}
\defV{r-langgaphi-n-ream25-usage}{+1.9}
\defV{r-langgapp-n-ream25-usage}{p = 1}
\defV{r-langen-n-ream50-usage}{\ensuremath{-}5.7}
\defV{r-langnonen-n-ream50-usage}{\ensuremath{-}3.0}
\defV{r-langgap-n-ream50-usage}{+2.7}
\defV{r-langgaplo-n-ream50-usage}{+0.8}
\defV{r-langgaphi-n-ream50-usage}{+4.7}
\defV{r-langgapp-n-ream50-usage}{p = 0.2}
\defV{r-langen-q-reap25-brand}{+4.2}
\defV{r-langnonen-q-reap25-brand}{+3.8}
\defV{r-langgap-q-reap25-brand}{\ensuremath{-}0.3}
\defV{r-langgaplo-q-reap25-brand}{\ensuremath{-}3.3}
\defV{r-langgaphi-q-reap25-brand}{+2.5}
\defV{r-langgapp-q-reap25-brand}{p = 1}
\defV{r-langen-q-reap50-brand}{+5.0}
\defV{r-langnonen-q-reap50-brand}{+3.9}
\defV{r-langgap-q-reap50-brand}{\ensuremath{-}1.0}
\defV{r-langgaplo-q-reap50-brand}{\ensuremath{-}5.1}
\defV{r-langgaphi-q-reap50-brand}{+3.0}
\defV{r-langgapp-q-reap50-brand}{p = 1}
\defV{r-langen-q-ream25-brand}{+0.4}
\defV{r-langnonen-q-ream25-brand}{\ensuremath{-}0.5}
\defV{r-langgap-q-ream25-brand}{\ensuremath{-}0.9}
\defV{r-langgaplo-q-ream25-brand}{\ensuremath{-}4.0}
\defV{r-langgaphi-q-ream25-brand}{+2.1}
\defV{r-langgapp-q-ream25-brand}{p = 1}
\defV{r-langen-q-ream50-brand}{+8.3}
\defV{r-langnonen-q-ream50-brand}{+5.1}
\defV{r-langgap-q-ream50-brand}{\ensuremath{-}3.2}
\defV{r-langgaplo-q-ream50-brand}{\ensuremath{-}7.3}
\defV{r-langgaphi-q-ream50-brand}{+0.8}
\defV{r-langgapp-q-ream50-brand}{p = 1}
\defV{r-langen-n-reap25-brand}{\ensuremath{-}0.4}
\defV{r-langnonen-n-reap25-brand}{\ensuremath{-}5.1}
\defV{r-langgap-n-reap25-brand}{\ensuremath{-}4.7}
\defV{r-langgaplo-n-reap25-brand}{\ensuremath{-}8.1}
\defV{r-langgaphi-n-reap25-brand}{\ensuremath{-}1.3}
\defV{r-langgapp-n-reap25-brand}{p = 0.24}
\defV{r-langen-n-reap50-brand}{\ensuremath{-}17.7}
\defV{r-langnonen-n-reap50-brand}{\ensuremath{-}22.2}
\defV{r-langgap-n-reap50-brand}{\ensuremath{-}4.5}
\defV{r-langgaplo-n-reap50-brand}{\ensuremath{-}9.1}
\defV{r-langgaphi-n-reap50-brand}{\ensuremath{-}0.1}
\defV{r-langgapp-n-reap50-brand}{p = 0.94}
\defV{r-langen-n-ream25-brand}{\ensuremath{-}7.9}
\defV{r-langnonen-n-ream25-brand}{\ensuremath{-}10.6}
\defV{r-langgap-n-ream25-brand}{\ensuremath{-}2.7}
\defV{r-langgaplo-n-ream25-brand}{\ensuremath{-}7.0}
\defV{r-langgaphi-n-ream25-brand}{+1.7}
\defV{r-langgapp-n-ream25-brand}{p = 1}
\defV{r-langen-n-ream50-brand}{\ensuremath{-}12.5}
\defV{r-langnonen-n-ream50-brand}{\ensuremath{-}16.7}
\defV{r-langgap-n-ream50-brand}{\ensuremath{-}4.2}
\defV{r-langgaplo-n-ream50-brand}{\ensuremath{-}8.8}
\defV{r-langgaphi-n-ream50-brand}{+0.5}
\defV{r-langgapp-n-ream50-brand}{p = 1}
\defV{r-langen-q-reap25-categ}{+2.2}
\defV{r-langnonen-q-reap25-categ}{\ensuremath{-}1.4}
\defV{r-langgap-q-reap25-categ}{\ensuremath{-}3.7}
\defV{r-langgaplo-q-reap25-categ}{\ensuremath{-}7.2}
\defV{r-langgaphi-q-reap25-categ}{\ensuremath{-}0.2}
\defV{r-langgapp-q-reap25-categ}{p = 0.79}
\defV{r-langen-q-reap50-categ}{\ensuremath{-}5.5}
\defV{r-langnonen-q-reap50-categ}{\ensuremath{-}6.9}
\defV{r-langgap-q-reap50-categ}{\ensuremath{-}1.4}
\defV{r-langgaplo-q-reap50-categ}{\ensuremath{-}6.5}
\defV{r-langgaphi-q-reap50-categ}{+3.9}
\defV{r-langgapp-q-reap50-categ}{p = 1}
\defV{r-langen-q-ream25-categ}{+2.2}
\defV{r-langnonen-q-ream25-categ}{\ensuremath{-}6.1}
\defV{r-langgap-q-ream25-categ}{\ensuremath{-}8.3}
\defV{r-langgaplo-q-ream25-categ}{\ensuremath{-}12.1}
\defV{r-langgaphi-q-ream25-categ}{\ensuremath{-}4.5}
\defV{r-langgapp-q-ream25-categ}{p = 0.008}
\defV{r-langen-q-ream50-categ}{\ensuremath{-}4.4}
\defV{r-langnonen-q-ream50-categ}{\ensuremath{-}11.3}
\defV{r-langgap-q-ream50-categ}{\ensuremath{-}6.9}
\defV{r-langgaplo-q-ream50-categ}{\ensuremath{-}12.0}
\defV{r-langgaphi-q-ream50-categ}{\ensuremath{-}1.8}
\defV{r-langgapp-q-ream50-categ}{p = 0.23}
\defV{r-langen-n-reap25-categ}{\ensuremath{-}0.7}
\defV{r-langnonen-n-reap25-categ}{\ensuremath{-}3.4}
\defV{r-langgap-n-reap25-categ}{\ensuremath{-}2.6}
\defV{r-langgaplo-n-reap25-categ}{\ensuremath{-}6.8}
\defV{r-langgaphi-n-reap25-categ}{+1.5}
\defV{r-langgapp-n-reap25-categ}{p = 1}
\defV{r-langen-n-reap50-categ}{\ensuremath{-}9.2}
\defV{r-langnonen-n-reap50-categ}{\ensuremath{-}11.6}
\defV{r-langgap-n-reap50-categ}{\ensuremath{-}2.4}
\defV{r-langgaplo-n-reap50-categ}{\ensuremath{-}7.9}
\defV{r-langgaphi-n-reap50-categ}{+3.2}
\defV{r-langgapp-n-reap50-categ}{p = 1}
\defV{r-langen-n-ream25-categ}{\ensuremath{-}13.6}
\defV{r-langnonen-n-ream25-categ}{\ensuremath{-}7.5}
\defV{r-langgap-n-ream25-categ}{+6.1}
\defV{r-langgaplo-n-ream25-categ}{+0.4}
\defV{r-langgaphi-n-ream25-categ}{+11.9}
\defV{r-langgapp-n-ream25-categ}{p = 0.78}
\defV{r-langen-n-ream50-categ}{\ensuremath{-}25.0}
\defV{r-langnonen-n-ream50-categ}{\ensuremath{-}17.2}
\defV{r-langgap-n-ream50-categ}{+7.8}
\defV{r-langgaplo-n-ream50-categ}{+1.5}
\defV{r-langgaphi-n-ream50-categ}{+14.3}
\defV{r-langgapp-n-ream50-categ}{p = 0.34}
\defV{r-langen-q-reap25-domains}{+1.2}
\defV{r-langnonen-q-reap25-domains}{\ensuremath{-}16.0}
\defV{r-langgap-q-reap25-domains}{\ensuremath{-}17.2}
\defV{r-langgaplo-q-reap25-domains}{\ensuremath{-}19.5}
\defV{r-langgaphi-q-reap25-domains}{\ensuremath{-}14.9}
\defV{r-langgapp-q-reap25-domains}{p = 0.008}
\defV{r-langen-q-reap50-domains}{\ensuremath{-}1.5}
\defV{r-langnonen-q-reap50-domains}{\ensuremath{-}26.5}
\defV{r-langgap-q-reap50-domains}{\ensuremath{-}25.1}
\defV{r-langgaplo-q-reap50-domains}{\ensuremath{-}28.3}
\defV{r-langgaphi-q-reap50-domains}{\ensuremath{-}21.9}
\defV{r-langgapp-q-reap50-domains}{p = 0.008}
\defV{r-langen-q-ream25-domains}{\ensuremath{-}0.6}
\defV{r-langnonen-q-ream25-domains}{\ensuremath{-}37.5}
\defV{r-langgap-q-ream25-domains}{\ensuremath{-}36.9}
\defV{r-langgaplo-q-ream25-domains}{\ensuremath{-}39.9}
\defV{r-langgaphi-q-ream25-domains}{\ensuremath{-}33.8}
\defV{r-langgapp-q-ream25-domains}{p = 0.008}
\defV{r-langen-q-ream50-domains}{\ensuremath{-}2.6}
\defV{r-langnonen-q-ream50-domains}{\ensuremath{-}35.8}
\defV{r-langgap-q-ream50-domains}{\ensuremath{-}33.1}
\defV{r-langgaplo-q-ream50-domains}{\ensuremath{-}36.6}
\defV{r-langgaphi-q-ream50-domains}{\ensuremath{-}29.6}
\defV{r-langgapp-q-ream50-domains}{p = 0.008}
\defV{r-langen-n-reap25-domains}{+0.6}
\defV{r-langnonen-n-reap25-domains}{\ensuremath{-}7.2}
\defV{r-langgap-n-reap25-domains}{\ensuremath{-}7.8}
\defV{r-langgaplo-n-reap25-domains}{\ensuremath{-}12.4}
\defV{r-langgaphi-n-reap25-domains}{\ensuremath{-}3.3}
\defV{r-langgapp-n-reap25-domains}{p = 0.008}
\defV{r-langen-n-reap50-domains}{+3.5}
\defV{r-langnonen-n-reap50-domains}{\ensuremath{-}27.3}
\defV{r-langgap-n-reap50-domains}{\ensuremath{-}30.8}
\defV{r-langgaplo-n-reap50-domains}{\ensuremath{-}35.0}
\defV{r-langgaphi-n-reap50-domains}{\ensuremath{-}26.5}
\defV{r-langgapp-n-reap50-domains}{p = 0.008}
\defV{r-langen-n-ream25-domains}{\ensuremath{-}6.4}
\defV{r-langnonen-n-ream25-domains}{\ensuremath{-}21.4}
\defV{r-langgap-n-ream25-domains}{\ensuremath{-}14.9}
\defV{r-langgaplo-n-ream25-domains}{\ensuremath{-}20.2}
\defV{r-langgaphi-n-ream25-domains}{\ensuremath{-}9.6}
\defV{r-langgapp-n-ream25-domains}{p = 0.008}
\defV{r-langen-n-ream50-domains}{\ensuremath{-}13.7}
\defV{r-langnonen-n-ream50-domains}{\ensuremath{-}29.8}
\defV{r-langgap-n-ream50-domains}{\ensuremath{-}16.1}
\defV{r-langgaplo-n-ream50-domains}{\ensuremath{-}21.6}
\defV{r-langgaphi-n-ream50-domains}{\ensuremath{-}10.4}
\defV{r-langgapp-n-ream50-domains}{p = 0.008}
\defV{r-langen-q-reap25-reach}{+1.5}
\defV{r-langnonen-q-reap25-reach}{+0.8}
\defV{r-langgap-q-reap25-reach}{\ensuremath{-}0.7}
\defV{r-langgaplo-q-reap25-reach}{\ensuremath{-}4.6}
\defV{r-langgaphi-q-reap25-reach}{+3.1}
\defV{r-langgapp-q-reap25-reach}{p = 1}
\defV{r-langen-q-reap50-reach}{\ensuremath{-}21.8}
\defV{r-langnonen-q-reap50-reach}{\ensuremath{-}25.6}
\defV{r-langgap-q-reap50-reach}{\ensuremath{-}3.9}
\defV{r-langgaplo-q-reap50-reach}{\ensuremath{-}9.1}
\defV{r-langgaphi-q-reap50-reach}{+1.3}
\defV{r-langgapp-q-reap50-reach}{p = 1}
\defV{r-langen-q-ream25-reach}{\ensuremath{-}26.7}
\defV{r-langnonen-q-ream25-reach}{\ensuremath{-}29.9}
\defV{r-langgap-q-ream25-reach}{\ensuremath{-}3.2}
\defV{r-langgaplo-q-ream25-reach}{\ensuremath{-}8.6}
\defV{r-langgaphi-q-ream25-reach}{+2.2}
\defV{r-langgapp-q-ream25-reach}{p = 1}
\defV{r-langen-q-ream50-reach}{\ensuremath{-}40.2}
\defV{r-langnonen-q-ream50-reach}{\ensuremath{-}46.1}
\defV{r-langgap-q-ream50-reach}{\ensuremath{-}6.0}
\defV{r-langgaplo-q-ream50-reach}{\ensuremath{-}11.2}
\defV{r-langgaphi-q-ream50-reach}{\ensuremath{-}0.9}
\defV{r-langgapp-q-ream50-reach}{p = 0.54}
\defV{r-langen-n-reap25-reach}{+3.1}
\defV{r-langnonen-n-reap25-reach}{\ensuremath{-}0.4}
\defV{r-langgap-n-reap25-reach}{\ensuremath{-}3.4}
\defV{r-langgaplo-n-reap25-reach}{\ensuremath{-}8.2}
\defV{r-langgaphi-n-reap25-reach}{+1.2}
\defV{r-langgapp-n-reap25-reach}{p = 1}
\defV{r-langen-n-reap50-reach}{\ensuremath{-}13.2}
\defV{r-langnonen-n-reap50-reach}{\ensuremath{-}9.1}
\defV{r-langgap-n-reap50-reach}{+4.1}
\defV{r-langgaplo-n-reap50-reach}{+0.4}
\defV{r-langgaphi-n-reap50-reach}{+8.0}
\defV{r-langgapp-n-reap50-reach}{p = 0.77}
\defV{r-langen-n-ream25-reach}{\ensuremath{-}11.5}
\defV{r-langnonen-n-ream25-reach}{\ensuremath{-}7.6}
\defV{r-langgap-n-ream25-reach}{+3.9}
\defV{r-langgaplo-n-ream25-reach}{0.0}
\defV{r-langgaphi-n-ream25-reach}{+7.9}
\defV{r-langgapp-n-ream25-reach}{p = 0.94}
\defV{r-langen-n-ream50-reach}{\ensuremath{-}16.3}
\defV{r-langnonen-n-ream50-reach}{\ensuremath{-}11.6}
\defV{r-langgap-n-ream50-reach}{+4.7}
\defV{r-langgaplo-n-ream50-reach}{+1.2}
\defV{r-langgaphi-n-ream50-reach}{+8.3}
\defV{r-langgapp-n-ream50-reach}{p = 0.22}
\defV{r-langgap-cells}{40}
\defV{r-langgap-sig}{9}
\defV{r-langgap-sig-neg}{9}
\defV{r-langgap-sig-pos}{0}
\defV{r-langgap-sig-nodomains}{1}
\defV{r-langgap-sig-neg-nodomains}{1}
\defV{r-langgap-sig-usage}{0}
\defV{r-langgap-mean-usage}{\ensuremath{-}0.2}
\defV{r-langen-mean-usage}{\ensuremath{-}1.0}
\defV{r-langnonen-mean-usage}{\ensuremath{-}1.2}
\defV{r-langgap-mean-usage-q}{\ensuremath{-}1.1}
\defV{r-langgap-mean-usage-n}{+0.8}
\defV{r-langgap-sig-brand}{0}
\defV{r-langgap-mean-brand}{\ensuremath{-}2.7}
\defV{r-langen-mean-brand}{\ensuremath{-}2.6}
\defV{r-langnonen-mean-brand}{\ensuremath{-}5.3}
\defV{r-langgap-mean-brand-q}{\ensuremath{-}1.4}
\defV{r-langgap-mean-brand-n}{\ensuremath{-}4.0}
\defV{r-langgap-sig-categ}{1}
\defV{r-langgap-mean-categ}{\ensuremath{-}1.4}
\defV{r-langen-mean-categ}{\ensuremath{-}6.8}
\defV{r-langnonen-mean-categ}{\ensuremath{-}8.2}
\defV{r-langgap-mean-categ-q}{\ensuremath{-}5.1}
\defV{r-langgap-mean-categ-n}{+2.2}
\defV{r-langgap-sig-domains}{8}
\defV{r-langgap-mean-domains}{\ensuremath{-}22.7}
\defV{r-langen-mean-domains}{\ensuremath{-}2.4}
\defV{r-langnonen-mean-domains}{\ensuremath{-}25.2}
\defV{r-langgap-mean-domains-q}{\ensuremath{-}28.1}
\defV{r-langgap-mean-domains-n}{\ensuremath{-}17.4}
\defV{r-langgap-sig-reach}{0}
\defV{r-langgap-mean-reach}{\ensuremath{-}0.5}
\defV{r-langen-mean-reach}{\ensuremath{-}15.6}
\defV{r-langnonen-mean-reach}{\ensuremath{-}16.2}
\defV{r-langgap-mean-reach-q}{\ensuremath{-}3.4}
\defV{r-langgap-mean-reach-n}{+2.3}
\defV{r-loops-n-base-domains}{0}
\defV{r-div-n-base-domains}{0.494}
\defV{r-loops-n-ream25-domains}{3}
\defV{r-div-n-ream25-domains}{0.410}
\defV{r-loops-n-ream50-domains}{20}
\defV{r-div-n-ream50-domains}{0.150}
\defV{r-loops-n-reap25-domains}{1}
\defV{r-div-n-reap25-domains}{0.411}
\defV{r-loops-n-reap50-domains}{16}
\defV{r-div-n-reap50-domains}{0.068}
\defV{r-loops-q-base-domains}{0}
\defV{r-div-q-base-domains}{0.378}
\defV{r-loops-q-ream25-domains}{0}
\defV{r-div-q-ream25-domains}{0.372}
\defV{r-loops-q-ream50-domains}{4}
\defV{r-div-q-ream50-domains}{0.130}
\defV{r-loops-q-reap25-domains}{0}
\defV{r-div-q-reap25-domains}{0.359}
\defV{r-loops-q-reap50-domains}{2}
\defV{r-div-q-reap50-domains}{0.172}
\defV{r-loops-n-base-reach}{39}
\defV{r-div-n-base-reach}{0.140}
\defV{r-loops-n-ream25-reach}{172}
\defV{r-div-n-ream25-reach}{0.057}
\defV{r-loops-n-ream50-reach}{312}
\defV{r-div-n-ream50-reach}{0.036}
\defV{r-loops-n-reap25-reach}{33}
\defV{r-div-n-reap25-reach}{0.205}
\defV{r-loops-n-reap50-reach}{237}
\defV{r-div-n-reap50-reach}{0.086}
\defV{r-loops-q-base-reach}{6}
\defV{r-div-q-base-reach}{0.690}
\defV{r-loops-q-ream25-reach}{61}
\defV{r-div-q-ream25-reach}{0.222}
\defV{r-loops-q-ream50-reach}{215}
\defV{r-div-q-ream50-reach}{0.121}
\defV{r-loops-q-reap25-reach}{14}
\defV{r-div-q-reap25-reach}{0.590}
\defV{r-loops-q-reap50-reach}{102}
\defV{r-div-q-reap50-reach}{0.263}
\defV{r-tau-xppl-q-usage}{+1.00}
\defV{r-tau-xppl-n-usage}{+0.80}
\defV{r-tau-xppl-q-brand}{\ensuremath{-}0.60}
\defV{r-tau-xppl-n-brand}{+0.80}
\defV{r-tau-xppl-q-categ}{+1.00}
\defV{r-tau-xppl-n-categ}{+1.00}
\defV{r-tau-xppl-q-domains}{+0.60}
\defV{r-tau-xppl-n-domains}{+0.80}
\defV{r-tau-xppl-q-reach}{+0.80}
\defV{r-tau-xppl-n-reach}{+0.80}
\defV{r-tau-xppl}{+0.70}
\defV{r-taupos-xppl}{9}
\defV{r-taucells-xppl}{10}
\defV{r-order-xppl-cells}{20}
\defV{r-order-xppl-reap}{20}
\defV{r-order-xppl-both}{18}
\defV{r-order-xppl-agree}{18}
\defV{r-tau-jsd-n-usage}{+1.00}
\defV{r-tau-jsd-q-usage}{0.00}
\defV{r-tau-jsd-n-brand}{+0.67}
\defV{r-tau-jsd-q-brand}{\ensuremath{-}0.67}
\defV{r-tau-jsd-n-categ}{+1.00}
\defV{r-tau-jsd-q-categ}{+1.00}
\defV{r-tau-jsd-n-domains}{0.00}
\defV{r-tau-jsd-q-domains}{+0.33}
\defV{r-tau-jsd-n-reach}{+1.00}
\defV{r-tau-jsd-q-reach}{+0.67}
\defV{r-tau-jsd}{+0.50}
\defV{r-taupos-jsd}{7}
\defV{r-taucells-jsd}{10}
\defV{r-order-jsd-cells}{20}
\defV{r-order-jsd-reap}{20}
\defV{r-order-jsd-both}{18}
\defV{r-order-jsd-agree}{18}
\defV{r-tau-erank-q-usage}{+1.00}
\defV{r-tau-erank-q-brand}{\ensuremath{-}0.60}
\defV{r-tau-erank-q-categ}{+1.00}
\defV{r-tau-erank-q-domains}{+0.60}
\defV{r-tau-erank-q-reach}{+0.60}
\defV{r-tau-erank-n-usage}{0.00}
\defV{r-tau-erank-n-brand}{+0.80}
\defV{r-tau-erank-n-categ}{+1.00}
\defV{r-tau-erank-n-domains}{0.00}
\defV{r-tau-erank-n-reach}{0.00}
\defV{r-tau-erank}{+0.44}
\defV{r-taupos-erank}{6}
\defV{r-taucells-erank}{10}
\defV{r-order-erank-cells}{20}
\defV{r-order-erank-reap}{15}
\defV{r-order-erank-both}{13}
\defV{r-order-erank-agree}{13}
\defV{r-tau-div-n-domains}{\ensuremath{-}0.20}
\defV{r-tau-div-q-domains}{+0.20}
\defV{r-tau-div-n-reach}{+0.40}
\defV{r-tau-div-q-reach}{+0.20}
\defV{r-tau-div}{+0.15}
\defV{r-taupos-div}{3}
\defV{r-taucells-div}{4}
\defV{r-order-div-cells}{8}
\defV{r-order-div-reap}{1}
\defV{r-order-div-both}{1}
\defV{r-order-div-agree}{1}
\defV{r-tau-divloops-n-domains}{+0.80}
\defV{r-tau-divloops-q-domains}{+0.40}
\defV{r-tau-divloops-n-reach}{+0.80}
\defV{r-tau-divloops-q-reach}{+0.80}
\defV{r-tau-divloops}{+0.70}
\defV{r-taupos-divloops}{4}
\defV{r-taucells-divloops}{4}
\defV{r-order-divloops-cells}{8}
\defV{r-order-divloops-reap}{6}
\defV{r-order-divloops-both}{6}
\defV{r-order-divloops-agree}{6}
\defV{r-tau-loop-n-domains}{+1.00}
\defV{r-tau-loop-q-domains}{+0.36}
\defV{r-tau-loop-n-reach}{+1.00}
\defV{r-tau-loop-q-reach}{+0.60}
\defV{r-tau-loop}{+0.74}
\defV{r-taupos-loop}{4}
\defV{r-taucells-loop}{4}
\defV{r-order-loop-cells}{8}
\defV{r-order-loop-reap}{7}
\defV{r-order-loop-both}{7}
\defV{r-order-loop-agree}{7}
\defV{r-tau-ownppl-q-usage}{+0.80}
\defV{r-tau-ownppl-n-usage}{+0.80}
\defV{r-tau-ownppl-q-brand}{+1.00}
\defV{r-tau-ownppl-n-brand}{+1.00}
\defV{r-tau-ownppl-q-categ}{+0.60}
\defV{r-tau-ownppl-n-categ}{+1.00}
\defV{r-tau-ownppl-q-domains}{+0.60}
\defV{r-tau-ownppl-n-domains}{+0.80}
\defV{r-tau-ownppl-q-reach}{+0.80}
\defV{r-tau-ownppl-n-reach}{+0.60}
\defV{r-tau-ownppl}{+0.80}
\defV{r-taupos-ownppl}{10}
\defV{r-taucells-ownppl}{10}
\defV{r-order-ownppl-cells}{20}
\defV{r-order-ownppl-reap}{18}
\defV{r-order-ownppl-both}{18}
\defV{r-order-ownppl-agree}{20}
\defV{r-order-quality-reap-all}{18}
\defV{r-order-quality-reap-gen}{8}
\defV{r-loops-ratio-q25-domains}{n/a}
\defV{r-loops-ratio-q25-reach}{4.4}
\defV{r-loops-ratio-q50-domains}{2.0}
\defV{r-loops-ratio-q50-reach}{2.1}
\defV{r-loops-ratio-n25-domains}{3.0}
\defV{r-loops-ratio-n25-reach}{5.2}
\defV{r-loops-ratio-n50-domains}{1.2}
\defV{r-loops-ratio-n50-reach}{1.3}
\defV{r-rm-answer-q25-usage}{19.7}
\defV{r-rm-prompt-q25-usage}{9.0}
\defV{r-rm-ratio-q25-usage}{2.18}
\defV{r-rm-answer-q25-brand}{7.9}
\defV{r-rm-prompt-q25-brand}{7.1}
\defV{r-rm-ratio-q25-brand}{1.11}
\defV{r-rm-answer-q25-categ}{7.3}
\defV{r-rm-prompt-q25-categ}{7.2}
\defV{r-rm-ratio-q25-categ}{1.01}
\defV{r-rm-answer-q25-domains}{6.5}
\defV{r-rm-prompt-q25-domains}{6.5}
\defV{r-rm-ratio-q25-domains}{1.00}
\defV{r-rm-answer-q25-reach}{17.6}
\defV{r-rm-prompt-q25-reach}{8.5}
\defV{r-rm-ratio-q25-reach}{2.07}
\defV{r-rm-ratio-q25-pooled}{2.13}
\defV{r-rm-ratio-q25-taskmean}{1.54}
\defV{r-rm-pooled-answer-q25}{17.4}
\defV{r-rm-taskmean-answer-q25}{11.8}
\defV{r-rm-pooled-prompt-q25}{8.2}
\defV{r-rm-taskmean-prompt-q25}{7.7}
\defV{r-rm-ansgt-q25}{3}
\defV{r-rm-answer-q50-usage}{38.8}
\defV{r-rm-prompt-q50-usage}{28.6}
\defV{r-rm-ratio-q50-usage}{1.36}
\defV{r-rm-answer-q50-brand}{24.8}
\defV{r-rm-prompt-q50-brand}{24.7}
\defV{r-rm-ratio-q50-brand}{1.01}
\defV{r-rm-answer-q50-categ}{24.3}
\defV{r-rm-prompt-q50-categ}{26.3}
\defV{r-rm-ratio-q50-categ}{0.92}
\defV{r-rm-answer-q50-domains}{26.0}
\defV{r-rm-prompt-q50-domains}{23.8}
\defV{r-rm-ratio-q50-domains}{1.09}
\defV{r-rm-answer-q50-reach}{35.4}
\defV{r-rm-prompt-q50-reach}{24.8}
\defV{r-rm-ratio-q50-reach}{1.43}
\defV{r-rm-ratio-q50-pooled}{1.41}
\defV{r-rm-ratio-q50-taskmean}{1.16}
\defV{r-rm-pooled-answer-q50}{35.3}
\defV{r-rm-taskmean-answer-q50}{29.9}
\defV{r-rm-pooled-prompt-q50}{25.0}
\defV{r-rm-taskmean-prompt-q50}{25.6}
\defV{r-rm-ansgt-q50}{3}
\defV{r-rm-answer-n25-usage}{18.9}
\defV{r-rm-prompt-n25-usage}{17.6}
\defV{r-rm-ratio-n25-usage}{1.07}
\defV{r-rm-answer-n25-brand}{16.1}
\defV{r-rm-prompt-n25-brand}{17.1}
\defV{r-rm-ratio-n25-brand}{0.94}
\defV{r-rm-answer-n25-categ}{16.8}
\defV{r-rm-prompt-n25-categ}{18.2}
\defV{r-rm-ratio-n25-categ}{0.92}
\defV{r-rm-answer-n25-domains}{16.8}
\defV{r-rm-prompt-n25-domains}{13.1}
\defV{r-rm-ratio-n25-domains}{1.28}
\defV{r-rm-answer-n25-reach}{21.8}
\defV{r-rm-prompt-n25-reach}{17.3}
\defV{r-rm-ratio-n25-reach}{1.26}
\defV{r-rm-ratio-n25-pooled}{1.26}
\defV{r-rm-ratio-n25-taskmean}{1.08}
\defV{r-rm-pooled-answer-n25}{21.7}
\defV{r-rm-taskmean-answer-n25}{18.1}
\defV{r-rm-pooled-prompt-n25}{17.3}
\defV{r-rm-taskmean-prompt-n25}{16.7}
\defV{r-rm-ansgt-n25}{3}
\defV{r-rm-answer-n50-usage}{39.2}
\defV{r-rm-prompt-n50-usage}{39.5}
\defV{r-rm-ratio-n50-usage}{0.99}
\defV{r-rm-answer-n50-brand}{36.1}
\defV{r-rm-prompt-n50-brand}{38.4}
\defV{r-rm-ratio-n50-brand}{0.94}
\defV{r-rm-answer-n50-categ}{38.1}
\defV{r-rm-prompt-n50-categ}{42.0}
\defV{r-rm-ratio-n50-categ}{0.91}
\defV{r-rm-answer-n50-domains}{39.6}
\defV{r-rm-prompt-n50-domains}{31.0}
\defV{r-rm-ratio-n50-domains}{1.28}
\defV{r-rm-answer-n50-reach}{44.7}
\defV{r-rm-prompt-n50-reach}{39.3}
\defV{r-rm-ratio-n50-reach}{1.14}
\defV{r-rm-ratio-n50-pooled}{1.14}
\defV{r-rm-ratio-n50-taskmean}{1.04}
\defV{r-rm-pooled-answer-n50}{44.5}
\defV{r-rm-taskmean-answer-n50}{39.6}
\defV{r-rm-pooled-prompt-n50}{39.2}
\defV{r-rm-taskmean-prompt-n50}{38.0}
\defV{r-rm-ansgt-n50}{2}
\defV{r-calib-sys-usage}{0.2}
\defV{r-calib-user-usage}{94.8}
\defV{r-calib-ans-usage}{5.0}
\defV{r-calib-entok-usage}{74.6}
\defV{r-calib-entext-usage}{74.6}
\defV{r-calib-enrows-usage}{77.9}
\defV{r-calib-nosys-usage}{99.5}
\defV{r-calib-sysdistinct-usage}{1}
\defV{r-calib-sysen-usage}{1}
\defV{r-calib-ktok-usage}{592}
\defV{r-calib-sys-brand}{59.4}
\defV{r-calib-user-brand}{39.7}
\defV{r-calib-ans-brand}{0.9}
\defV{r-calib-entok-brand}{18.5}
\defV{r-calib-entext-brand}{66.7}
\defV{r-calib-enrows-brand}{18.8}
\defV{r-calib-nosys-brand}{0.0}
\defV{r-calib-sysdistinct-brand}{2}
\defV{r-calib-sysen-brand}{2}
\defV{r-calib-ktok-brand}{593}
\defV{r-calib-sys-categ}{20.9}
\defV{r-calib-user-categ}{76.6}
\defV{r-calib-ans-categ}{2.5}
\defV{r-calib-entok-categ}{15.1}
\defV{r-calib-entext-categ}{31.9}
\defV{r-calib-enrows-categ}{20.2}
\defV{r-calib-nosys-categ}{19.5}
\defV{r-calib-sysdistinct-categ}{4}
\defV{r-calib-sysen-categ}{4}
\defV{r-calib-ktok-categ}{592}
\defV{r-calib-sys-domains}{64.3}
\defV{r-calib-user-domains}{24.3}
\defV{r-calib-ans-domains}{11.4}
\defV{r-calib-entok-domains}{70.0}
\defV{r-calib-entext-domains}{88.8}
\defV{r-calib-enrows-domains}{71.0}
\defV{r-calib-nosys-domains}{16.9}
\defV{r-calib-sysdistinct-domains}{2}
\defV{r-calib-sysen-domains}{2}
\defV{r-calib-ktok-domains}{592}
\defV{r-calib-sys-reach}{74.6}
\defV{r-calib-user-reach}{1.4}
\defV{r-calib-ans-reach}{24.0}
\defV{r-calib-entok-reach}{17.6}
\defV{r-calib-entext-reach}{79.2}
\defV{r-calib-enrows-reach}{18.0}
\defV{r-calib-nosys-reach}{0.0}
\defV{r-calib-sysdistinct-reach}{50}
\defV{r-calib-sysen-reach}{50}
\defV{r-calib-ktok-reach}{592}
\defV{r-calib-sys-all}{43.9}
\defV{r-calib-user-all}{47.4}
\defV{r-calib-ans-all}{8.7}
\defV{r-calib-entok-all}{39.1}
\defV{r-calib-entext-all}{68.2}
\defV{r-calib-enrows-all}{59.8}
\defV{r-calib-nosys-all}{32.5}
\defV{r-calib-sysdistinct-all}{59}
\defV{r-calib-sysen-all}{59}
\defV{r-calib-ktok-all}{2{,}962}
\defV{r-kv-bytes-q}{98{,}304}
\defV{r-kv-kib-q}{96}
\defV{r-kv-attnlayers-q}{48}
\defV{r-kv-mambalayers-n}{23}
\defV{r-kv-mamba-mib-n}{46.8}
\defV{r-kv-bytes-n}{6{,}144}
\defV{r-kv-kib-n}{6}
\defV{r-kv-attnlayers-n}{6}
\defV{r-kv-wgib-q-base}{56.9}
\defV{r-kv-free-q-base-48}{does not fit}
\defV{r-kv-tokens-q-base-48}{does not fit}
\defV{r-kv-seq4k-q-base-48}{does not fit}
\defV{r-kv-free-q-base-80}{15.1}
\defV{r-kv-tokens-q-base-80}{165}
\defV{r-kv-seq4k-q-base-80}{40}
\defV{r-kv-free-q-base-96}{29.5}
\defV{r-kv-tokens-q-base-96}{323}
\defV{r-kv-seq4k-q-base-96}{78}
\defV{r-kv-wgib-q-reap25}{43.4}
\defV{r-kv-free-q-reap25-48}{does not fit}
\defV{r-kv-tokens-q-reap25-48}{does not fit}
\defV{r-kv-seq4k-q-reap25-48}{does not fit}
\defV{r-kv-gain-q-reap25-48}{n/a}
\defV{r-kv-free-q-reap25-80}{28.6}
\defV{r-kv-tokens-q-reap25-80}{313}
\defV{r-kv-seq4k-q-reap25-80}{76}
\defV{r-kv-gain-q-reap25-80}{+148}
\defV{r-kv-free-q-reap25-96}{43.0}
\defV{r-kv-tokens-q-reap25-96}{470}
\defV{r-kv-seq4k-q-reap25-96}{114}
\defV{r-kv-gain-q-reap25-96}{+148}
\defV{r-kv-wgib-q-reap50}{29.9}
\defV{r-kv-free-q-reap50-48}{13.3}
\defV{r-kv-tokens-q-reap50-48}{146}
\defV{r-kv-seq4k-q-reap50-48}{35}
\defV{r-kv-gain-q-reap50-48}{n/a}
\defV{r-kv-free-q-reap50-80}{42.1}
\defV{r-kv-tokens-q-reap50-80}{460}
\defV{r-kv-seq4k-q-reap50-80}{112}
\defV{r-kv-gain-q-reap50-80}{+295}
\defV{r-kv-free-q-reap50-96}{56.5}
\defV{r-kv-tokens-q-reap50-96}{618}
\defV{r-kv-seq4k-q-reap50-96}{150}
\defV{r-kv-gain-q-reap50-96}{+295}
\defV{r-kv-wgib-q-ream25}{43.4}
\defV{r-kv-free-q-ream25-48}{does not fit}
\defV{r-kv-tokens-q-ream25-48}{does not fit}
\defV{r-kv-seq4k-q-ream25-48}{does not fit}
\defV{r-kv-gain-q-ream25-48}{n/a}
\defV{r-kv-free-q-ream25-80}{28.6}
\defV{r-kv-tokens-q-ream25-80}{313}
\defV{r-kv-seq4k-q-ream25-80}{76}
\defV{r-kv-gain-q-ream25-80}{+148}
\defV{r-kv-free-q-ream25-96}{43.0}
\defV{r-kv-tokens-q-ream25-96}{470}
\defV{r-kv-seq4k-q-ream25-96}{114}
\defV{r-kv-gain-q-ream25-96}{+148}
\defV{r-kv-wgib-q-ream50}{29.9}
\defV{r-kv-free-q-ream50-48}{13.3}
\defV{r-kv-tokens-q-ream50-48}{146}
\defV{r-kv-seq4k-q-ream50-48}{35}
\defV{r-kv-gain-q-ream50-48}{n/a}
\defV{r-kv-free-q-ream50-80}{42.1}
\defV{r-kv-tokens-q-ream50-80}{460}
\defV{r-kv-seq4k-q-ream50-80}{112}
\defV{r-kv-gain-q-ream50-80}{+295}
\defV{r-kv-free-q-ream50-96}{56.5}
\defV{r-kv-tokens-q-ream50-96}{618}
\defV{r-kv-seq4k-q-ream50-96}{150}
\defV{r-kv-gain-q-ream50-96}{+295}
\defV{r-kv-wgib-n-base}{61.3}
\defV{r-kv-free-n-base-48}{does not fit}
\defV{r-kv-tokens-n-base-48}{does not fit}
\defV{r-kv-seq4k-n-base-48}{does not fit}
\defV{r-kv-free-n-base-80}{10.7}
\defV{r-kv-tokens-n-base-80}{1{,}869}
\defV{r-kv-seq4k-n-base-80}{154}
\defV{r-kv-free-n-base-96}{25.1}
\defV{r-kv-tokens-n-base-96}{4{,}385}
\defV{r-kv-seq4k-n-base-96}{362}
\defV{r-kv-wgib-n-reap25}{47.6}
\defV{r-kv-free-n-reap25-48}{does not fit}
\defV{r-kv-tokens-n-reap25-48}{does not fit}
\defV{r-kv-seq4k-n-reap25-48}{does not fit}
\defV{r-kv-gain-n-reap25-48}{n/a}
\defV{r-kv-free-n-reap25-80}{24.4}
\defV{r-kv-tokens-n-reap25-80}{4{,}260}
\defV{r-kv-seq4k-n-reap25-80}{352}
\defV{r-kv-gain-n-reap25-80}{+2{,}391}
\defV{r-kv-free-n-reap25-96}{38.8}
\defV{r-kv-tokens-n-reap25-96}{6{,}777}
\defV{r-kv-seq4k-n-reap25-96}{560}
\defV{r-kv-gain-n-reap25-96}{+2{,}391}
\defV{r-kv-wgib-n-reap50}{33.9}
\defV{r-kv-free-n-reap50-48}{9.3}
\defV{r-kv-tokens-n-reap50-48}{1{,}618}
\defV{r-kv-seq4k-n-reap50-48}{133}
\defV{r-kv-gain-n-reap50-48}{n/a}
\defV{r-kv-free-n-reap50-80}{38.1}
\defV{r-kv-tokens-n-reap50-80}{6{,}651}
\defV{r-kv-seq4k-n-reap50-80}{550}
\defV{r-kv-gain-n-reap50-80}{+4{,}782}
\defV{r-kv-free-n-reap50-96}{52.5}
\defV{r-kv-tokens-n-reap50-96}{9{,}168}
\defV{r-kv-seq4k-n-reap50-96}{758}
\defV{r-kv-gain-n-reap50-96}{+4{,}782}
\defV{r-kv-wgib-n-ream25}{47.6}
\defV{r-kv-free-n-ream25-48}{does not fit}
\defV{r-kv-tokens-n-ream25-48}{does not fit}
\defV{r-kv-seq4k-n-ream25-48}{does not fit}
\defV{r-kv-gain-n-ream25-48}{n/a}
\defV{r-kv-free-n-ream25-80}{24.4}
\defV{r-kv-tokens-n-ream25-80}{4{,}260}
\defV{r-kv-seq4k-n-ream25-80}{352}
\defV{r-kv-gain-n-ream25-80}{+2{,}391}
\defV{r-kv-free-n-ream25-96}{38.8}
\defV{r-kv-tokens-n-ream25-96}{6{,}777}
\defV{r-kv-seq4k-n-ream25-96}{560}
\defV{r-kv-gain-n-ream25-96}{+2{,}391}
\defV{r-kv-wgib-n-ream50}{33.9}
\defV{r-kv-free-n-ream50-48}{9.3}
\defV{r-kv-tokens-n-ream50-48}{1{,}618}
\defV{r-kv-seq4k-n-ream50-48}{133}
\defV{r-kv-gain-n-ream50-48}{n/a}
\defV{r-kv-free-n-ream50-80}{38.1}
\defV{r-kv-tokens-n-ream50-80}{6{,}651}
\defV{r-kv-seq4k-n-ream50-80}{550}
\defV{r-kv-gain-n-ream50-80}{+4{,}782}
\defV{r-kv-free-n-ream50-96}{52.5}
\defV{r-kv-tokens-n-ream50-96}{9{,}168}
\defV{r-kv-seq4k-n-ream50-96}{758}
\defV{r-kv-gain-n-ream50-96}{+4{,}782}
\defV{r-gentok-q-base-usage}{46}
\defV{r-gens-q-base-usage}{36}
\defV{r-gentok-q-base-brand}{47}
\defV{r-gens-q-base-brand}{75}
\defV{r-gentok-q-base-categ}{23}
\defV{r-gens-q-base-categ}{27}
\defV{r-gentok-q-base-domains}{42}
\defV{r-gens-q-base-domains}{11}
\defV{r-gentok-q-base-reach}{5{,}731}
\defV{r-gens-q-base-reach}{5{,}566}
\defV{r-gentok-q-base-all}{5{,}889}
\defV{r-gens-q-base-all}{5{,}713}
\defV{r-gentok-q-reap25-usage}{46}
\defV{r-gens-q-reap25-usage}{36}
\defV{r-gentok-q-reap25-brand}{43}
\defV{r-gens-q-reap25-brand}{73}
\defV{r-gentok-q-reap25-categ}{23}
\defV{r-gens-q-reap25-categ}{26}
\defV{r-gentok-q-reap25-domains}{40}
\defV{r-gens-q-reap25-domains}{11}
\defV{r-gentok-q-reap25-reach}{5{,}917}
\defV{r-gens-q-reap25-reach}{5{,}441}
\defV{r-gentok-q-reap25-all}{6{,}069}
\defV{r-gens-q-reap25-all}{5{,}586}
\defV{r-gentok-chg-q-reap25-all}{+3}
\defV{r-gens-chg-q-reap25-all}{\ensuremath{-}2}
\defV{r-gentok-chg-q-reap25-domains}{\ensuremath{-}7}
\defV{r-gens-chg-q-reap25-domains}{+2}
\defV{r-gentok-chg-q-reap25-reach}{+3}
\defV{r-gens-chg-q-reap25-reach}{\ensuremath{-}2}
\defV{r-gentok-q-reap50-usage}{46}
\defV{r-gens-q-reap50-usage}{35}
\defV{r-gentok-q-reap50-brand}{40}
\defV{r-gens-q-reap50-brand}{71}
\defV{r-gentok-q-reap50-categ}{24}
\defV{r-gens-q-reap50-categ}{26}
\defV{r-gentok-q-reap50-domains}{107}
\defV{r-gens-q-reap50-domains}{367}
\defV{r-gentok-q-reap50-reach}{8{,}462}
\defV{r-gens-q-reap50-reach}{9{,}305}
\defV{r-gentok-q-reap50-all}{8{,}679}
\defV{r-gens-q-reap50-all}{9{,}803}
\defV{r-gentok-chg-q-reap50-all}{+47}
\defV{r-gens-chg-q-reap50-all}{+72}
\defV{r-gentok-chg-q-reap50-domains}{+154}
\defV{r-gens-chg-q-reap50-domains}{+3{,}360}
\defV{r-gentok-chg-q-reap50-reach}{+48}
\defV{r-gens-chg-q-reap50-reach}{+67}
\defV{r-gentok-q-ream25-usage}{46}
\defV{r-gens-q-ream25-usage}{36}
\defV{r-gentok-q-ream25-brand}{46}
\defV{r-gens-q-ream25-brand}{74}
\defV{r-gentok-q-ream25-categ}{23}
\defV{r-gens-q-ream25-categ}{26}
\defV{r-gentok-q-ream25-domains}{43}
\defV{r-gens-q-ream25-domains}{10}
\defV{r-gentok-q-ream25-reach}{6{,}490}
\defV{r-gens-q-ream25-reach}{7{,}534}
\defV{r-gentok-q-ream25-all}{6{,}648}
\defV{r-gens-q-ream25-all}{7{,}680}
\defV{r-gentok-chg-q-ream25-all}{+13}
\defV{r-gens-chg-q-ream25-all}{+34}
\defV{r-gentok-chg-q-ream25-domains}{+2}
\defV{r-gens-chg-q-ream25-domains}{\ensuremath{-}4}
\defV{r-gentok-chg-q-ream25-reach}{+13}
\defV{r-gens-chg-q-ream25-reach}{+35}
\defV{r-gentok-q-ream50-usage}{46}
\defV{r-gens-q-ream50-usage}{34}
\defV{r-gentok-q-ream50-brand}{40}
\defV{r-gens-q-ream50-brand}{70}
\defV{r-gentok-q-ream50-categ}{22}
\defV{r-gens-q-ream50-categ}{26}
\defV{r-gentok-q-ream50-domains}{174}
\defV{r-gens-q-ream50-domains}{699}
\defV{r-gentok-q-ream50-reach}{11{,}364}
\defV{r-gens-q-ream50-reach}{15{,}506}
\defV{r-gentok-q-ream50-all}{11{,}647}
\defV{r-gens-q-ream50-all}{16{,}335}
\defV{r-gentok-chg-q-ream50-all}{+98}
\defV{r-gens-chg-q-ream50-all}{+186}
\defV{r-gentok-chg-q-ream50-domains}{+312}
\defV{r-gens-chg-q-ream50-domains}{+6{,}497}
\defV{r-gentok-chg-q-ream50-reach}{+98}
\defV{r-gens-chg-q-ream50-reach}{+179}
\defV{r-gentok-n-base-usage}{46}
\defV{r-gens-n-base-usage}{35}
\defV{r-gentok-n-base-brand}{39}
\defV{r-gens-n-base-brand}{112}
\defV{r-gentok-n-base-categ}{21}
\defV{r-gens-n-base-categ}{31}
\defV{r-gentok-n-base-domains}{62}
\defV{r-gens-n-base-domains}{28}
\defV{r-gentok-n-base-reach}{4{,}643}
\defV{r-gens-n-base-reach}{3{,}203}
\defV{r-gentok-n-base-all}{4{,}811}
\defV{r-gens-n-base-all}{3{,}409}
\defV{r-gentok-n-reap25-usage}{46}
\defV{r-gens-n-reap25-usage}{34}
\defV{r-gentok-n-reap25-brand}{38}
\defV{r-gens-n-reap25-brand}{111}
\defV{r-gentok-n-reap25-categ}{22}
\defV{r-gens-n-reap25-categ}{36}
\defV{r-gentok-n-reap25-domains}{101}
\defV{r-gens-n-reap25-domains}{216}
\defV{r-gentok-n-reap25-reach}{4{,}150}
\defV{r-gens-n-reap25-reach}{2{,}746}
\defV{r-gentok-n-reap25-all}{4{,}358}
\defV{r-gens-n-reap25-all}{3{,}143}
\defV{r-gentok-chg-n-reap25-all}{\ensuremath{-}9}
\defV{r-gens-chg-n-reap25-all}{\ensuremath{-}8}
\defV{r-gentok-chg-n-reap25-domains}{+63}
\defV{r-gens-chg-n-reap25-domains}{+665}
\defV{r-gentok-chg-n-reap25-reach}{\ensuremath{-}11}
\defV{r-gens-chg-n-reap25-reach}{\ensuremath{-}14}
\defV{r-gentok-n-reap50-usage}{46}
\defV{r-gens-n-reap50-usage}{34}
\defV{r-gentok-n-reap50-brand}{38}
\defV{r-gens-n-reap50-brand}{109}
\defV{r-gentok-n-reap50-categ}{21}
\defV{r-gens-n-reap50-categ}{30}
\defV{r-gentok-n-reap50-domains}{583}
\defV{r-gens-n-reap50-domains}{1{,}256}
\defV{r-gentok-n-reap50-reach}{10{,}512}
\defV{r-gens-n-reap50-reach}{4{,}273}
\defV{r-gentok-n-reap50-all}{11{,}200}
\defV{r-gens-n-reap50-all}{5{,}702}
\defV{r-gentok-chg-n-reap50-all}{+133}
\defV{r-gens-chg-n-reap50-all}{+67}
\defV{r-gentok-chg-n-reap50-domains}{+840}
\defV{r-gens-chg-n-reap50-domains}{+4{,}354}
\defV{r-gentok-chg-n-reap50-reach}{+126}
\defV{r-gens-chg-n-reap50-reach}{+33}
\defV{r-gentok-n-ream25-usage}{46}
\defV{r-gens-n-ream25-usage}{35}
\defV{r-gentok-n-ream25-brand}{38}
\defV{r-gens-n-ream25-brand}{111}
\defV{r-gentok-n-ream25-categ}{21}
\defV{r-gens-n-ream25-categ}{30}
\defV{r-gentok-n-ream25-domains}{145}
\defV{r-gens-n-ream25-domains}{592}
\defV{r-gentok-n-ream25-reach}{7{,}807}
\defV{r-gens-n-ream25-reach}{3{,}951}
\defV{r-gentok-n-ream25-all}{8{,}058}
\defV{r-gens-n-ream25-all}{4{,}719}
\defV{r-gentok-chg-n-ream25-all}{+68}
\defV{r-gens-chg-n-ream25-all}{+38}
\defV{r-gentok-chg-n-ream25-domains}{+134}
\defV{r-gens-chg-n-ream25-domains}{+1{,}998}
\defV{r-gentok-chg-n-ream25-reach}{+68}
\defV{r-gens-chg-n-ream25-reach}{+23}
\defV{r-gentok-n-ream50-usage}{46}
\defV{r-gens-n-ream50-usage}{33}
\defV{r-gentok-n-ream50-brand}{39}
\defV{r-gens-n-ream50-brand}{109}
\defV{r-gentok-n-ream50-categ}{21}
\defV{r-gens-n-ream50-categ}{30}
\defV{r-gentok-n-ream50-domains}{717}
\defV{r-gens-n-ream50-domains}{1{,}424}
\defV{r-gentok-n-ream50-reach}{12{,}327}
\defV{r-gens-n-ream50-reach}{4{,}630}
\defV{r-gentok-n-ream50-all}{13{,}150}
\defV{r-gens-n-ream50-all}{6{,}226}
\defV{r-gentok-chg-n-ream50-all}{+173}
\defV{r-gens-chg-n-ream50-all}{+83}
\defV{r-gentok-chg-n-ream50-domains}{+1{,}055}
\defV{r-gens-chg-n-ream50-domains}{+4{,}949}
\defV{r-gentok-chg-n-ream50-reach}{+165}
\defV{r-gens-chg-n-ream50-reach}{+45}
\defV{r-h20-reap}{16}
\defV{r-h20-ream}{2}
\defV{r-h20-ns}{2}
\defV{r-h-reap-noemail}{12}
\defV{r-h-reap-noemail-q}{5}
\defV{r-h-reap-noemail-n}{7}
\defV{r-h-ream-noemail}{2}
\defV{r-h-ream-noemail-q}{1}
\defV{r-h-ream-noemail-n}{1}
\defV{r-h-ns-noemail}{2}
\defV{r-h-ns-noemail-q}{2}
\defV{r-h-ns-noemail-n}{0}
\defV{r-h-total-noemail}{16}
\defV{r-ret-n-reap25-gen-noemail}{90.0}
\defV{r-retlo-n-reap25-gen-noemail}{86.4}
\defV{r-rethi-n-reap25-gen-noemail}{93.9}
\defV{r-ret-n-reap25-gen-email}{101.8}
\defV{r-retlo-n-reap25-gen-email}{89.8}
\defV{r-rethi-n-reap25-gen-email}{115.4}
\defV{r-ret-n-reap50-gen-noemail}{62.6}
\defV{r-retlo-n-reap50-gen-noemail}{59.1}
\defV{r-rethi-n-reap50-gen-noemail}{66.3}
\defV{r-ret-n-reap50-gen-email}{23.0}
\defV{r-retlo-n-reap50-gen-email}{18.2}
\defV{r-rethi-n-reap50-gen-email}{28.5}
\defV{r-ret-n-ream25-gen-noemail}{68.0}
\defV{r-retlo-n-ream25-gen-noemail}{64.4}
\defV{r-rethi-n-ream25-gen-noemail}{72.0}
\defV{r-ret-n-ream25-gen-email}{35.2}
\defV{r-retlo-n-ream25-gen-email}{29.0}
\defV{r-rethi-n-ream25-gen-email}{42.2}
\defV{r-ret-n-ream50-gen-noemail}{54.0}
\defV{r-retlo-n-ream50-gen-noemail}{50.5}
\defV{r-rethi-n-ream50-gen-noemail}{57.7}
\defV{r-ret-n-ream50-gen-email}{3.0}
\defV{r-retlo-n-ream50-gen-email}{1.4}
\defV{r-rethi-n-ream50-gen-email}{4.8}

\newcommand{\figref}[1]{Figure~\ref{#1}}
\newcommand{\tabref}[1]{Table~\ref{#1}}
\newcommand{\appref}[1]{Appendix~\ref{#1}}
\newcommand{\secref}[1]{Section~\ref{#1}}
\newcommand{\pp}{\,pp}
\newcolumntype{L}[1]{>{\raggedright\arraybackslash}p{#1}}
\title{Prune, Don't Merge:\\ Expert Compression of Mixture-of-Experts LLMs\\ on Multilingual Production Tasks}
\author{Tomas Rasymas\thanks{Corresponding author:
  \href{mailto:tomas.rasymas@hostinger.com}{tomas.rasymas@hostinger.com}.}\qquad Povilas Kvedaras\\[2pt]
  Hostinger Research}
\date{}

\begin{document}
\maketitle

\begin{abstract}
Mixture-of-Experts language models are cheap to run per token but expensive to hold in GPU memory, which drives the
cost of serving them in production. One-shot expert compression shrinks them after a single calibration pass. Its two
current state-of-the-art methods are REAP for expert pruning, which deletes the least salient experts, and REAM for
expert merging, which merges them into the most salient ones. Both have been evaluated only on academic benchmarks.
We compare the two on five multilingual production tasks of a web-hosting company, from classification to the
generation of marketing HTML email templates, with two 30B-parameter models (Qwen3-30B-A3B-Instruct-2507 and the
hybrid Mamba-attention Nemotron-3.5-Lightning) at 25\% and 50\% of experts removed and the same calibration data.
Pruning wins: on the same rows, REAP is significantly better in \V{h-reap} of \V{h-total} comparisons and REAM in
\V{h-ream}. At 25\%, REAP keeps \V{ret-q-reap25-gen}\% of Qwen3-30B's generative quality and REAM
\V{ret-q-ream25-gen}\%, although on benchmarks the two are within a point for the same model. The damage concentrates
in the output contract rather than the content: \V{reach-gateshare-min}--\V{reach-gateshare-max}\% of each significant
loss on email templates comes from hard-gate failures such as invalid JSON, broken mobile layouts and runaway
generation, and most failed non-English domain descriptions copy the prompt's English opening phrase. Cross-perplexity
and logit divergence, two diagnostics from the REAP paper, also track production quality, as does the rate of repetition
loops; n-gram diversity does not. We close with practical guidance for compressing MoE models for production.
\end{abstract}

\begin{quote}\small
\textbf{Keywords:} mixture of experts; expert pruning; expert merging; model compression; production evaluation;
multilingual generation
\end{quote}

\section{Introduction}\label{sec:intro}

Serving large language models (LLMs) in production is paid for in GPU memory and GPU hours. Mixture-of-Experts (MoE)
models change the balance between the two. A model with 30B parameters, of which about 3B are active for each token,
computes like a small dense model but must hold all of its experts in memory: \V{ckpt-q-base}\,GB for
Qwen3-30B-A3B-Instruct-2507 and \V{ckpt-n-base}\,GB for Nemotron-3.5-Lightning-30B-A3B in BF16. The weights compete with
the key--value cache for memory~\citep{kwon2023efficient,liu2026fluxmoe}, and decoding is bound by memory
bandwidth~\citep{yu2025efficient}. For high-volume workloads, the memory a model needs therefore decides how many GPUs
it takes.

\emph{Expert compression} attacks this cost directly. It removes whole routed experts from every MoE layer and leaves a
smaller checkpoint of the same architecture, which serving stacks such as vLLM load unchanged; the one-shot methods need
only a calibration pass. Two families compete, pruning and merging, and we compare the current state-of-the-art method
of each. REAP~\citep{lasby2025reap} is the state of the art in expert pruning: it outperforms earlier pruning and
merging methods, argues that merging introduces an irreducible error in fine-grained MoEs, and shows pruning ahead on
generative benchmarks. REAM~\citep{jha2026ream} is the state of the art in expert merging: it builds on REAP's own
saliency and, with the best of ten calibration mixtures, reports slightly higher generative scores than REAP at 25\%
compression of the same Qwen3 model that we study.

Both papers measure on academic benchmarks: multiple-choice suites, code, mathematics and creative writing. Production
tasks differ in ways those benchmarks do not test. Answers must satisfy an output contract (a JSON schema, HTML rules,
hard format gates). They serve users in dozens of languages, and a product's own judge or rubric accepts them. Some
answers run to thousands of tokens. And calibration data must come from product traffic, chosen without access to the
test metric.

We therefore evaluate REAP and REAM where they would be deployed. We compress two architecturally different 30B-A3B
MoE models with each method at 25\% and 50\%, on the same calibration data of production conversations. We then
evaluate all ten models on five production tasks of Hostinger, a web-hosting company, with 2,000 rows per task in
\V{langs-min}--\V{langs-max} languages and the production metrics. REAM chooses its merge centres with REAP's
saliency, so the two methods keep almost the same experts, and the comparison isolates what happens to the others.

\begin{figure}[t]
\centering
\includegraphics[width=\textwidth]{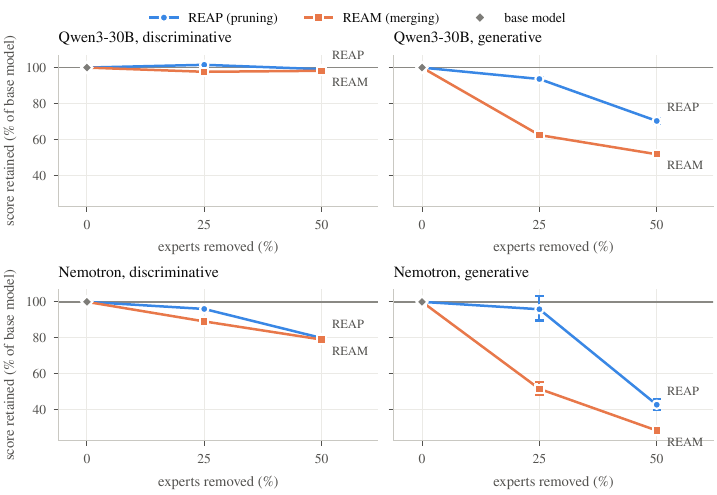}
\caption{\textbf{Merging falls behind on generation.} Mean score retained (\% of the base model's score) against the
share of experts removed, for the discriminative tasks (website cues, brand names, categorisation; left) and the
generative tasks (domain descriptions, email templates; right), one row per base model. Blue circles are REAP, orange
squares REAM, and the grey diamond is the base model (100\% by definition). Error bars are 95\% bootstrap CIs, and tasks
have equal weight. On Qwen3-30B's generative tasks, REAP at 50\% keeps more than REAM at 25\%.}
\label{fig:retention}
\end{figure}

We ask how much production quality each method keeps, and which of the two is better once the evaluation looks like
production. We then ask how the compressed models fail, and whether the damage is the same in every language. Next we
ask whether the diagnostics that the two papers use to explain their results agree with production outcomes.
\figref{fig:retention} previews the main answer.

\paragraph{Contributions.}
\begin{itemize}
  \item The first evaluation of REAP and REAM on production tasks: ten models, five tasks, 2,000 rows each, with paired
        bootstrap intervals and multiple-comparison correction (\secref{sec:setup}).
  \item A head-to-head result: REAP is significantly better than REAM in \V{h-reap} of \V{h-total} comparisons. On
        marketing email template generation it leads by up to \V{h-q25-reach} rubric points, and by
        \V{r-h2hprod-q25} points under the production scoring rules (\secref{sec:versus}).
  \item A failure analysis: compression mostly breaks the output contract, through format gates, runaway generation and
        English template phrases copied into non-English answers (\secref{sec:failures}).
  \item A test of which benchmark claims and diagnostics carry over to production (\secref{sec:diagnostics},
        \secref{sec:transfer}), and practical guidance (\secref{sec:discussion}).
\end{itemize}

\section{Related work}\label{sec:related}

\paragraph{Expert pruning and merging.} One-shot expert compression removes whole experts after a calibration pass.
Merging methods were reported to do better than pruning on discriminative benchmarks: M-SMoE aligns and merges
experts~\citep{li2024merge}, and HC-SMoE clusters them by their outputs~\citep{chen2025retraining}. REAP
\citep{lasby2025reap} reversed the verdict for generation. It argues that merging with summed gates loses the router's
independent control of the merged experts, and it prunes by router-weighted activation instead. REAM
\citep{jha2026ream} merges into REAP's most salient experts without summing gates. Hybrids choose per expert between
pruning and merging \citep{huang2026prime}, or repair the router afterwards \citep{hyeon2026is}.

Almost all of this work reports benchmark averages, and two studies show what such averages hide. With a fixed pruning
mask, multiple-choice accuracy stays at 92--97\% while GSM8K and code scores collapse \citep{ali2026what}. A
benchmark of composable MoE compression finds that removing experts is the main source of loss, and that REAM is
slightly behind REAP at 25\% \citep{benazir2026benchmarking}. That is the only independent REAP--REAM comparison we
know of. Studies closer to deployment prune for one domain, for example biomedicine on the same Qwen3 checkpoint and a
Nemotron~3 Nano model \citep{yamaguchi2026on}. None compares pruning with merging on production tasks, on output
contracts or across many languages.

\paragraph{Serving MoE models.} Every expert of an MoE model must stay resident in GPU memory, where the experts
compete with the key--value cache~\citep{liu2026fluxmoe}, and decoding cost grows with the number of distinct experts
read per step~\citep{yu2025efficient}. Offloading experts to host memory saves GPU memory but adds transfer
latency~\citep{liu2026fluxmoe}. Quantisation lowers the bits per expert instead. Expert compression is complementary:
it removes experts and can be combined with quantisation, although the effects do not simply
add~\citep{benazir2026benchmarking}.

\paragraph{Calibration data.} Compression quality depends on the calibration data~\citep{ji2025beware}. At high
sparsity, calibration choices that look alike on commonsense accuracy differ by 51.9 points on
code~\citep{xu2026rethinking}. For MoE models, the calibration mixture moves REAM between multiple-choice and
generative strength~\citep{jha2026ream}. Standard calibration also covers mostly prompt tokens, and adding the model's
own generated tokens helps decode-heavy tasks~\citep{lucas2026reasoning}.

Evidence on the calibration \emph{language} is split:
\begin{itemize}
  \item English-only calibration transferred across seven languages for expert pruning of
        gpt-oss-20b~\citep{salim2026towards};
  \item in dense pruning, target-language calibration kept perplexity but not downstream
        accuracy~\citep{kurz2026limitations};
  \item English calibration eroded Lithuanian in expert-pruned MoE models~\citep{rasymas2026english}.
\end{itemize}

\paragraph{Multilingual effects and language confusion.} Compression hurts other languages more than English.
Quantisation damages non-Latin-script languages in ways that automatic metrics understate~\citep{marchisio2024how}.
Multilingual pruning hurts under-represented languages first~\citep{choenni2025mwanda}. Pruned models often still
recognise an answer in multiple choice but fail to generate it: for 12\% of English items, and for 34--50\% of items in
other languages~\citep{wen2026the}. Inside MoE models, masking late-layer language-specific experts disrupts language
consistency~\citep{chen2026understanding}.

English-centric models also produce \emph{language confusion}: whole lines or single words in the wrong
language~\citep{marchisio2024understanding}, a failure traced to the final layers~\citep{nie2025mechanistic}. The
domain-description failures in \secref{sec:failures} are of the word-level kind.

\paragraph{Evaluating compressed models beyond benchmarks.} Aggregate accuracy hides how many answers flip: per-item
churn can be several times the net change~\citep{dutta2024accuracy}. Compression also damages capabilities that
averages miss: parseable output~\citep{ning2026acecompression}, instruction following and JSON-schema
adherence~\citep{salim2026the}, and freedom from repetition loops~\citep{lee2026focus}.

Evaluations of compression on production data are rare~\citep{behdin2025scaling}; we found none for MoE expert
compression. LLM judges are themselves inconsistent~\citep{haldar2025rating}. We therefore score with the production
judge that accepted the references, and add rule-based checks.

\section{The two methods}\label{sec:methods}

An MoE layer with $N$ routed experts $E_1,\dots,E_N$ sends each token $x$ to the top-$k$ experts $T(x)$ chosen by its
router, and combines their outputs with gate weights $g_j(x)$: $y(x) = \sum_{j \in T(x)} g_j(x)\, E_j(x)$, plus a
shared expert in Nemotron. Both methods remove a fixed share $r$ of the routed experts in every MoE layer, delete their
router rows and keep $k$. Shared experts, attention and Mamba layers are untouched.

\paragraph{REAP: pruning.} REAP~\citep{lasby2025reap} scores each expert by its mean gate-weighted output norm over the
calibration tokens routed to it, $X_j = \{x : j \in T(x)\}$:
\begin{equation}
  S_j = \frac{1}{|X_j|} \sum_{x \in X_j} g_j(x)\, \lVert E_j(x) \rVert_2 ,
\end{equation}
and deletes the $rN$ experts with the lowest $S_j$. The router then chooses its top-$k$ among the experts that remain.

\paragraph{REAM: merging.} REAM~\citep{jha2026ream} ranks experts by the same saliency and makes the $(1-r)N$ most
salient ones group \emph{centres}. In order of saliency, each centre takes the unassigned experts most similar to it,
measured by router logits and gate-weighted outputs, until its group holds $C$ experts. Following REAM, $C=16$ at 25\%
and $C=32$ at 50\%. Members are aligned to their centre neuron by neuron (Hungarian matching) and averaged with saliency
weights. The router keeps only the centres' rows, so REAM, unlike earlier merging methods, does not sum gates.

\paragraph{Implementation.} REAP is llm-compressor's \texttt{REAPPruningModifier}~\citep{redhatai2024llm}. REAM is our
port of the official code to the same pipeline, needed because the original supports only Qwen-style layers. Both run
sequentially, each layer calibrated on the outputs of the already compressed layers before it; REAM's paper calibrated
its baselines, REAP included, in a single pass. The port deviates from the original code in four documented ways,
among them that the group size $C$ counts the centre. Unit tests check that grouping and alignment reproduce the
original functions, and all runs use deterministic kernels (\appref{app:implementation} lists the deviations).

\section{Experimental setup}\label{sec:setup}

\paragraph{Tasks.} The five tasks are LLM calls from Hostinger products (\tabref{tab:tasks}); \appref{app:samples}
shows an anonymised reference answer for each. Three are discriminative. \emph{Website cues} asks whether a website description
mentions colours or fonts. \emph{Brand names} asks the model to copy the user's own brand name from a description.
\emph{Product categorisation} assigns an e-commerce product to a store category from an open label set, usually from a
candidate list. Two are generative. \emph{Domain descriptions} asks for a one- or two-sentence marketing text for a
domain name in a requested language, and the prompt asks it to open with ``This domain is''. \emph{Email templates}
asks an email-marketing product for a complete, responsive HTML template of a marketing email, such as a newsletter, a
promotion, or a transactional email sent after a purchase or a sign-up, returned inside JSON.

References are production outputs of an earlier production model that the product's LLM judge accepted, not human
labels; the brand task keeps only rows with a stated brand. Each task has 2,000 validation rows, the same for every
model, balanced over five reporting languages (en, pt, es, fr, id) and an ``other'' group as far as the data allow.

\begin{table}[t]
\caption{\textbf{The five production tasks.} Lang.\ = number of languages in the 2,000 validation rows. Tokens = median
answer length of the base Qwen3-30B.}
\label{tab:tasks}
\centering\small
\resizebox{\textwidth}{!}{\setlength{\tabcolsep}{3.5pt}
\begin{tabular}{@{}L{2.25cm}L{1.95cm}L{1.75cm}lrL{1.9cm}r@{}}
\toprule
Task & Type & Output & Decoding & Lang. & Primary metric & Tokens \\
\midrule
Website cues & classification & JSON, 2 booleans & greedy & 19 & exact match & 23 \\
Brand names & span extraction & JSON (schema) & greedy & 24 & exact match & 24 \\
Categorisation & open-label categorisation & JSON (schema) & greedy & 33 & judge pass rate & 11 \\
Domain descriptions & short generation & plain text & sampled & 17 & judge pass rate & 20 \\
Email templates & long structured generation & JSON with HTML & sampled & 35 & rubric score (0--100) & 2{,}724 \\
\bottomrule
\end{tabular}
}
\end{table}

\paragraph{Models.} Both base models have about 30B total and 3B active parameters and 128 routed experts per MoE layer.
Qwen3-30B-A3B-Instruct-2507~\citep{qwen2025qwen3} is a Transformer with 48 MoE layers, top-8 routing, a softmax router
and gated experts. NVIDIA-Nemotron-3.5-Lightning-30B-A3B~\citep{nvidia2026nemotron} is a hybrid of
Mamba-2~\citep{dao2024transformers}, attention and MoE layers, derived from Nemotron~3 Nano~\citep{nvidia2025nemotron}.
It has 23 MoE layers, top-6 routing, a sigmoid router with bias, one shared expert and gateless squared-ReLU experts.
Thinking is off. We call them Qwen3-30B and Nemotron.

\paragraph{Compression and calibration.} Each method removes 25\% or 50\% of the routed experts in every MoE layer
(128 $\to$ 96 or 64), giving eight compressed models. All runs of a model share one calibration set, disjoint from
validation: whole production conversations (system prompt, user message, accepted answer) in the model's chat template,
about 590k tokens per task and \V{calib-samples} conversations in total. Answers are only \V{r-calib-ans-all}\% of the
calibration tokens. All \V{r-calib-sysdistinct-all} distinct system prompts are English, so at least
\V{r-calib-entext-all}\% of the calibration text is English, even though the rows of three tasks are
language-balanced. Website cues and domain descriptions are also \V{calib-en-usage}\% and \V{calib-en-domains}\%
English by rows (\appref{app:calibration}).

\paragraph{Generation.} All models are served with vLLM~0.32~\citep{kwon2023efficient}. Following REAP, decoding is
greedy for the three discriminative tasks. Domain descriptions and email templates use each model's default sampling,
with the seed fixed per row so that every model draws the same random numbers on a row. As in production, brand names
and categorisation are constrained to their JSON schema, with vLLM's structured outputs, which use the XGrammar
constrained-decoding engine~\citep{dong2025xgrammar}. Answers are capped at 32,768 tokens
(1,024 for the schema-constrained tasks).

\paragraph{Metrics.} The primary metrics are those of production. Website cues and brand names use exact match.
Categorisation and domain descriptions use the pass rate of the production LLM judge, OpenAI gpt-6.1-sol, the judge
that accepted the references. Email templates use the production rubric. It scores 13 categories with static checks,
checks on the rendered email at desktop and mobile width, and LLM text and vision judges, and it sets the score to 0 if
any of five \emph{hard gates} fails: output parseable, HTML present, HTML complete, unsubscribe link, mobile layout
intact.

The production email score swung on one habit. The base Qwen3-30B often returns a complete \texttt{<body>} without the
\texttt{<html>} wrapper: it fails \texttt{html\_complete} in \V{gateprod-htmlcomplete-n-q-base} of 2,000 answers and
scores \V{reach-prod-q-base}. We therefore relax two gates: a complete \texttt{<body>} passes \texttt{html\_complete},
and any unsubscribe link passes \texttt{unsubscribe\_link}. This lifts the base model to \V{reach-rel-q-base}. We
report the production score as well (\appref{app:reach}).

\paragraph{Statistics.} Every comparison is paired, on the same 2,000 rows: a compressed model against its own base
model, or REAP against REAM. We report 95\% paired bootstrap confidence intervals (CIs; 10,000
resamples~\citep{efron1993introduction}). We test with McNemar's exact test for pass/fail metrics~\citep{mcnemar1947note}
and a paired sign-flip permutation test on the mean for email scores, Holm-corrected~\citep{holm1979simple} over the
\V{n-comp} base comparisons and, separately, over the \V{h-total} head-to-head comparisons. \emph{Significant} means
Holm-corrected $p<0.05$. Pass/fail metrics are percentages, so their changes are percentage points (pp); email scores
are rubric points. Retention is the compressed score divided by the base score. Each configuration was compressed once,
so the tests compare these checkpoints on these rows; they do not cover the variance from compressing with another
calibration sample.

\section{Results}\label{sec:results}

\subsection{How much quality survives compression}\label{sec:quality}

\tabref{tab:main} gives every change from the base model (\figref{fig:deltas} in \appref{app:full} plots them with
their CIs). Of the \V{n-comp} changes, \V{n-sig} are significant: \V{n-sig-worse} losses and \V{n-sig-better} gains,
all three gains on Qwen3-30B's brand names.

REAP at 25\% is close to lossless for Qwen3-30B outside generation. Website cues (\V{d-q-reap25-usage}\pp),
categorisation (\V{d-q-reap25-categ}\pp) and email templates (\V{d-q-reap25-reach} points) change by amounts that are
not significant, and brand names gain \V{d-q-reap25-brand}\pp. Its one significant loss is domain descriptions
(\V{d-q-reap25-domains}\pp).

Nemotron is more fragile: REAP at 25\% already loses \V{d-n-reap25-brand}\pp{} on brand names,
\V{d-n-reap25-categ}\pp{} on categorisation and \V{d-n-reap25-domains}\pp{} on domain descriptions. Removing half the
experts costs generation most. Qwen3-30B with REAP at 50\% keeps \V{ret-q-reap50-disc}\% of its discriminative score,
but loses \V{d-q-reap50-domains}\pp{} on domain descriptions and \V{d-q-reap50-reach} points on email templates. The CIs
are \V{ciw-other-min}--\V{ciw-other-max}\pp{} wide (\V{ciw-usage-min}--\V{ciw-usage-max}\pp{} for website cues), far
narrower than the large changes.

\begin{table}[t]
\caption{\textbf{Primary metric of each base model and change of each compressed model} (compressed $-$ base on the
same 2,000 rows, in pp; rubric points for email templates). \textbf{Bold}: significant after Holm correction over the
\V{n-comp} comparisons. Retention rows: mean score retained (\% of base) over the discriminative (first three) and the
generative (last two) tasks. \appref{app:full} gives every CI and $p$-value.}
\label{tab:main}
\centering\small
\resizebox{\textwidth}{!}{\begin{tabular}{@{}lr rrrr@{}}
\toprule
 & Base & REAP 25\% & REAP 50\% & REAM 25\% & REAM 50\% \\
\midrule
\multicolumn{6}{@{}l}{\textbf{Qwen3-30B}} \\
\quad Website cues & 98.8 & $-$0.2 & \textbf{$-$1.0} & $-$0.5 & \textbf{$-$1.7} \\
\quad Brand names & 64.8 & \textbf{+3.9} & \textbf{+4.2} & $-$0.3 & \textbf{+5.9} \\
\quad Categorisation & 81.7 & $-$1.0 & \textbf{$-$6.7} & \textbf{$-$5.0} & \textbf{$-$10.4} \\
\quad Domain descriptions & 93.5 & \textbf{$-$13.0} & \textbf{$-$22.2} & \textbf{$-$31.1} & \textbf{$-$30.1} \\
\quad Email templates & 70.2 & +0.9 & \textbf{$-$25.0} & \textbf{$-$29.3} & \textbf{$-$45.1} \\
\quad \textit{Retention, discr.\ (\%)} & 100 & 101.5 & 99.1 & 97.7 & 98.2 \\
\quad \textit{Retention, gen.\ (\%)} & 100 & 93.6 & 70.3 & 62.5 & 51.8 \\
\midrule
\multicolumn{6}{@{}l}{\textbf{Nemotron}} \\
\quad Website cues & 96.0 & +0.5 & \textbf{$-$1.3} & $-$0.5 & \textbf{$-$4.4} \\
\quad Brand names & 49.8 & \textbf{$-$4.0} & \textbf{$-$21.1} & \textbf{$-$10.0} & \textbf{$-$15.6} \\
\quad Categorisation & 68.2 & \textbf{$-$3.0} & \textbf{$-$11.3} & \textbf{$-$8.3} & \textbf{$-$18.3} \\
\quad Domain descriptions & 58.8 & \textbf{$-$5.9} & \textbf{$-$22.0} & \textbf{$-$18.8} & \textbf{$-$27.1} \\
\quad Email templates & 12.8 & +0.2 & \textbf{$-$9.8} & \textbf{$-$8.3} & \textbf{$-$12.4} \\
\quad \textit{Retention, discr.\ (\%)} & 100 & 96.0 & 79.9 & 89.1 & 79.0 \\
\quad \textit{Retention, gen.\ (\%)} & 100 & 95.9 & 42.8 & 51.6 & 28.5 \\

\bottomrule
\end{tabular}
}
\end{table}

\subsection{Pruning versus merging}\label{sec:versus}

\begin{figure}[t]
\centering
\includegraphics[width=\textwidth]{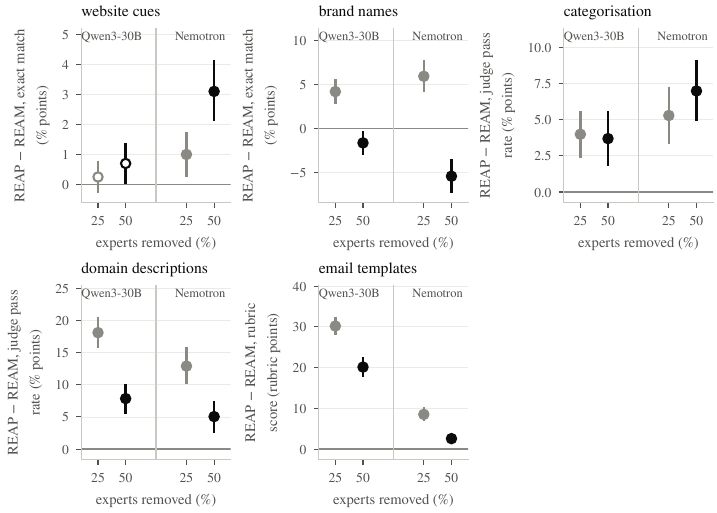}
\caption{\textbf{REAP beats REAM on the same rows in 16 of 20 comparisons.} REAP $-$ REAM in each task's metric
(positive = REAP better), at 25\% (grey) and 50\% (black) of experts removed for each base model. Bars are 95\%
paired-bootstrap CIs. Filled markers are significant after Holm correction over the \V{h-total} comparisons. REAM is
better only on brand names at 50\%.}
\label{fig:h2h}
\end{figure}

Compared directly on the same rows (\figref{fig:h2h}), REAP is significantly better in \V{h-reap} of \V{h-total}
comparisons and REAM in \V{h-ream}. The remaining \V{h-ns}, website cues on Qwen3-30B, show no significant difference.
The largest gaps are on generation, both for Qwen3-30B at 25\%: \V{h-q25-reach} rubric points [\V{hlo-q25-reach},
\V{hhi-q25-reach}] on email templates and \V{h-q25-domains}\pp{} [\V{hlo-q25-domains}, \V{hhi-q25-domains}] on domain
descriptions.

\figref{fig:retention} shows the pattern that REAP's analysis predicts. On Qwen3-30B's discriminative tasks both methods
keep almost everything at 25\% (\V{ret-q-reap25-disc}\% and \V{ret-q-ream25-disc}\%), while on generation REAM falls
far behind (\V{ret-q-ream25-gen}\% against \V{ret-q-reap25-gen}\%). REAP with half the experts removed
(\V{ret-q-reap50-gen}\%) even keeps more than REAM with a quarter removed. On Nemotron, merging also loses on the
discriminative tasks at 25\% (\V{ret-n-ream25-disc}\% against \V{ret-n-reap25-disc}\%).

The result depends neither on the relaxed email-template gates nor on Nemotron's weak results on that task. Under the
production gates
REAP's lead grows to \V{r-h2hprod-q25} rubric points at 25\% and \V{r-h2hprod-q50} at 50\% on Qwen3-30B, with all four
production-rule comparisons significant (\appref{app:reach}). Without the email-template task the count is
\V{r-h-reap-noemail} for REAP, \V{r-h-ream-noemail} for REAM and \V{r-h-ns-noemail} ties, out of
\V{r-h-total-noemail}. Nemotron's base score on email templates is only \V{reach-rel-n-base}, which makes its
generative retention noisy: at REAP 25\% it is \V{r-ret-n-reap25-gen-email}\% [\V{r-retlo-n-reap25-gen-email},
\V{r-rethi-n-reap25-gen-email}] on email templates, against
\V{r-ret-n-reap25-gen-noemail}\% on domain descriptions alone.

REAM's two wins are both on brand names at 50\%, and they differ in kind (\appref{app:secondary}). On Qwen3-30B it is a
detection effect: REAM returns a brand more often (detection \V{r-brand-det-q-ream50}\% against
\V{r-brand-det-q-reap50}\%), while REAP's returned brands are slightly more often right (precision
\V{r-brand-prec-q-reap50}\% against \V{r-brand-prec-q-ream50}\%). Every validation row states a brand, so abstaining is
always wrong and REAM's gain is entirely detection; on real traffic, where about a quarter of requests name no brand, it
could cost accuracy. The same effect explains Qwen3-30B's gains over its base model on this task, since the base model
answers ``no brand'' too often (detection \V{r-brand-det-q-base}\%). On Nemotron, REAM's win is genuine precision
(\V{r-brand-prec-n-ream50}\% against \V{r-brand-prec-n-reap50}\%).

\subsection{How compressed models fail}\label{sec:failures}

The failures cluster on the two generative tasks, and most are failures to keep the output contract
(\figref{fig:failures} in \appref{app:reach} gives every failure rate). On the three discriminative tasks the output
stays valid: the schema keeps JSON valid in at least 99.9\% of brand and categorisation answers, and website cues stay
at least 99.7\% valid, so the failures there are wrong answers. Small net changes hide many of them. Qwen3-30B REAP 25\%
loses \V{lost-q-reap25-categ}\% of the categorisation rows its base model passed and gains
\V{gained-q-reap25-categ}\%, and only \V{same-q-reap25-categ}\% of its categories equal the base model's
(\figref{fig:agreement} in \appref{app:agreement}).

\paragraph{Email templates: compression breaks the contract, not the content.} The template score is the gate pass rate
times the mean score of the answers that pass, so each loss splits into a gate part and a quality part (a symmetric
split; \tabref{tab:reach}). For every significant loss, \V{reach-gateshare-min}--\V{reach-gateshare-max}\% is the gate
part. The templates that pass change much less: Qwen3-30B's average \V{reach-meanpass-q-base} for the base model,
\V{reach-meanpass-q-reap50} for REAP 50\% and \V{reach-meanpass-q-ream25} for REAM 25\%.

Three failures dominate (\appref{app:reach}). Qwen3-30B's rendered templates overflow a narrow screen in
\V{gate-mobilelayoutintact-q-base}\% of the base model's answers, but in \V{gate-mobilelayoutintact-q-reap50}\% at
REAP 50\%, \V{gate-mobilelayoutintact-q-ream25}\% at REAM 25\% and \V{gate-mobilelayoutintact-q-ream50}\% at REAM 50\%.
Its JSON becomes invalid more often (\V{reach-jsoninvalid-q-base}\% $\to$ \V{reach-jsoninvalid-q-ream50}\% at REAM
50\%). And answers run away to the 32k-token cap, repeating text: \V{reach-capped-q-base} of 2,000 for the base model,
\V{reach-capped-q-reap50} at REAP 50\% and \V{reach-capped-q-ream50} at REAM 50\%. Runaway answers also cost GPU time:
Qwen3-30B REAM 50\% generates \V{r-gentok-q-ream50-reach} thousand email-template tokens, where the base model generates
\V{r-gentok-q-base-reach} thousand.

Nemotron's base model already fails the contract. \V{reach-jsoninvalid-n-n-base} of its answers are invalid JSON, and
it passes every gate in only \V{reach-passall-n-base}\%. Its low score therefore reflects instruction following rather
than template quality: its passing templates average \V{reach-meanpass-n-base}.

\begin{table}[t]
\caption{\textbf{Email templates: where the score goes.} Production and relaxed rubric scores; the share of answers
passing every (relaxed) gate, and their mean relaxed score; the share of the loss against the base model that comes
from gate failures (a symmetric split of score = pass rate $\times$ mean of passing answers; significant losses only);
and the answers that ran to the 32k-token cap (of 2,000).}
\label{tab:reach}
\centering\small
\resizebox{\textwidth}{!}{\begin{tabular}{@{}lrrrrrr@{}}
\toprule
 & \multicolumn{2}{c}{Score (0--100)} & Passes all & Mean score if & Loss from & Runaway \\
\cmidrule(lr){2-3}
Model & production & relaxed & gates (\%) & all gates pass & gates (\%) & answers \\
\midrule
Qwen3-30B & 26.1 & 70.2 & 88.7 & 79.2 &  & 6 \\
\quad REAP 25\% & 51.3 & 71.1 & 89.9 & 79.1 &  & 14 \\
\quad REAP 50\% & 37.7 & 45.3 & 59.2 & 76.5 & 92.1 & 102 \\
\quad REAM 25\% & 0.4 & 40.9 & 54.5 & 75.0 & 89.8 & 61 \\
\quad REAM 50\% & 1.9 & 25.1 & 34.4 & 73.1 & 91.7 & 215 \\
\midrule
Nemotron & 11.9 & 12.8 & 17.8 & 72.0 &  & 38 \\
\quad REAP 25\% & 12.1 & 13.0 & 17.9 & 72.5 &  & 32 \\
\quad REAP 50\% & 2.6 & 2.9 & 4.3 & 67.7 & 95.2 & 226 \\
\quad REAM 25\% & 2.9 & 4.5 & 6.7 & 67.7 & 93.7 & 166 \\
\quad REAM 50\% & 0.1 & 0.4 & 0.6 & 63.1 & 93.4 & 299 \\
\bottomrule
\end{tabular}
}
\end{table}

\paragraph{Domain descriptions: the English opener.} The prompt is in English and asks for an answer ``in such form
\emph{This domain is}'' in the requested language; the references translate that opener. Compressed Qwen3-30B models
increasingly copy it untranslated and continue in the requested language (``This domain adalah\ldots'', ``This dominio
es\ldots''). This is a word-level form of language confusion~\citep{marchisio2024understanding}, and the judge rejects
it (\tabref{tab:domains}).

At REAM 25\%, \V{r-opener-q-ream25}\% of the non-English rows open in English, against \V{r-opener-q-base}\% for the
base model, while only \V{r-deten-q-ream25}\% are entirely English. The answers that keep the target language pass
\V{r-pass-neither-q-ream25}\% of the time, close to the base model's \V{r-pass-neither-q-base}\%. At the base model's
opener rate, REAM 25\%'s non-English pass rate would be \V{r-cf-pass-q-ream25}\% instead of \V{r-passnonen-q-ream25}\%.
The opener therefore explains \V{r-drift-explains-share-q-ream25}\% of REAM 25\%'s non-English loss, but only
\V{r-drift-explains-share-q-reap25}\% of REAP 25\%'s. Nemotron, by contrast, answers in the wrong language outright,
mostly English: \V{r-wronglang-n-base}\% of its base model's non-English rows, rising to \V{r-wronglang-n-reap50}\% at
REAP 50\%.

\begin{table}[t]
\caption{\textbf{Domain descriptions: compressed Qwen3-30B copies the English opener.} Judge pass rate on English and
non-English rows; among non-English rows, the share that opens with the English ``This'' but is otherwise in the
requested language, the share in the wrong language (and how many of those are English), the share the language
detector could not classify, and the pass rate of rows with neither failure; and the non-English pass rate the model
would have at its base model's opener and wrong-language rates.}
\label{tab:domains}
\centering\small
\resizebox{\textwidth}{!}{\setlength{\tabcolsep}{3pt}
\begin{tabular}{@{}lrrrrrrrr@{}}
\toprule
 & \multicolumn{2}{c}{Pass (\%)} & \multicolumn{4}{c}{Non-English rows (\%)} & \multicolumn{2}{c}{Pass non-En.} \\
\cmidrule(lr){2-3}\cmidrule(lr){4-7}\cmidrule(l){8-9}
 & & non- & opener & wrong & of which & & & at base \\
Model & En. & En. & only & lang. & En. & undet. & neither & drift \\
\midrule
Qwen3-30B & 98.2 & 92.6 & 3.3 & 0.2 & 0 & 1.7 & 95.8 & -- \\
\quad REAP 25\% & 99.4 & 76.6 & 11.5 & 0.2 & 0 & 1.8 & 86.3 & 82.6 \\
\quad REAP 50\% & 96.8 & 66.0 & 17.0 & 6.7 & 97 & 2.2 & 86.7 & 82.5 \\
\quad REAM 25\% & 97.7 & 55.1 & 39.6 & 2.4 & 54 & 1.8 & 94.4 & 89.7 \\
\quad REAM 50\% & 95.6 & 56.8 & 23.2 & 10.0 & 90 & 0.7 & 84.5 & 80.2 \\
\midrule
Nemotron & 92.7 & 51.7 & 19.4 & 14.0 & 84 & 1.4 & 69.4 & -- \\
\quad REAP 25\% & 93.3 & 44.5 & 25.8 & 15.0 & 90 & 0.7 & 61.9 & 47.3 \\
\quad REAP 50\% & 96.2 & 24.5 & 36.9 & 25.4 & 90 & 1.3 & 45.1 & 33.8 \\
\quad REAM 25\% & 86.3 & 30.4 & 26.2 & 12.7 & 53 & 11.2 & 45.0 & 34.8 \\
\quad REAM 50\% & 78.9 & 22.0 & 18.8 & 16.0 & 62 & 5.7 & 32.3 & 23.4 \\
\bottomrule
\end{tabular}
}
\end{table}

\subsection{Is the damage the same in every language?}\label{sec:languages}

Mostly yes, with one task-specific exception (\figref{fig:languages} in \appref{app:languages}). For every task and
compressed model we tested the difference between the change on non-English rows and the change on English rows.
\V{r-langgap-sig} of the \V{r-langgap-cells} cells are significant after Holm correction, all with non-English worse,
and \V{r-langgap-sig-domains} of them are domain descriptions, with a mean gap of \V{r-langgap-mean-domains}\pp. Outside
that task the gaps are small (means of \V{r-langgap-mean-usage} to \V{r-langgap-mean-brand}\pp, and
\V{r-langgap-mean-reach} rubric points on email templates), and only \V{r-langgap-sig-nodomains} cell is significant.

The domain-description gap is partly built into the task, because English rows cannot fail by leaving the opener
untranslated. It is also consistent with the calibration, which is at least \V{r-calib-entext-all}\% English text and
\V{calib-en-domains}\% English rows for this task, and with studies of language-specific
experts~\citep{chen2026understanding} and of expert pruning with English calibration~\citep{rasymas2026english}. Our
data cannot separate these explanations. Nemotron's wrong-language answers are the
clearer case of non-English capacity loss.

\subsection{Do the benchmark diagnostics agree with production?}\label{sec:diagnostics}

REAP explains its result with three measurements of generated text: cross-perplexity (the base model's perplexity of a
compressed model's answers), completion-logit divergence from the base model, and n-gram diversity, aimed at the
repetitiveness of merged models. REAM proposes the rank of the pre-logit embeddings as a quality signal. We scored each
diagnostic by how well it orders the five variants of each task and model by quality (Kendall $\tau$), and by whether it
ranks REAP above REAM where quality does (\tabref{tab:diag} in \appref{app:diagnostics}).

Cross-perplexity agrees best ($\tau=\V{r-tau-xppl}$, positive in \V{r-taupos-xppl} of \V{r-taucells-xppl} cells).
REAM's answers are less likely under the base model than REAP's in all \V{xppl-cells} cells, for example
\V{xppl-q-ream25-domains} against \V{xppl-q-reap25-domains} on Qwen3-30B's domain descriptions at 25\%. Logit divergence
agrees in level ($\tau=\V{r-tau-jsd}$) and is larger for REAM in \V{jsd-ream-gt} of \V{jsd-cells} cells. Repetition is
best measured directly. At 25\%, REAM produces \V{r-loops-ratio-q25-reach} times (Qwen3-30B) and
\V{r-loops-ratio-n25-reach} times (Nemotron) as many runaway email-template answers as REAP, and the loop rate tracks quality
($\tau=\V{r-tau-loop}$). n-gram diversity tracks quality only when the loops are included ($\tau=\V{r-tau-divloops}$
with them, \V{r-tau-div} without). Effective rank is a weaker positive signal ($\tau=\V{r-tau-erank}$), and numerical
rank is full for every model on four of the five tasks. These diagnostics come without CIs and over few cells, so they
support evaluation on the task but do not replace it.

\section{What transfers from benchmarks to production}\label{sec:transfer}

\tabref{tab:transfer} sets the claims of the two papers against our results. The settings differ in three ways that
matter. REAP evaluates Qwen3-30B-A3B with thinking off, not the Instruct-2507 checkpoint, with domain-specific
calibration. REAM evaluates the same checkpoint as we do but reports each method at its best of ten calibration
mixtures; averaged over the ten, the two methods are level (GEN 64.0 for REAP and 63.8 for REAM at 25\%, with REAM
ahead in 6 of 10 mixtures). And REAM calibrated its baselines in one pass and itself sequentially, a choice its own
ablation values at about one point. Our verdicts therefore concern 128-expert 30B-A3B models with production
calibration. REAP's theory concerns merging with summed gates, which REAM avoids; that REAP's conclusion also holds
against REAM is an empirical extension, not a test of the theory.

\begin{table}[t]
\caption{\textbf{Claims of the REAP and REAM papers against our production results.} Benchmark numbers are quoted from
the papers (section in brackets); ours are from this study.}
\label{tab:transfer}
\centering\footnotesize
\setlength{\tabcolsep}{3pt}
\begin{tabular}{@{}L{4.4cm}L{5.2cm}L{1.9cm}@{}}
\toprule
Claim (source) & Production evidence (ours) & Verdict \\
\midrule
``Expert pruning is a superior strategy for generative tasks'' (REAP, abstract); at 25\% on Qwen3-30B-A3B pruning and
M-SMoE are within about a point on generation (\S5) &
REAP better in \V{h-reap} of \V{h-total} comparisons; Qwen3-30B generative retention at 25\% \V{ret-q-reap25-gen}\% vs
\V{ret-q-ream25-gen}\% &
Confirmed and larger, also against REAM \\
\addlinespace
REAM beats REAP on generation at 25\%: GEN 69.8 vs 68.6, best of ten mixtures (REAM \S5.1) &
REAP ahead on 4 of 5 tasks at 25\% on Qwen3-30B, by up to \V{h-q25-reach} points &
Not seen in production \\
\addlinespace
At 50\%, REAP leads on GEN (62.8 vs 60.7), REAM on multiple-choice (REAM App.~A.3) &
REAP ahead on three tasks; REAM ahead only on brand names (detection) &
Partly confirmed \\
\addlinespace
Merging collapses expert functions (``functional subspace collapse'', REAP \S3.3) &
Not measured on the experts; REAM's outputs diverge further from the base model's (next row) &
Consistent; not tested directly \\
\addlinespace
Merged outputs have higher cross-perplexity and diverge faster with position (REAP \S5) &
Cross-perplexity higher for REAM in all \V{xppl-cells} cells; JSD in \V{jsd-ream-gt}; position trend not testable on
our short and HTML answers &
Level confirmed \\
\addlinespace
Merged outputs are ``more repetitive'' (REAP \S5) &
REAM has \V{r-loops-ratio-n50-reach}--\V{r-loops-ratio-n25-reach} times as many runaway email-template answers; distinct-2
without loops does not differ &
Confirmed via loops \\
\addlinespace
Rank of pre-logit embeddings predicts quality (REAM \S5.4) &
Effective rank $\tau=\V{r-tau-erank}$; numerical rank saturates &
Weak \\
\addlinespace
Group size $C$ ``is not heavily tuned'' (REAM App.~A.1) &
We use REAM's $C$ (16 at 25\%, 32 at 50\%) without tuning, also for Nemotron &
Not tested \\
\addlinespace
Code generation survives 50\% pruning (REAP: 97\% of Qwen3-30B-A3B's code score, \S5) &
REAP 50\% keeps \V{ret-q-reap50-gen}\% (Qwen3-30B) and \V{ret-n-reap50-gen}\% (Nemotron) of generative quality &
Not for these tasks \\
\bottomrule
\end{tabular}
\end{table}

Four properties of production tasks are consistent with the gaps, although we do not test them causally. Production
fails on the output contract (JSON, HTML gates, language rules), which benchmarks that parse leniently do not count. It
asks for generation in many languages, which neither paper evaluates. Its calibration comes from product traffic, chosen
without looking at the test metric. And long sampled answers give a compressed model many chances to fall into a loop:
up to \V{reach-capped-max-pct}\% of email-template answers at 50\% run to the cap.

\section{Discussion and recommendations}\label{sec:discussion}

Why does deleting beat merging here? REAP attributes merging's weakness to averaging experts that the router used for
different inputs~\citep{lasby2025reap}. REAM avoids summed gates, but it still replaces surviving experts with
saliency-weighted averages of up to $C$ experts, while pruning leaves every survivor intact. The damage concentrates
where exact behaviour matters (format rules, a translated template phrase, long structured output), and REAM's outputs
drift further from the base model by every measure. A smaller $C$, a capped group size, router
repair~\citep{hyeon2026is} or a per-expert choice between pruning and merging~\citep{huang2026prime} may narrow the gap.

For practitioners we draw four recommendations:
\begin{enumerate}
  \item \textbf{Prefer pruning.} REAP wins most comparisons and gives checkpoints of the same size as REAM.
  \item \textbf{Treat 25\% as the limit for generative products.} At 25\% REAP is close to lossless for Qwen3-30B
        outside generation. At 50\% even REAP loses \V{d-q-reap50-domains}\pp{} on domain descriptions and
        \V{d-q-reap50-reach} points on email templates.
  \item \textbf{Evaluate the output contract and every language, not only the mean.} JSON validity, format gates,
        untranslated template text and runaway generation explain most of the loss, and benchmark averages do not show
        them.
  \item \textbf{Cover the tokens that matter in calibration.} Answers are only \V{r-calib-ans-all}\% of the calibration
        tokens, yet the failures are in what the model generates. Weighting answer tokens, or calibrating on the
        model's own generations~\citep{lucas2026reasoning}, is a hypothesis worth testing.
\end{enumerate}

\section{Limitations}\label{sec:limitations}

Each configuration was compressed once, with one calibration set, so the CIs cover which rows were evaluated, not
compression randomness or the choice of calibration data, and the \V{h-total} head-to-head comparisons are correlated
comparisons among eight checkpoints rather than independent replications. We study two base models of one size class,
two ratios, and no other pruning or merging methods (frequency pruning, HC-SMoE, M-SMoE). REAM is our port, with
documented deviations including the group-size convention, and $C$ follows REAM's choice for Qwen3 without tuning for
Nemotron. References are judge-approved outputs of an earlier production model rather than human labels; two tasks are
scored by the same LLM judge that accepted their references, and the email-template rubric includes LLM judges. The
relaxed email-template gates are our decision (the relaxed unsubscribe rule also accepts placeholder links), and both scores are reported. The
brand task excludes rows without a brand, which rewards detection. Two tasks are sampled, as in production. The
calibration is English-heavy in text for every task and in rows for two, and the per-language cell intervals in
\figref{fig:languages} are uncorrected. We do not measure serving memory, latency or throughput. The
tasks are private product data, so the study cannot be repeated on the same rows.

\section{Conclusion}\label{sec:conclusion}

On production tasks, deleting experts beats merging them. With the same calibration data and nearly the same experts
kept, REAP is better than REAM in \V{h-reap} of \V{h-total} comparisons. The margins are widest on generation, where
the small benchmark difference between the methods does not carry over. The failures are mostly broken output
contracts: invalid JSON, broken layouts, runaway answers and untranslated template phrases. Compression at 25\% with
REAP is close to lossless for Qwen3-30B outside generation, with a quarter of the routed experts removed. Compressed MoE models should be evaluated per failure type and per
language, on the tasks they will serve.

\FloatBarrier

\section*{Data and code availability}
The datasets are private logs of Hostinger products and contain customer input, so they cannot be released. The code,
compressed models and per-row results are not released either. Sections~\ref{sec:methods} and~\ref{sec:setup} and the
appendices describe the methods, the REAM port and the evaluation in enough detail to repeat the study on other data.

\section*{Competing interests}
Both authors are employees of Hostinger, whose products provide the tasks and data. The study received no external
funding, and the authors have no other competing interests.

\section*{Use of generative AI}
An AI assistant was used as an aid in writing the analysis code and in drafting and editing the text. All numbers in the
paper are generated by scripts from the experiment results, and the authors reviewed the analyses and the text and take
responsibility for them.

\bibliographystyle{plainnat}
\bibliography{references}

\begin{thebibliography}{39}
\providecommand{\natexlab}[1]{#1}
\providecommand{\url}[1]{\texttt{#1}}
\expandafter\ifx\csname urlstyle\endcsname\relax
  \providecommand{\doi}[1]{doi: #1}\else
  \providecommand{\doi}{doi: \begingroup \urlstyle{rm}\Url}\fi

\bibitem[Ali et~al.(2026)Ali, Bibi, Moetesum, and Shafait]{ali2026what}
Ezza Ali, Sehrish Bibi, Momina Moetesum, and Faisal Shafait.
\newblock What we keep of what we forget: Replacement operators and routing
  semantics in training-free {MoE} expert pruning.
\newblock \emph{arXiv preprint
  \href{https://arxiv.org/abs/2610.41366}{arXiv:2610.41366}}, 2026.

\bibitem[Behdin et~al.(2025)Behdin, Fatahibaarzi, Song,
  et~al.]{behdin2025scaling}
Kayhan Behdin, Ata Fatahibaarzi, Qingquan Song, et~al.
\newblock Scaling down, serving fast: Compressing and deploying efficient
  {LLM}s for recommendation systems.
\newblock In \emph{Proceedings of the 2025 Conference on Empirical Methods in
  Natural Language Processing: Industry Track}, pages 1687--1702. Association
  for Computational Linguistics, 2025.
\newblock URL \url{https://aclanthology.org/2025.emnlp-industry.119/}.
\newblock arXiv:2502.14305.

\bibitem[Benazir et~al.(2026)Benazir, Chen, Qu, Huang, Li, and
  Lyu]{benazir2026benchmarking}
Afsara Benazir, Chen Chen, Rongxiao Qu, Jiabo Huang, Jingtao Li, and Lingjuan
  Lyu.
\newblock Benchmarking composable compression techniques in mixture-of-experts
  {LLM}s.
\newblock \emph{arXiv preprint
  \href{https://arxiv.org/abs/2608.21693}{arXiv:2608.21693}}, 2026.

\bibitem[Chen et~al.(2025)Chen, Liu, Sun, Chao, Hsu, and
  Lee]{chen2025retraining}
I-Chun Chen, Hsu-Shen Liu, Wei-Fang Sun, Chen-Hao Chao, Yen-Chang Hsu, and
  Chun-Yi Lee.
\newblock Retraining-free merging of sparse {MoE} via hierarchical clustering.
\newblock In \emph{Proceedings of the 42nd International Conference on Machine
  Learning}, volume 267 of \emph{Proceedings of Machine Learning Research},
  pages 8594--8620. PMLR, 2025.
\newblock URL \url{https://proceedings.mlr.press/v267/chen25aq.html}.
\newblock arXiv:2410.08589.

\bibitem[Chen et~al.(2026)Chen, Cai, Ji, Zhao, Zhang, Wang, and
  Chua]{chen2026understanding}
Yuxin Chen, Zhengzhou Cai, Xiangtian Ji, Weixiang Zhao, An~Zhang, Xiang Wang,
  and Tat-Seng Chua.
\newblock Understanding multilingualism in mixture-of-experts {LLM}s: Routing
  mechanism, expert specialization, and layerwise steering.
\newblock \emph{arXiv preprint
  \href{https://arxiv.org/abs/2601.14050}{arXiv:2601.14050}}, 2026.

\bibitem[Choenni and Titov(2025)]{choenni2025mwanda}
Rochelle Choenni and Ivan Titov.
\newblock {M}-{W}anda: Improving one-shot pruning for multilingual {LLM}s.
\newblock In \emph{Proceedings of the 2025 Conference on Empirical Methods in
  Natural Language Processing}, pages 26951--26964, Suzhou, China, 2025.
  Association for Computational Linguistics.
\newblock \doi{10.18653/v1/2025.emnlp-main.1369}.

\bibitem[Dao and Gu(2024)]{dao2024transformers}
Tri Dao and Albert Gu.
\newblock Transformers are {SSMs}: Generalized models and efficient algorithms
  through structured state space duality.
\newblock In \emph{Proceedings of the 41st International Conference on Machine
  Learning}, volume 235 of \emph{Proceedings of Machine Learning Research},
  pages 10041--10071, 2024.
\newblock URL \url{https://proceedings.mlr.press/v235/dao24a.html}.
\newblock arXiv:2405.21060.

\bibitem[Dong et~al.(2025)Dong, Ruan, Cai, Xu, Zhao, Lai, and
  Chen]{dong2025xgrammar}
Yixin Dong, Charlie~F. Ruan, Yaxing Cai, Ziyi Xu, Yilong Zhao, Ruihang Lai, and
  Tianqi Chen.
\newblock {XGrammar}: Flexible and efficient structured generation engine for
  large language models.
\newblock In \emph{Proceedings of Machine Learning and Systems 7 (MLSys 2025)},
  2025.
\newblock URL
  \url{https://proceedings.mlsys.org/paper_files/paper/2025/hash/5c20ca4b0b20b0bd2f1d839dc605e70f-Abstract-Conference.html}.
\newblock arXiv:2411.15100.

\bibitem[Dutta et~al.(2024)Dutta, Krishnan, Kwatra, and
  Ramjee]{dutta2024accuracy}
Abhinav Dutta, Sanjeev Krishnan, Nipun Kwatra, and Ramachandran Ramjee.
\newblock Accuracy is not all you need.
\newblock In \emph{Advances in Neural Information Processing Systems},
  volume~37, pages 124347--124390, 2024.
\newblock URL
  \url{https://proceedings.neurips.cc/paper_files/paper/2024/hash/e0e956681b04ac126679e8c7dd706b2e-Abstract-Conference.html}.
\newblock arXiv:2407.09141.

\bibitem[Efron and Tibshirani(1994)]{efron1993introduction}
Bradley Efron and Robert~J. Tibshirani.
\newblock \emph{An Introduction to the Bootstrap}.
\newblock Chapman and Hall/CRC, 1994.
\newblock \doi{10.1201/9780429246593}.

\bibitem[Haldar and Hockenmaier(2025)]{haldar2025rating}
Rajarshi Haldar and Julia Hockenmaier.
\newblock Rating roulette: Self-inconsistency in {LLM}-as-a-judge frameworks.
\newblock In \emph{Findings of the Association for Computational Linguistics:
  EMNLP 2025}, pages 24986--25004. Association for Computational Linguistics,
  2025.
\newblock URL \url{https://aclanthology.org/2025.findings-emnlp.1361/}.

\bibitem[Holm(1979)]{holm1979simple}
Sture Holm.
\newblock A simple sequentially rejective multiple test procedure.
\newblock \emph{Scandinavian Journal of Statistics}, 6\penalty0 (2):\penalty0
  65--70, 1979.

\bibitem[Huang et~al.(2026)Huang, Zhang, Fang, Xiao, and Du]{huang2026prime}
Shanfeng Huang, Rongqian Zhang, Zhou Fang, Song Xiao, and Hai Du.
\newblock {PRIME}: Frequency-aware unified pruning and merging for {MoE}
  compression.
\newblock \emph{arXiv preprint
  \href{https://arxiv.org/abs/2610.13982}{arXiv:2610.13982}}, 2026.

\bibitem[Hyeon and Do(2026)]{hyeon2026is}
Sieun Hyeon and Jaeyoung Do.
\newblock Is retraining-free enough? the necessity of router calibration for
  efficient {MoE} compression.
\newblock \emph{arXiv preprint
  \href{https://arxiv.org/abs/2603.02217}{arXiv:2603.02217}}, 2026.

\bibitem[Jha et~al.(2026)Jha, Hashemzadeh, Saheb~Pasand, Parviz, Lee, and
  Knyazev]{jha2026ream}
Saurav Jha, Maryam Hashemzadeh, Ali Saheb~Pasand, Ali Parviz, Min-Joong Lee,
  and Boris Knyazev.
\newblock {REAM}: Merging improves pruning of experts in {LLMs}.
\newblock \emph{arXiv preprint
  \href{https://arxiv.org/abs/2604.04356}{arXiv:2604.04356}}, 2026.

\bibitem[Ji et~al.(2025)Ji, Xiang, Li, Xia, Li, Duan, Wang, and
  Zhang]{ji2025beware}
Yixin Ji, Yang Xiang, Juntao Li, Qingrong Xia, Ping Li, Xinyu Duan, Zhefeng
  Wang, and Min Zhang.
\newblock Beware of calibration data for pruning large language models.
\newblock In \emph{The Thirteenth International Conference on Learning
  Representations (ICLR)}, 2025.
\newblock URL
  \url{https://proceedings.iclr.cc/paper_files/paper/2025/hash/2ede933e10afa991a10b6f36b6522129-Abstract-Conference.html}.
\newblock arXiv:2410.17711.

\bibitem[Kurz et~al.(2026)Kurz, Chen, Flek, and Zhao]{kurz2026limitations}
Simon Kurz, Jian-Jia Chen, Lucie Flek, and Zhixue Zhao.
\newblock On the limitations of language-targeted pruning: Investigating the
  calibration language impact in multilingual {LLM} pruning.
\newblock \emph{Transactions of the Association for Computational Linguistics},
  14:\penalty0 167--192, 2026.
\newblock \doi{10.1162/tacl.a.599}.

\bibitem[Kwon et~al.(2023)Kwon, Li, Zhuang, Sheng, Zheng, Yu, Gonzalez, Zhang,
  and Stoica]{kwon2023efficient}
Woosuk Kwon, Zhuohan Li, Siyuan Zhuang, Ying Sheng, Lianmin Zheng, Cody~Hao Yu,
  Joseph~E. Gonzalez, Hao Zhang, and Ion Stoica.
\newblock Efficient memory management for large language model serving with
  {PagedAttention}.
\newblock In \emph{Proceedings of the 29th Symposium on Operating Systems
  Principles (SOSP '23)}, pages 611--626. ACM, 2023.
\newblock \doi{10.1145/3600006.3613165}.

\bibitem[Lasby et~al.(2026)Lasby, Lazarevich, Sinnadurai, Lie, Ioannou, and
  Thangarasa]{lasby2025reap}
Mike Lasby, Ivan Lazarevich, Nish Sinnadurai, Sean Lie, Yani Ioannou, and
  Vithursan Thangarasa.
\newblock {REAP} the experts: Why pruning prevails for one-shot {MoE}
  compression.
\newblock In \emph{The Fourteenth International Conference on Learning
  Representations (ICLR)}, 2026.
\newblock URL \url{https://arxiv.org/abs/2510.13999}.
\newblock arXiv:2510.13999.

\bibitem[Lee et~al.(2026)Lee, Park, Jang, Ryu, Kim, Lee, Suh, and
  Kim]{lee2026focus}
Junyoung Lee, Sehyeon Park, Shinhyoung Jang, Seonha Ryu, Hojeong Kim, Hyunsei
  Lee, Il~Hong Suh, and Yeseong Kim.
\newblock {FOCUS} \& {RePAIR}: Mitigating text degeneration via token-level
  guidance for pruned large language models.
\newblock In \emph{International Conference on Machine Learning (ICML)}, 2026.
\newblock URL \url{https://icml.cc/virtual/2026/poster/66464}.
\newblock arXiv:2608.26676.

\bibitem[Li et~al.(2024)Li, Zhang, Yadav, Sung, Cheng, Bansal, and
  Chen]{li2024merge}
Pingzhi Li, Zhenyu Zhang, Prateek Yadav, Yi-Lin Sung, Yu~Cheng, Mohit Bansal,
  and Tianlong Chen.
\newblock Merge, then compress: Demystify efficient {SMoE} with hints from its
  routing policy.
\newblock In \emph{The Twelfth International Conference on Learning
  Representations (ICLR)}, 2024.
\newblock URL \url{https://arxiv.org/abs/2310.01334}.
\newblock arXiv:2310.01334.

\bibitem[Liu et~al.(2026)Liu, He, Jiang, Wang, Zhao, Zhang, and
  Lee]{liu2026fluxmoe}
Qingxiu Liu, Yongchao He, Runhan Jiang, Zion Wang, Bohan Zhao, Mi~Zhang, and
  Patrick P.~C. Lee.
\newblock {FluxMoE}: Decoupling expert residency for high-performance {MoE}
  serving.
\newblock \emph{arXiv preprint
  \href{https://arxiv.org/abs/2604.02715}{arXiv:2604.02715}}, 2026.

\bibitem[Lucas et~al.(2026)Lucas, Behdin, Wang, Song, Tang, and
  Mazumder]{lucas2026reasoning}
Ryan Lucas, Kayhan Behdin, Zhipeng Wang, Qingquan Song, Shao Tang, and Rahul
  Mazumder.
\newblock Reasoning models can be accurately pruned via chain-of-thought
  reconstruction.
\newblock In \emph{The Fourteenth International Conference on Learning
  Representations (ICLR)}, 2026.
\newblock URL
  \url{https://proceedings.iclr.cc/paper_files/paper/2026/hash/e70ffb3f05096e62b5077d8e1b62e668-Abstract-Conference.html}.
\newblock arXiv:2509.12464.

\bibitem[Marchisio et~al.(2024{\natexlab{a}})Marchisio, Dash, Chen, Aumiller,
  {\"U}st{\"u}n, Hooker, and Ruder]{marchisio2024how}
Kelly Marchisio, Saurabh Dash, Hongyu Chen, Dennis Aumiller, Ahmet
  {\"U}st{\"u}n, Sara Hooker, and Sebastian Ruder.
\newblock How does quantization affect multilingual {LLM}s?
\newblock In \emph{Findings of the Association for Computational Linguistics:
  EMNLP 2024}, pages 15928--15947. Association for Computational Linguistics,
  2024{\natexlab{a}}.
\newblock URL \url{https://aclanthology.org/2024.findings-emnlp.935/}.
\newblock arXiv:2407.03211.

\bibitem[Marchisio et~al.(2024{\natexlab{b}})Marchisio, Ko, B{\'e}rard, Dehaze,
  and Ruder]{marchisio2024understanding}
Kelly Marchisio, Wei-Yin Ko, Alexandre B{\'e}rard, Th{\'e}o Dehaze, and
  Sebastian Ruder.
\newblock Understanding and mitigating language confusion in {LLM}s.
\newblock In \emph{Proceedings of the 2024 Conference on Empirical Methods in
  Natural Language Processing}, pages 6653--6677, Miami, Florida, USA,
  2024{\natexlab{b}}. Association for Computational Linguistics.
\newblock \doi{10.18653/v1/2024.emnlp-main.380}.

\bibitem[McNemar(1947)]{mcnemar1947note}
Quinn McNemar.
\newblock Note on the sampling error of the difference between correlated
  proportions or percentages.
\newblock \emph{Psychometrika}, 12\penalty0 (2):\penalty0 153--157, 1947.
\newblock \doi{10.1007/BF02295996}.

\bibitem[Nie et~al.(2025)Nie, Schmid, and Sch{\"u}tze]{nie2025mechanistic}
Ercong Nie, Helmut Schmid, and Hinrich Sch{\"u}tze.
\newblock Mechanistic understanding and mitigation of language confusion in
  {E}nglish-centric large language models.
\newblock In \emph{Findings of the Association for Computational Linguistics:
  EMNLP 2025}, pages 690--706, Suzhou, China, 2025. Association for
  Computational Linguistics.
\newblock \doi{10.18653/v1/2025.findings-emnlp.37}.

\bibitem[Ning et~al.(2026)Ning, Chen, Liang, Zhang, and
  Fan]{ning2026acecompression}
Jiahui Ning, Liang Chen, Yuchen Liang, Yi~Zhang, and Fenglei Fan.
\newblock {ACE-Compression}: An ability-centric evaluation of compressed large
  language models.
\newblock \emph{arXiv preprint
  \href{https://arxiv.org/abs/2610.27913}{arXiv:2610.27913}}, 2026.

\bibitem[{NVIDIA}(2025)]{nvidia2025nemotron}
{NVIDIA}.
\newblock Nemotron 3 nano: Open, efficient mixture-of-experts hybrid
  {Mamba}-transformer model for agentic reasoning.
\newblock \emph{arXiv preprint
  \href{https://arxiv.org/abs/2512.20848}{arXiv:2512.20848}}, 2025.

\bibitem[{NVIDIA}(2026)]{nvidia2026nemotron}
{NVIDIA}.
\newblock Nemotron 3.5 super and lightning: Open, efficient mixture-of-experts
  hybrid {Mamba}-transformer models for agentic {AI}.
\newblock \emph{arXiv preprint
  \href{https://arxiv.org/abs/2610.38095}{arXiv:2610.38095}}, 2026.

\bibitem[Rasymas and Kvedaras(2026)]{rasymas2026english}
Tomas Rasymas and Povilas Kvedaras.
\newblock {E}nglish calibration erodes {L}ithuanian in expert-pruned
  mixture-of-experts language models.
\newblock \emph{arXiv preprint
  \href{https://arxiv.org/abs/2611.07654}{arXiv:2611.07654}}, 2026.

\bibitem[{Red Hat AI} and {vLLM Project}(2024)]{redhatai2024llm}
{Red Hat AI} and {vLLM Project}.
\newblock {LLM Compressor}, 2024.
\newblock URL \url{https://github.com/vllm-project/llm-compressor}.
\newblock Software.

\bibitem[Salim et~al.(2026{\natexlab{a}})Salim, Rinartha, Wu, Nhiem, Ku, and
  Li]{salim2026the}
Luis~Frentzen Salim, Komang Rinartha, Chia-Chun Wu, Tran~Van Nhiem, Lun-Wei Ku,
  and Yung-Hui Li.
\newblock The fragility and dynamics of {LLM} pruning-recovery steps.
\newblock \emph{arXiv preprint
  \href{https://arxiv.org/abs/2610.15574}{arXiv:2610.15574}},
  2026{\natexlab{a}}.

\bibitem[Salim et~al.(2026{\natexlab{b}})Salim, Wu, Van~Nhiem, Ku, and
  Li]{salim2026towards}
Luis~Frentzen Salim, Chia-Chun Wu, Tran Van~Nhiem, Lun-Wei Ku, and Yung-Hui Li.
\newblock Towards robust calibration dataset for domain-general mixture of
  experts pruning.
\newblock \emph{arXiv preprint
  \href{https://arxiv.org/abs/2610.15578}{arXiv:2610.15578}},
  2026{\natexlab{b}}.

\bibitem[Wen et~al.(2026)Wen, Sun, Liu, Xu, Cong, and Li]{wen2026the}
Rui Wen, Lu~Sun, Jiayang Liu, Zesheng Xu, Tianshuo Cong, and Zheng Li.
\newblock The benchmark illusion: Pruned {LLM}s can pass multiple choice but
  fail to answer.
\newblock In \emph{Proceedings of the 2026 Conference on Empirical Methods in
  Natural Language Processing}, pages 52565--52575. Association for
  Computational Linguistics, 2026.
\newblock URL \url{https://aclanthology.org/2026.emnlp-main.2313/}.
\newblock arXiv:2606.17609.

\bibitem[Xu et~al.(2026)Xu, Xing, Liu, Wang, Jiang, Huang, Chen, and
  Xu]{xu2026rethinking}
Hu~Xu, Zhaolong Xing, Congcong Liu, Jiaxing Wang, Zhida Jiang, Junshi Huang,
  Zhen Chen, and Jianfeng Xu.
\newblock Rethinking calibration for high-sparsity {LLM} pruning: From single
  sources to capability coverage.
\newblock \emph{arXiv preprint
  \href{https://arxiv.org/abs/2606.03328}{arXiv:2606.03328}}, 2026.

\bibitem[Yamaguchi et~al.(2026)Yamaguchi, Palucha, Bijar, Villavicencio, and
  Aletras]{yamaguchi2026on}
Atsuki Yamaguchi, Szymon Palucha, L{\'e}o Bijar, Aline Villavicencio, and
  Nikolaos Aletras.
\newblock On the utility and factual reliability of pruned mixture-of-experts
  models in the biomedical domain.
\newblock \emph{arXiv preprint
  \href{https://arxiv.org/abs/2607.01444}{arXiv:2607.01444}}, 2026.

\bibitem[Yang et~al.(2025)Yang, Li, Yang, et~al.]{qwen2025qwen3}
An~Yang, Anfeng Li, Baosong Yang, et~al.
\newblock Qwen3 technical report.
\newblock \emph{arXiv preprint
  \href{https://arxiv.org/abs/2505.09388}{arXiv:2505.09388}}, 2025.

\bibitem[Yu et~al.(2025)Yu, Ma, Agarwal, Oswald, Huang, Linsenmaier, Mei, Zhao,
  Borkar, Rouhani, Nellans, Krashinsky, and Khandelwal]{yu2025efficient}
Yanpeng Yu, Haiyue Ma, Krish Agarwal, Nicolai Oswald, Qijing Huang, Hugo
  Linsenmaier, Chunhui Mei, Ritchie Zhao, Ritika Borkar, Bita~Darvish Rouhani,
  David Nellans, Ronny Krashinsky, and Anurag Khandelwal.
\newblock Efficient {MoE} serving in the memory-bound regime: Balance activated
  experts, not tokens.
\newblock \emph{arXiv preprint
  \href{https://arxiv.org/abs/2512.09277}{arXiv:2512.09277}}, 2025.

\end{thebibliography}

\clearpage
\begin{appendix}
\sloppy
\section*{Appendices}
Conventions shared by the appendices:
\begin{itemize}
  \item changes are compressed $-$ base on the same rows, in percentage points (rubric points for email templates);
  \item REAP $-$ REAM is positive when REAP is better;
  \item \emph{significant} means Holm-corrected $p<0.05$;
  \item filled markers in figures are significant, hollow ones are not.
\end{itemize}

\FloatBarrier
\section{Dataset samples}\label{app:samples}

This appendix shows, for each task, what the model must return: a short description of the task, one reference answer
and how answers are scored. The prompts (system prompt and user message) describe internal systems and are not shown.
\begin{itemize}
  \item Each reference answer is the output of an earlier production model on a real validation row, accepted by the
        product's judge, and anonymised: every brand, company or person name, domain, email address, URL, phone number
        and street address is replaced by a fictitious one.
  \item Long parts are abridged with [\ldots].
\end{itemize}

\ifsamples{In the boxes, \texttt{[...]} marks omitted text.

\subsection*{Website-description cues}

\noindent\textbf{Task.} From a user's description of the website they want, the website builder decides whether the description names specific colours and whether it names specific fonts, and returns the two booleans as JSON.\\
\textbf{Row language.} English (\texttt{en}).

\begin{samplebox}{Reference answer}
\begin{lstlisting}
{
  "isColorMentioned": true,
  "isFontMentioned": false
}
\end{lstlisting}
\end{samplebox}

\noindent\textbf{Scoring.} Exact match: an answer is correct only if it is JSON with exactly the two keys and both booleans equal the reference, a production answer accepted by an LLM judge.

\subsection*{Brand-name extraction}

\noindent\textbf{Task.} Find the user's own brand name in a website description and return it exactly as written, with \texttt{source} set to \texttt{detected}; every row used here states a brand.\\
\textbf{Row language.} English (\texttt{en}).

\begin{samplebox}{Reference answer}
\begin{lstlisting}
{"brand_name":"Bella's Bakery","source":"detected"}
\end{lstlisting}
\end{samplebox}

\noindent\textbf{Scoring.} Exact match: an answer is correct only if \texttt{source} is \texttt{detected} and \texttt{brand\_name} is identical to the reference, including case and punctuation. As in production, answers are constrained to a JSON schema with exactly these two keys (structured output).

\subsection*{Product categorisation}

\noindent\textbf{Task.} Assign a new online-store product to a category: reuse one of the store's categories when one fits, otherwise propose a new generic two-word category.\\
\textbf{Row language.} English (\texttt{en}).

\begin{samplebox}{Reference answer}
\begin{lstlisting}
{"category": "Cards"}
\end{lstlisting}
\end{samplebox}

\noindent\textbf{Scoring.} LLM-judge pass rate: the production judge that accepted the reference sees the production request and the answer and returns pass or fail; a new generic category passes unless a listed category is clearly better. As in production, answers are constrained to a JSON schema with the single key \texttt{category} (structured output).

\subsection*{Domain descriptions}

\noindent\textbf{Task.} Say in a short marketing text what a domain name is good for, in the requested language, starting with ``This domain is'' and without repeating the domain name.\\
\textbf{Row language.} English (\texttt{en}).

\begin{samplebox}{Reference answer}
\begin{lstlisting}
This domain is suitable for floral and gifting services. It's ideal for an online flower shop.
\end{lstlisting}
\end{samplebox}

\noindent\textbf{Scoring.} LLM-judge pass rate: the production judge that accepted the reference checks the language, the required opening (a faithful translation counts in other languages), that the domain is not named, and the length, leniently.

\subsection*{Email template generation}

\noindent\textbf{Task.} Write a complete, responsive HTML template of a marketing email (here an educational newsletter for a coffee brand) from the user's chat request and the sender's brand information, returned as JSON with the keys \texttt{html}, \texttt{content} (a short chat reply to the user) and \texttt{template\_name}.\\
\textbf{Row language.} English (\texttt{en}).

\noindent The reference answer is abridged to its skeleton: the JSON keys, the opening of the \texttt{html} string, one content block, the footer with the unsubscribe link and the start of the \texttt{content} reply; long style attributes are shortened. The \texttt{html} string is shown unescaped, with its \texttt{\textbackslash n} escapes as line breaks and \texttt{\textbackslash"} as \texttt{"}.

\begin{samplebox}{Reference answer}
\begin{lstlisting}
{"html": "<!DOCTYPE html>
<html lang="en">
<head> [...] <title>The Impact of Your Cup</title> </head>
<body>
<div style="background-color: #4B2E2A; font-family: 'DM Sans', Arial, sans-serif; margin: 0; padding: 0;">
[...] header with logo, hero heading, first button, mission section [...]
<h3 style="margin: 0; font-size: 26px; color: #FFFFFF; font-weight: 700;">Guatemala Highlands</h3>
<p style="margin: 8px 0 20px 0; font-size: 16px; color: #FFFFFF; line-height: 24px;">
Medium Roast | $21.99 USD<br>
Smooth and balanced with a rich cocoa base, roasted almond, and a gentle cherry sweetness.
</p>
[...] product image and button, second product, image grid, closing call to action, social icons [...]
<p style="margin: 0;">Copper Kettle Coffee<br>123 Main St #100 Anytown, OH 43000<br>
<a href="https://example-coffee.com/privacy-policy" [...]>Privacy Policy</a></p>
<div id="reach-unsubscribe" class="unsubscribe active" style="margin-top: 20px;">
<a clicktracking='off' href="{{unsubscribe-link}}" [...]>Unsubscribe</a>
</div>
<div id="reach-signature" style="margin-top: 20px;">{{signature}}</div>
[...]
</body>
</html>",
"content": "This email template highlights Copper Kettle Coffee's commitment to direct trade and community impact, [...]",
"template_name": "Direct Trade Impact Newsletter"}
\end{lstlisting}
\end{samplebox}

\noindent\textbf{Scoring.} Production rubric (version 1.2), from 0 to 100. Five hard gates set the score to 0 when they fail: \texttt{output\_parseable}, \texttt{html\_present}, \texttt{html\_complete}, \texttt{unsubscribe\_link} and \texttt{mobile\_layout\_intact}. Otherwise the score combines 13 category scores (\texttt{output\_validity}, \texttt{client\_safety}, \texttt{structure}, \texttt{typography}, \texttt{spacing}, \texttt{required\_blocks}, \texttt{links\_media}, \texttt{conditional}, \texttt{content\_field}, \texttt{language}, \texttt{rendered}, \texttt{text\_judge}, \texttt{vision\_judge}), from deterministic checks, the email rendered in a browser, and a text and a vision LLM judge. The score used in this paper relaxes two gates (\texttt{html\_complete}, \texttt{unsubscribe\_link}); the production score with every gate is reported next to it.
\par}\else{%
  \textbf{\textcolor{red}{[dataset samples pending]}}}\fi

\FloatBarrier
\section{All comparisons}\label{app:full}

\tabref{tab:full} lists every compressed model's score, its change from the base model with the 95\% paired-bootstrap
CI, its retention with CI, and the Holm-corrected $p$-value. The tests are McNemar's exact test for pass/fail tasks and
the paired sign-flip permutation test for email templates. \tabref{tab:h2h} gives the head-to-head differences.

\begin{table}[p]
\caption{\textbf{Every comparison with the base model} (2,000 rows each). Change = compressed $-$ base with 95\%
paired-bootstrap CI; retention = compressed / base.}
\label{tab:full}
\centering\scriptsize
\resizebox{\textwidth}{!}{\begin{tabular}{@{}llllllr@{}}
\toprule
Task, model & Base & Variant & Score & Change [95\% CI] & Retention (\%) [95\% CI] & $p$ (Holm) \\
\midrule
Website cues, Qwen3-30B & 98.8 & REAP 25\% & 98.5 & $-$0.2 [$-$0.6, 0.0] & 99.7 [99.4, 100.1] & 1.000 \\
 &  & REAP 50\% & 97.8 & $-$1.0 [$-$1.5, $-$0.4] & 99.0 [98.5, 99.6] & 0.009 \\
 &  & REAM 25\% & 98.2 & $-$0.5 [$-$1.0, 0.0] & 99.5 [99.0, 100.0] & 0.529 \\
 &  & REAM 50\% & 97.1 & $-$1.7 [$-$2.4, $-$1.0] & 98.3 [97.6, 99.0] & $<$0.001 \\
\addlinespace
Website cues, Nemotron & 96.0 & REAP 25\% & 96.5 & +0.5 [0.0, 1.1] & 100.6 [100.1, 101.1] & 0.491 \\
 &  & REAP 50\% & 94.7 & $-$1.3 [$-$2.1, $-$0.5] & 98.6 [97.8, 99.5] & 0.020 \\
 &  & REAM 25\% & 95.5 & $-$0.5 [$-$1.2, 0.2] & 99.5 [98.8, 100.3] & 1.000 \\
 &  & REAM 50\% & 91.5 & $-$4.4 [$-$5.4, $-$3.4] & 95.4 [94.3, 96.4] & $<$0.001 \\
\addlinespace
Brand names, Qwen3-30B & 64.8 & REAP 25\% & 68.7 & +3.9 [2.7, 5.1] & 106.0 [104.1, 108.0] & $<$0.001 \\
 &  & REAP 50\% & 69.0 & +4.2 [2.6, 5.8] & 106.5 [103.9, 109.2] & $<$0.001 \\
 &  & REAM 25\% & 64.5 & $-$0.3 [$-$1.6, 1.0] & 99.5 [97.6, 101.6] & 1.000 \\
 &  & REAM 50\% & 70.7 & +5.9 [4.1, 7.6] & 109.0 [106.3, 111.9] & $<$0.001 \\
\addlinespace
Brand names, Nemotron & 49.8 & REAP 25\% & 45.8 & $-$4.0 [$-$5.5, $-$2.5] & 92.0 [89.1, 94.9] & $<$0.001 \\
 &  & REAP 50\% & 28.6 & $-$21.1 [$-$23.1, $-$19.1] & 57.6 [54.2, 60.9] & $<$0.001 \\
 &  & REAM 25\% & 39.8 & $-$10.0 [$-$11.8, $-$8.0] & 80.0 [76.6, 83.6] & $<$0.001 \\
 &  & REAM 50\% & 34.1 & $-$15.6 [$-$17.7, $-$13.6] & 68.5 [64.9, 72.2] & $<$0.001 \\
\addlinespace
Categorisation, Qwen3-30B & 81.7 & REAP 25\% & 80.7 & $-$1.0 [$-$2.2, 0.4] & 98.8 [97.3, 100.4] & 1.000 \\
 &  & REAP 50\% & 75.0 & $-$6.7 [$-$8.6, $-$4.9] & 91.8 [89.6, 94.0] & $<$0.001 \\
 &  & REAM 25\% & 76.7 & $-$5.0 [$-$6.5, $-$3.3] & 93.9 [92.0, 95.9] & $<$0.001 \\
 &  & REAM 50\% & 71.2 & $-$10.4 [$-$12.5, $-$8.4] & 87.3 [84.9, 89.6] & $<$0.001 \\
\addlinespace
Categorisation, Nemotron & 68.2 & REAP 25\% & 65.2 & $-$3.0 [$-$4.7, $-$1.4] & 95.6 [93.2, 97.9] & 0.004 \\
 &  & REAP 50\% & 56.9 & $-$11.3 [$-$13.4, $-$9.2] & 83.4 [80.6, 86.3] & $<$0.001 \\
 &  & REAM 25\% & 59.9 & $-$8.3 [$-$10.3, $-$6.2] & 87.8 [85.0, 90.7] & $<$0.001 \\
 &  & REAM 50\% & 49.9 & $-$18.3 [$-$20.5, $-$16.0] & 73.1 [70.2, 76.2] & $<$0.001 \\
\addlinespace
Domain descr., Qwen3-30B & 93.5 & REAP 25\% & 80.5 & $-$13.0 [$-$14.7, $-$11.5] & 86.1 [84.4, 87.7] & $<$0.001 \\
 &  & REAP 50\% & 71.3 & $-$22.2 [$-$24.3, $-$20.2] & 76.2 [74.1, 78.3] & $<$0.001 \\
 &  & REAM 25\% & 62.4 & $-$31.1 [$-$33.3, $-$29.0] & 66.7 [64.5, 68.9] & $<$0.001 \\
 &  & REAM 50\% & 63.5 & $-$30.1 [$-$32.3, $-$27.9] & 67.8 [65.6, 70.1] & $<$0.001 \\
\addlinespace
Domain descr., Nemotron & 58.8 & REAP 25\% & 52.9 & $-$5.9 [$-$8.1, $-$3.5] & 90.0 [86.4, 93.9] & $<$0.001 \\
 &  & REAP 50\% & 36.8 & $-$22.0 [$-$24.4, $-$19.4] & 62.6 [59.1, 66.3] & $<$0.001 \\
 &  & REAM 25\% & 40.0 & $-$18.8 [$-$21.3, $-$16.2] & 68.0 [64.4, 72.0] & $<$0.001 \\
 &  & REAM 50\% & 31.7 & $-$27.1 [$-$29.6, $-$24.4] & 54.0 [50.5, 57.7] & $<$0.001 \\
\addlinespace
Email templates, Qwen3-30B & 70.2 & REAP 25\% & 71.1 & +0.9 [$-$0.5, 2.3] & 101.2 [99.3, 103.3] & 1.000 \\
 &  & REAP 50\% & 45.3 & $-$25.0 [$-$26.8, $-$23.1] & 64.4 [62.0, 66.9] & 0.002 \\
 &  & REAM 25\% & 40.9 & $-$29.3 [$-$31.2, $-$27.4] & 58.3 [55.8, 60.8] & 0.002 \\
 &  & REAM 50\% & 25.1 & $-$45.1 [$-$46.9, $-$43.3] & 35.8 [33.6, 37.9] & 0.002 \\
\addlinespace
Email templates, Nemotron & 12.8 & REAP 25\% & 13.0 & +0.2 [$-$1.4, 1.8] & 101.8 [89.8, 115.4] & 1.000 \\
 &  & REAP 50\% & 2.9 & $-$9.8 [$-$11.2, $-$8.5] & 23.0 [18.2, 28.5] & 0.002 \\
 &  & REAM 25\% & 4.5 & $-$8.3 [$-$9.7, $-$6.9] & 35.2 [29.0, 42.2] & 0.002 \\
 &  & REAM 50\% & 0.4 & $-$12.4 [$-$13.7, $-$11.2] & 3.0 [1.4, 4.8] & 0.002 \\
\bottomrule
\end{tabular}
}
\end{table}

\begin{table}[!htbp]
\caption{\textbf{REAP $-$ REAM on the same rows,} with 95\% paired-bootstrap CIs. \textbf{Bold}: REAP significantly
better; \textit{italic}$^\dagger$: REAM significantly better (Holm over the 20 comparisons).}
\label{tab:h2h}
\centering\small
\resizebox{\textwidth}{!}{\begin{tabular}{@{}lllll@{}}
\toprule
 & \multicolumn{2}{c}{Qwen3-30B} & \multicolumn{2}{c}{Nemotron} \\
\cmidrule(lr){2-3}\cmidrule(l){4-5}
Task & 25\% & 50\% & 25\% & 50\% \\
\midrule
Website cues & +0.2 {\scriptsize[$-$0.2, 0.8]} & +0.7 {\scriptsize[0.0, 1.4]} & \textbf{+1.0 {\scriptsize[0.3, 1.7]}} & \textbf{+3.1 {\scriptsize[2.2, 4.1]}} \\
Brand names & \textbf{+4.2 {\scriptsize[3.0, 5.5]}} & \textit{$-$1.7 {\scriptsize[$-$2.9, $-$0.4]}}$^\dagger$ & \textbf{+6.0 {\scriptsize[4.3, 7.6]}} & \textit{$-$5.5 {\scriptsize[$-$7.2, $-$3.6]}}$^\dagger$ \\
Categorisation & \textbf{+4.0 {\scriptsize[2.5, 5.5]}} & \textbf{+3.7 {\scriptsize[1.9, 5.5]}} & \textbf{+5.3 {\scriptsize[3.4, 7.2]}} & \textbf{+7.0 {\scriptsize[5.0, 9.0]}} \\
Domain descr. & \textbf{+18.1 {\scriptsize[15.9, 20.3]}} & \textbf{+7.8 {\scriptsize[5.7, 9.9]}} & \textbf{+12.9 {\scriptsize[10.3, 15.6]}} & \textbf{+5.1 {\scriptsize[2.7, 7.3]}} \\
Email templates & \textbf{+30.2 {\scriptsize[28.4, 32.0]}} & \textbf{+20.1 {\scriptsize[18.0, 22.2]}} & \textbf{+8.5 {\scriptsize[7.2, 9.9]}} & \textbf{+2.6 {\scriptsize[2.0, 3.2]}} \\
\bottomrule
\end{tabular}
}
\end{table}

\begin{figure}[!htbp]
\centering
\includegraphics[width=\textwidth]{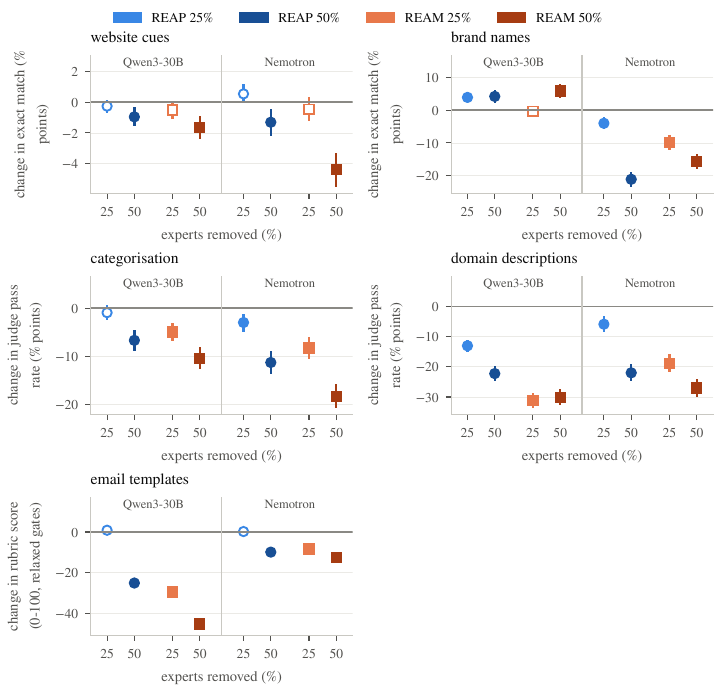}
\caption{\textbf{REAM is below REAP at the same ratio in almost every task and model.} Change from the base model in
each task's metric (named on the y-axis; negative = worse), for Qwen3-30B (left half of each panel) and Nemotron (right
half), REAP (circles) and REAM (squares) at 25\% and 50\%. Bars are 95\% paired-bootstrap CIs; filled markers are
significant after Holm correction, hollow ones are not.}
\label{fig:deltas}
\end{figure}

\FloatBarrier
\section{Brand names: detection and precision}\label{app:secondary}

Every validation row of the brand task states a brand, so an answer can fail in two ways: it returns no brand (a
detection failure), or it returns the wrong one (a precision failure). \tabref{tab:brand} splits each change in exact
match into these two parts.

\begin{table}[!htbp]
\caption{\textbf{Brand names: detection versus precision.} Detection = answers that return a brand; precision = exact
matches among those. The change in exact match from the base model splits into a detection part and a precision part,
which sum to the total. Every validation row states a brand.}
\label{tab:brand}
\centering\small
\resizebox{\textwidth}{!}{\setlength{\tabcolsep}{3.5pt}
\begin{tabular}{@{}lrrrlrrr@{}}
\toprule
 & Exact & Detec- & \multicolumn{2}{c}{Precision (\%)} & \multicolumn{3}{c}{Change in exact (points)} \\
\cmidrule(lr){4-5}\cmidrule(l){6-8}
Model & (\%) & tion (\%) & & 95\% CI & total & detect. & precis. \\
\midrule
Qwen3-30B & 64.8 & 87.2 & 74.3 & [72.3, 76.3] & -- & -- & -- \\
\quad REAP 25\% & 68.7 & 91.0 & 75.5 & [73.5, 77.4] & +3.9 & +2.8 & +1.1 \\
\quad REAP 50\% & 69.0 & 91.5 & 75.5 & [73.5, 77.4] & +4.2 & +3.2 & +1.0 \\
\quad REAM 25\% & 64.5 & 88.2 & 73.1 & [71.1, 75.2] & $-$0.3 & +0.7 & $-$1.0 \\
\quad REAM 50\% & 70.7 & 95.2 & 74.3 & [72.2, 76.2] & +5.9 & +5.9 & $-$0.1 \\
\midrule
Nemotron & 49.8 & 78.5 & 63.4 & [61.1, 65.8] & -- & -- & -- \\
\quad REAP 25\% & 45.8 & 72.0 & 63.5 & [61.0, 66.0] & $-$4.0 & $-$4.1 & +0.1 \\
\quad REAP 50\% & 28.6 & 54.5 & 52.6 & [49.7, 55.5] & $-$21.1 & $-$13.9 & $-$7.2 \\
\quad REAM 25\% & 39.8 & 61.6 & 64.6 & [61.9, 67.3] & $-$9.9 & $-$10.8 & +0.8 \\
\quad REAM 50\% & 34.1 & 53.5 & 63.7 & [60.8, 66.6] & $-$15.6 & $-$15.9 & +0.2 \\
\bottomrule
\end{tabular}
}
\end{table}

\FloatBarrier
\section{Email templates: rubric, gates and production scores}\label{app:reach}

\paragraph{The rubric.} The production rubric (version 1.2) scores an answer in 13 categories with weighted checks:
output validity, client safety, structure, typography, spacing, required blocks, links and media, conditional content,
content fields, language, the rendered page, a text judge and a vision judge. The rendered checks load the email in a
headless browser at desktop and mobile width.

\paragraph{The gates.} Five checks are hard gates:
\begin{itemize}
  \item \emph{output parseable}: the answer is valid JSON with the required keys;
  \item \emph{HTML present};
  \item \emph{HTML complete};
  \item \emph{unsubscribe link};
  \item \emph{mobile layout intact}: the rendered page does not overflow a narrow screen.
\end{itemize}

\paragraph{The relaxed rules.} We relax two gates:
\begin{itemize}
  \item \texttt{html\_complete} passes a complete \texttt{<body>\ldots</body>} without the \texttt{<html>} wrapper. A
        bare \texttt{<table>} fragment still fails.
  \item \texttt{unsubscribe\_link} passes any unsubscribe link in any of the dataset's languages, for example
        \texttt{<a href="\#">Unsubscribe</a>}, not only the template placeholder.
\end{itemize}
Nothing else changes, and all 2,000 production references are full documents.

\begin{figure}[!htbp]
\centering
\includegraphics[width=\textwidth]{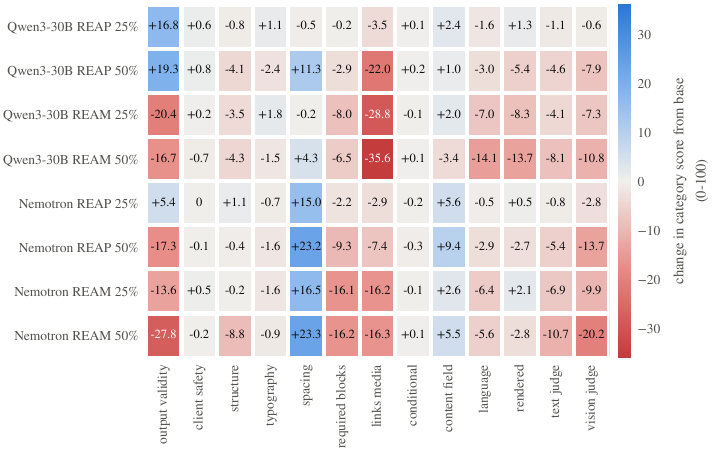}
\caption{\textbf{Merging hurts links and media, language and the rendered page most.} Change of each rubric category
score from the base model (points of 100; red = worse). Categories use the production checks, so \emph{output validity}
includes the \texttt{<html>} wrapper; that is why it rises for Qwen3-30B REAP.}
\label{fig:categories}
\end{figure}

\paragraph{Results.}
\begin{itemize}
  \item \textbf{The production score orders the models by a formatting habit.} Under the production rules Qwen3-30B
        scores \V{reach-prod-q-base}, REAP 25\% \V{reach-prod-q-reap25} and REAM 25\% \V{reach-prod-q-ream25}. That
        order follows how often each model omits the \texttt{<html>} wrapper, not template quality
        (\tabref{tab:gates}).
  \item \textbf{REAP still leads under the production rules,} in all four comparisons (\tabref{tab:h2h-prod}).
  \item \textbf{Per category, merging hurts content most:} links and media, language, the rendered page and the vision
        judge (\figref{fig:categories}).
\end{itemize}

\begin{table}[!htbp]
\caption{\textbf{Email templates: answers failing each gate} (\%). The relaxed rules are the ones scored; the last two
columns apply the production rules to the two relaxed gates.}
\label{tab:gates}
\centering\small
\resizebox{\textwidth}{!}{\begin{tabular}{@{}l rrrrr rr@{}}
\toprule
 & \multicolumn{5}{c}{Relaxed rules (as scored)} & \multicolumn{2}{c}{Production rules} \\
\cmidrule(lr){2-6}\cmidrule(l){7-8}
Model & parseable & HTML & complete & unsubscribe & mobile & complete & unsubscribe \\
\midrule
Qwen3-30B & 1.1 & 0.0 & 0.1 & 0.1 & 10.0 & 61.5 & 0.2 \\
\quad REAP 25\% & 1.1 & 0.0 & 0.1 & 0.1 & 8.8 & 28.0 & 0.1 \\
\quad REAP 50\% & 5.4 & 0.1 & 0.4 & 0.1 & 35.1 & 14.3 & 0.2 \\
\quad REAM 25\% & 5.1 & 0.0 & 0.1 & 0.4 & 40.1 & 94.3 & 0.8 \\
\quad REAM 50\% & 12.2 & 0.0 & 2.6 & 0.4 & 51.7 & 82.5 & 0.5 \\
\midrule
Nemotron & 27.2 & 0.1 & 34.9 & 35.0 & 13.9 & 35.3 & 36.9 \\
\quad REAP 25\% & 22.8 & 0.2 & 32.3 & 40.3 & 15.2 & 32.8 & 42.5 \\
\quad REAP 50\% & 47.8 & 0.1 & 27.9 & 36.4 & 16.2 & 28.1 & 37.5 \\
\quad REAM 25\% & 34.0 & 0.4 & 30.2 & 54.1 & 10.4 & 44.9 & 54.9 \\
\quad REAM 50\% & 46.7 & 0.1 & 46.4 & 43.4 & 15.2 & 49.5 & 44.9 \\
\bottomrule
\end{tabular}
}
\end{table}

\begin{table}[!htbp]
\caption{\textbf{REAP $-$ REAM on email templates under the production and the relaxed rules} (rubric points, 95\%
paired-bootstrap CI; production: Holm over these four comparisons).}
\label{tab:h2h-prod}
\centering\small
\resizebox{\textwidth}{!}{\setlength{\tabcolsep}{4pt}
\begin{tabular}{@{}lrrrlrr@{}}
\toprule
 & \multicolumn{5}{c}{Production gates} & Relaxed \\
\cmidrule(lr){2-6}\cmidrule(l){7-7}
Model & REAP & REAM & diff. & 95\% CI & Holm $p$ & diff. \\
\midrule
Qwen3-30B 25\% & 51.3 & 0.4 & \textbf{+51.0} & [+49.3, +52.6] & $<$0.001 & \textbf{+30.2} \\
Qwen3-30B 50\% & 37.7 & 1.9 & \textbf{+35.8} & [+34.0, +37.5] & $<$0.001 & \textbf{+20.1} \\
Nemotron 25\% & 12.1 & 2.9 & \textbf{+9.2} & [+7.9, +10.4] & $<$0.001 & \textbf{+8.5} \\
Nemotron 50\% & 2.6 & 0.1 & \textbf{+2.5} & [+1.9, +3.1] & $<$0.001 & \textbf{+2.6} \\
\bottomrule
\end{tabular}
}
\end{table}

\begin{figure}[!htbp]
\centering
\includegraphics[width=\textwidth]{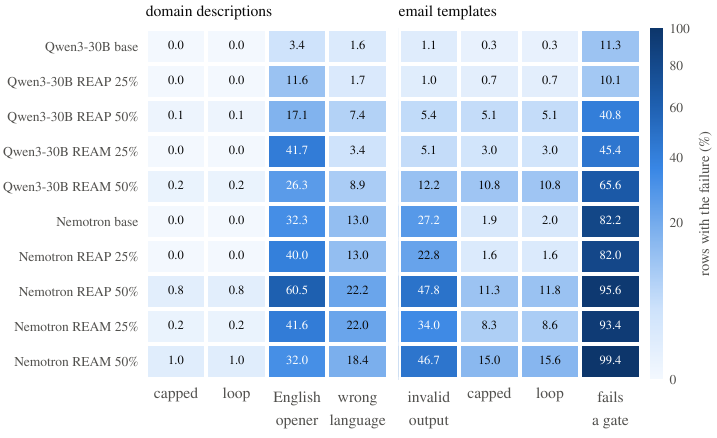}
\caption{\textbf{Failures concentrate in the output contract of the generative tasks.} Share of the 2,000 rows with each
failure (colour scale, square-root scaled). Domain descriptions: answers that ran to the 32k-token cap, repetition
loops, non-English answers that open in English, and answers in the wrong language. Email templates: invalid JSON,
capped answers, loops, and answers failing at least one hard gate. The three discriminative tasks are left out: every
rate there is below 1\%.}
\label{fig:failures}
\end{figure}

\FloatBarrier
\section{Languages}\label{app:languages}

\tabref{tab:lang} tests, for every task and compressed model, whether the change from the base model differs between
English and non-English rows.
\begin{itemize}
  \item \textbf{The gap} = non-English change $-$ English change, with a 95\% bootstrap CI (rows resampled within each
        group).
  \item \textbf{Correction:} Holm over the 40 cells.
\end{itemize}

\begin{figure}[!htbp]
\centering
\includegraphics[width=\textwidth]{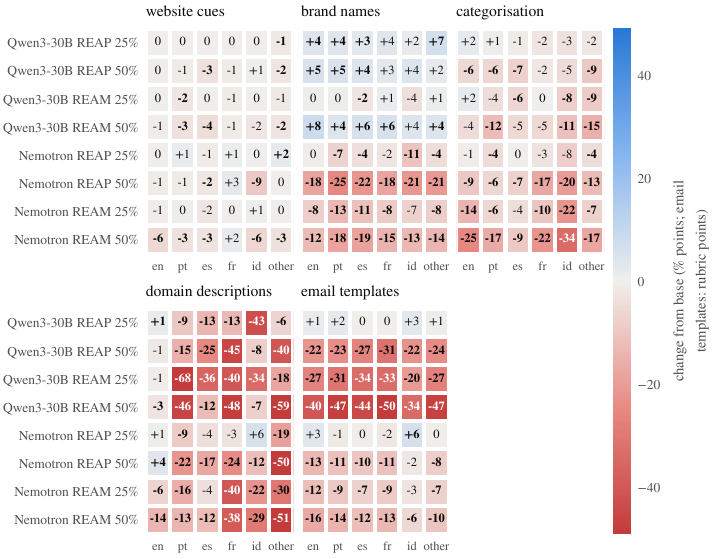}
\caption{\textbf{Only domain descriptions show a consistent English advantage.} Change from the base model per language
group (columns; ``other'' pools the remaining languages) for each compressed model (rows) and task (panels), in pp
(rubric points for email templates); red = worse, blue = better. Bold: the cell's 95\% bootstrap CI excludes 0,
uncorrected; the corrected tests are in \tabref{tab:lang}.}
\label{fig:languages}
\end{figure}

\begin{table}[p]
\caption{\textbf{English versus non-English change} per task and compressed model (pp; rubric points for email
templates). \textbf{Bold}: Holm-significant over the 40 cells.}
\label{tab:lang}
\centering\scriptsize
\resizebox{\textwidth}{!}{\setlength{\tabcolsep}{4pt}
\begin{tabular}{@{}lrrlr@{}}
\toprule
 & \multicolumn{2}{c}{Change from base} & Gap: non-En.\ $-$ En. & \\
\cmidrule(lr){2-3}
Task, model & English & non-English & [95\% CI] & Holm $p$ \\
\midrule
\multicolumn{5}{@{}l}{\textbf{Website cues}} \\
\quad Qwen3-30B REAP 25\% & $-$0.1 & $-$0.4 & $-$0.3 [$-$1.0, +0.3] & 1 \\
\quad Qwen3-30B REAP 50\% & $-$0.5 & $-$1.5 & $-$1.0 [$-$2.1, +0.1] & 1 \\
\quad Qwen3-30B REAM 25\% & 0.0 & $-$1.1 & $-$1.1 [$-$2.1, $-$0.1] & 0.78 \\
\quad Qwen3-30B REAM 50\% & $-$0.7 & $-$2.7 & $-$2.1 [$-$3.5, $-$0.7] & 0.068 \\
\quad Nemotron REAP 25\% & +0.3 & +0.8 & +0.6 [$-$0.5, +1.6] & 1 \\
\quad Nemotron REAP 50\% & $-$1.0 & $-$1.6 & $-$0.5 [$-$2.2, +1.1] & 1 \\
\quad Nemotron REAM 25\% & $-$0.7 & $-$0.2 & +0.5 [$-$1.0, +1.9] & 1 \\
\quad Nemotron REAM 50\% & $-$5.7 & $-$3.0 & +2.7 [+0.8, +4.7] & 0.2 \\
\midrule
\multicolumn{5}{@{}l}{\textbf{Brand names}} \\
\quad Qwen3-30B REAP 25\% & +4.2 & +3.8 & $-$0.3 [$-$3.3, +2.5] & 1 \\
\quad Qwen3-30B REAP 50\% & +5.0 & +3.9 & $-$1.0 [$-$5.1, +3.0] & 1 \\
\quad Qwen3-30B REAM 25\% & +0.4 & $-$0.5 & $-$0.9 [$-$4.0, +2.1] & 1 \\
\quad Qwen3-30B REAM 50\% & +8.3 & +5.1 & $-$3.2 [$-$7.3, +0.8] & 1 \\
\quad Nemotron REAP 25\% & $-$0.4 & $-$5.1 & $-$4.7 [$-$8.1, $-$1.3] & 0.24 \\
\quad Nemotron REAP 50\% & $-$17.7 & $-$22.2 & $-$4.5 [$-$9.1, $-$0.1] & 0.94 \\
\quad Nemotron REAM 25\% & $-$7.9 & $-$10.6 & $-$2.7 [$-$7.0, +1.7] & 1 \\
\quad Nemotron REAM 50\% & $-$12.5 & $-$16.7 & $-$4.2 [$-$8.8, +0.5] & 1 \\
\midrule
\multicolumn{5}{@{}l}{\textbf{Categorisation}} \\
\quad Qwen3-30B REAP 25\% & +2.2 & $-$1.4 & $-$3.7 [$-$7.2, $-$0.2] & 0.79 \\
\quad Qwen3-30B REAP 50\% & $-$5.5 & $-$6.9 & $-$1.4 [$-$6.5, +3.9] & 1 \\
\quad Qwen3-30B REAM 25\% & +2.2 & $-$6.1 & \textbf{$-$8.3 [$-$12.1, $-$4.5]} & 0.008 \\
\quad Qwen3-30B REAM 50\% & $-$4.4 & $-$11.3 & $-$6.9 [$-$12.0, $-$1.8] & 0.23 \\
\quad Nemotron REAP 25\% & $-$0.7 & $-$3.4 & $-$2.6 [$-$6.8, +1.5] & 1 \\
\quad Nemotron REAP 50\% & $-$9.2 & $-$11.6 & $-$2.4 [$-$7.9, +3.2] & 1 \\
\quad Nemotron REAM 25\% & $-$13.6 & $-$7.5 & +6.1 [+0.4, +11.9] & 0.78 \\
\quad Nemotron REAM 50\% & $-$25.0 & $-$17.2 & +7.8 [+1.5, +14.3] & 0.34 \\
\midrule
\multicolumn{5}{@{}l}{\textbf{Domain descr.}} \\
\quad Qwen3-30B REAP 25\% & +1.2 & $-$16.0 & \textbf{$-$17.2 [$-$19.5, $-$14.9]} & 0.008 \\
\quad Qwen3-30B REAP 50\% & $-$1.5 & $-$26.5 & \textbf{$-$25.1 [$-$28.3, $-$21.9]} & 0.008 \\
\quad Qwen3-30B REAM 25\% & $-$0.6 & $-$37.5 & \textbf{$-$36.9 [$-$39.9, $-$33.8]} & 0.008 \\
\quad Qwen3-30B REAM 50\% & $-$2.6 & $-$35.8 & \textbf{$-$33.1 [$-$36.6, $-$29.6]} & 0.008 \\
\quad Nemotron REAP 25\% & +0.6 & $-$7.2 & \textbf{$-$7.8 [$-$12.4, $-$3.3]} & 0.008 \\
\quad Nemotron REAP 50\% & +3.5 & $-$27.3 & \textbf{$-$30.8 [$-$35.0, $-$26.5]} & 0.008 \\
\quad Nemotron REAM 25\% & $-$6.4 & $-$21.4 & \textbf{$-$14.9 [$-$20.2, $-$9.6]} & 0.008 \\
\quad Nemotron REAM 50\% & $-$13.7 & $-$29.8 & \textbf{$-$16.1 [$-$21.6, $-$10.4]} & 0.008 \\
\midrule
\multicolumn{5}{@{}l}{\textbf{Email templates}} \\
\quad Qwen3-30B REAP 25\% & +1.5 & +0.8 & $-$0.7 [$-$4.6, +3.1] & 1 \\
\quad Qwen3-30B REAP 50\% & $-$21.8 & $-$25.6 & $-$3.9 [$-$9.1, +1.3] & 1 \\
\quad Qwen3-30B REAM 25\% & $-$26.7 & $-$29.9 & $-$3.2 [$-$8.6, +2.2] & 1 \\
\quad Qwen3-30B REAM 50\% & $-$40.2 & $-$46.1 & $-$6.0 [$-$11.2, $-$0.9] & 0.54 \\
\quad Nemotron REAP 25\% & +3.1 & $-$0.4 & $-$3.4 [$-$8.2, +1.2] & 1 \\
\quad Nemotron REAP 50\% & $-$13.2 & $-$9.1 & +4.1 [+0.4, +8.0] & 0.77 \\
\quad Nemotron REAM 25\% & $-$11.5 & $-$7.6 & +3.9 [0.0, +7.9] & 0.94 \\
\quad Nemotron REAM 50\% & $-$16.3 & $-$11.6 & +4.7 [+1.2, +8.3] & 0.22 \\
\bottomrule
\end{tabular}
}
\end{table}

\FloatBarrier
\section{Agreement with the base model}\label{app:agreement}

\begin{figure}[!htbp]
\centering
\includegraphics[width=\textwidth]{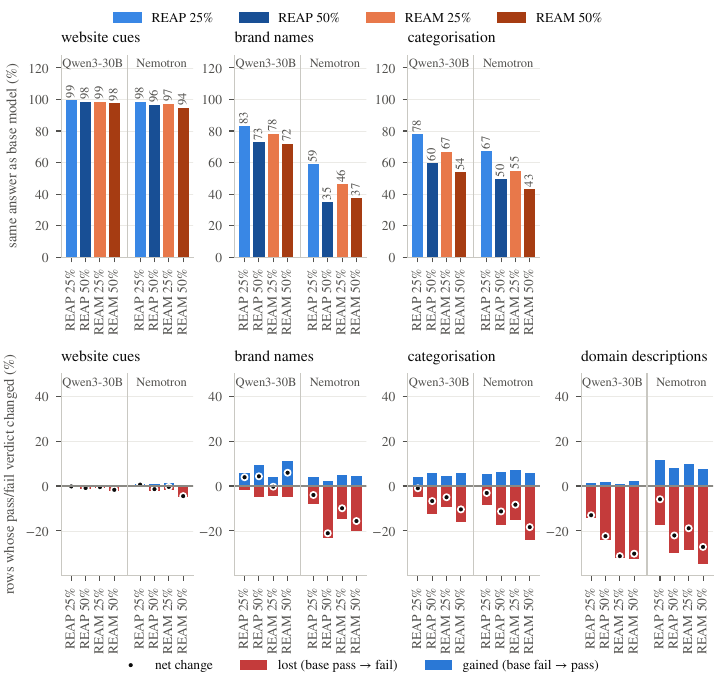}
\caption{\textbf{Small net changes hide many flipped answers.} Top: share of rows where a compressed model gives exactly
the base model's answer (greedy tasks; categorisation compares the normalised category). Bottom: rows whose pass/fail
verdict changed. Red bars below zero are rows lost (the base model passes, the compressed model fails); blue bars above
zero are rows gained; the dot is the net change.}
\label{fig:agreement}
\end{figure}

A small net change can hide many answers that changed. \figref{fig:agreement} counts, for every compressed model,
how often it gives exactly the base model's answer, and how many rows it newly fails or newly passes.

\FloatBarrier
\section{Diagnostics from the benchmark papers}\label{app:diagnostics}

\paragraph{Definitions.}
\begin{itemize}
  \item \textbf{Cross-perplexity} is the base model's perplexity of a compressed model's answers, token-weighted over
        all answers of a task. Answers that ran to the 32k-token cap are excluded.
  \item \textbf{Own perplexity} is each model's perplexity of its own answers.
  \item \textbf{Completion-logit JSD} is the Jensen--Shannon divergence (bits) between the compressed and the base
        model's next-token distributions, along the compressed model's answers on 200 rows per task.
    \begin{itemize}
      \item Each distribution is restricted to the compressed model's top-20 tokens plus a remainder bucket, so the
            value is a lower bound.
      \item For brand names and categorisation the first answer token is excluded from all log-probability measures,
            because the schema forces it.
    \end{itemize}
  \item \textbf{Distinct-2} is the share of unique bigrams: \emph{within} averages over answers, \emph{across} pools
        all of a model's answers. Repetition loops are excluded unless stated otherwise.
  \item \textbf{Loop rate} is the share of answers that end in a repetition loop. A loop is an answer that hit the
        token cap, or one whose last 3,000 characters compress below 10\%.
  \item \textbf{Embedding rank} uses the final normalised hidden states at 64 evenly spaced positions of 200
        prompt-plus-reference rows per task (12,800 embeddings).
    \begin{itemize}
      \item The numerical rank counts singular values above $s_{\max}\cdot\max(n,d)\cdot\epsilon_{\text{fp32}}$.
      \item The effective rank is the exponential of the entropy of the normalised singular values, after
            mean-centring.
    \end{itemize}
\end{itemize}

\begin{figure}[!htbp]
\centering
\includegraphics[width=\textwidth]{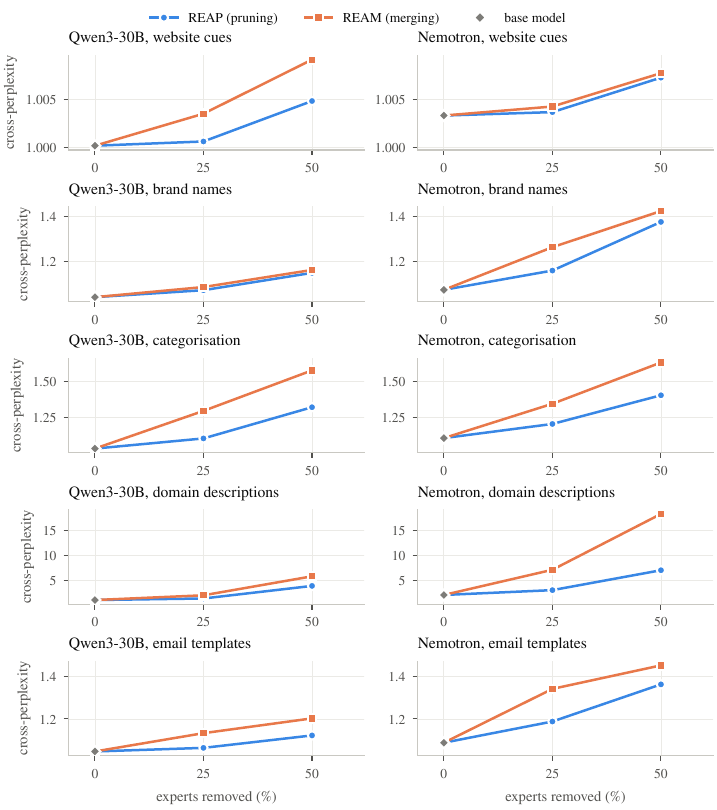}
\caption{\textbf{REAM's answers are less likely under the base model than REAP's in every task.} Cross-perplexity
(y-axis) against the share of experts removed, one row per task and one column per base model. For the base model the
value is its perplexity of its own answers.}
\label{fig:xppl}
\end{figure}

\paragraph{Scoring the diagnostics.} \tabref{tab:diag} scores each diagnostic in two ways:
\begin{itemize}
  \item the Kendall $\tau$ between the diagnostic, signed so that higher is better, and the primary metric, over the
        variants of each task and base model;
  \item over matched cells, how often the diagnostic ranks REAP and REAM the same way as quality does.
\end{itemize}
\tabref{tab:xppl} gives cross-perplexity and JSD per task and \tabref{tab:divrank} gives distinct-2 and effective rank.
\figref{fig:xppl} plots cross-perplexity against the share of experts removed.

\begin{table}[!htbp]
\caption{\textbf{Cross-perplexity, logit divergence and loops track production quality.} Mean Kendall $\tau$ between each
diagnostic (signed so that higher = better) and the primary metric over the variants of each task and base model, with
the number of positive cells; and, over matched (task, model, ratio) cells, how often the diagnostic favours REAP, how
often quality agrees, and how often both order REAP and REAM the same way. $^a$JSD has no base-model value, so $\tau$
uses four variants.}
\label{tab:diag}
\centering\small
\setlength{\tabcolsep}{3pt}
\begin{tabular}{@{}lccrrrrr@{}}
\toprule
 & & & \multicolumn{2}{c}{Kendall $\tau$} & \multicolumn{3}{c}{REAP vs REAM} \\
\cmidrule(lr){4-5}\cmidrule(l){6-8}
 & Better & & & & REAP & quality & same \\
Diagnostic & when & Tasks & mean & $> 0$ & better & agrees & order \\
\midrule
Cross-perplexity & lower & all & +0.70 & 9/10 & 20/20 & 18/20 & 18/20 \\
Completion JSD$^a$ & lower & all & +0.50 & 7/10 & 20/20 & 18/20 & 18/20 \\
Effective rank & higher & all & +0.44 & 6/10 & 15/20 & 13/15 & 13/20 \\
Distinct-2, no loops & higher & gen. & +0.15 & 3/4 & 1/8 & 1/1 & 1/8 \\
Distinct-2, with loops & higher & gen. & +0.70 & 4/4 & 6/8 & 6/6 & 6/8 \\
Loop rate & lower & gen. & +0.74 & 4/4 & 7/8 & 7/7 & 7/8 \\
Own perplexity & lower & all & +0.80 & 10/10 & 18/20 & 18/18 & 20/20 \\
\bottomrule
\end{tabular}

\end{table}

\begin{table}[!htbp]
\caption{\textbf{Cross-perplexity and completion-logit JSD} per task.}
\label{tab:xppl}
\centering\scriptsize
\resizebox{\textwidth}{!}{\begin{tabular}{@{}l rrrrr rrrrr@{}}
\toprule
 & \multicolumn{5}{c}{Cross-perplexity} & \multicolumn{5}{c}{Completion-logit JSD (bits)} \\
\cmidrule(lr){2-6}\cmidrule(l){7-11}
Model & Website & Brand & Categorisation & Domain & Email & Website & Brand & Categorisation & Domain & Email \\
\midrule
Qwen3-30B & 1.000 & 1.042 & 1.038 & 1.19 & 1.048 & -- & -- & -- & -- & -- \\
\quad REAP 25\% & 1.001 & 1.073 & 1.11 & 1.50 & 1.065 & 0.000 & 0.008 & 0.013 & 0.056 & 0.005 \\
\quad REAP 50\% & 1.005 & 1.15 & 1.32 & 4.00 & 1.12 & 0.000 & 0.024 & 0.034 & 0.168 & 0.018 \\
\quad REAM 25\% & 1.003 & 1.086 & 1.30 & 2.13 & 1.13 & 0.001 & 0.010 & 0.030 & 0.101 & 0.018 \\
\quad REAM 50\% & 1.009 & 1.16 & 1.57 & 5.90 & 1.20 & 0.001 & 0.029 & 0.049 & 0.215 & 0.029 \\
\midrule
Nemotron & 1.003 & 1.074 & 1.11 & 2.20 & 1.089 & -- & -- & -- & -- & -- \\
\quad REAP 25\% & 1.004 & 1.16 & 1.21 & 3.16 & 1.19 & 0.000 & 0.023 & 0.018 & 0.062 & 0.023 \\
\quad REAP 50\% & 1.007 & 1.37 & 1.40 & 7.11 & 1.36 & 0.001 & 0.052 & 0.051 & 0.060 & 0.045 \\
\quad REAM 25\% & 1.004 & 1.26 & 1.35 & 7.21 & 1.34 & 0.001 & 0.040 & 0.047 & 0.142 & 0.038 \\
\quad REAM 50\% & 1.008 & 1.42 & 1.63 & 18.31 & 1.45 & 0.002 & 0.059 & 0.071 & 0.123 & 0.048 \\
\bottomrule
\end{tabular}
}
\end{table}

\begin{table}[!htbp]
\caption{\textbf{Distinct-2 (repetition loops excluded) and effective rank of the pre-logit embeddings.}}
\label{tab:divrank}
\centering\scriptsize
\resizebox{\textwidth}{!}{\begin{tabular}{@{}l rrrr rrrrr@{}}
\toprule
 & \multicolumn{4}{c}{distinct-2} & \multicolumn{5}{c}{Effective rank of pre-logit embeddings} \\
\cmidrule(lr){2-5}\cmidrule(l){6-10}
 & \multicolumn{2}{c}{Domain descr.} & \multicolumn{2}{c}{Email templates} & & & & & \\
Model & within & across & within & across & Website & Brand & Categorisation & Domain & Email \\
\midrule
Qwen3-30B & 1.000 & 0.378 & 0.950 & 0.690 & 1{,}184 & 1{,}242 & 1{,}413 & 892 & 1{,}272 \\
\quad REAP 25\% & 1.000 & 0.359 & 0.950 & 0.698 & 1{,}165 & 1{,}230 & 1{,}396 & 876 & 1{,}257 \\
\quad REAP 50\% & 0.998 & 0.332 & 0.928 & 0.696 & 1{,}104 & 1{,}180 & 1{,}332 & 839 & 1{,}226 \\
\quad REAM 25\% & 1.000 & 0.372 & 0.934 & 0.702 & 1{,}154 & 1{,}215 & 1{,}379 & 873 & 1{,}241 \\
\quad REAM 50\% & 0.998 & 0.346 & 0.912 & 0.664 & 1{,}091 & 1{,}150 & 1{,}289 & 832 & 1{,}207 \\
\midrule
Nemotron & 0.998 & 0.494 & 0.880 & 0.572 & 1{,}709 & 1{,}764 & 1{,}899 & 1{,}105 & 1{,}911 \\
\quad REAP 25\% & 0.995 & 0.492 & 0.907 & 0.622 & 1{,}704 & 1{,}757 & 1{,}878 & 1{,}092 & 1{,}909 \\
\quad REAP 50\% & 0.994 & 0.477 & 0.817 & 0.491 & 1{,}688 & 1{,}734 & 1{,}824 & 1{,}074 & 1{,}909 \\
\quad REAM 25\% & 0.997 & 0.557 & 0.899 & 0.625 & 1{,}701 & 1{,}754 & 1{,}878 & 1{,}130 & 1{,}912 \\
\quad REAM 50\% & 0.988 & 0.572 & 0.821 & 0.530 & 1{,}709 & 1{,}732 & 1{,}815 & 1{,}116 & 1{,}910 \\
\bottomrule
\end{tabular}
}
\end{table}

\FloatBarrier
\section{Calibration}\label{app:calibration}

\tabref{tab:calib} describes the calibration set, tokenised with the Qwen3 tokenizer and chat template, as compression
saw it.

\begin{table}[!ht]
\caption{\textbf{Calibration composition per task.} Shares of tokens in system prompts, user messages and answers;
share of tokens from English rows; a lower bound on the share of English text (English rows plus the English system
prompts of the other rows); distinct system prompts detected as English.}
\label{tab:calib}
\centering\small
\resizebox{\textwidth}{!}{\setlength{\tabcolsep}{3pt}
\begin{tabular}{@{}lrrrrrrrrr@{}}
\toprule
 & & Tokens & \multicolumn{3}{c}{Share of tokens (\%)} & \multicolumn{3}{c}{English (\%)} & Sys.\ pr. \\
\cmidrule(lr){4-6}\cmidrule(lr){7-9}
Task & Rows & (k) & system & user & answer & rows & tokens & text $\geq$ & En. \\
\midrule
Website cues & 1{,}103 & 592 & 0.2 & 94.8 & 5.0 & 77.9 & 74.6 & 74.6 & 1/1 \\
Brand names & 272 & 593 & 59.4 & 39.7 & 0.9 & 18.8 & 18.5 & 66.7 & 2/2 \\
Categorisation & 1{,}067 & 592 & 20.9 & 76.6 & 2.5 & 20.2 & 15.1 & 31.9 & 4/4 \\
Domain descr. & 3{,}174 & 592 & 64.3 & 24.3 & 11.4 & 71.0 & 70.0 & 88.8 & 2/2 \\
Email templates & 50 & 592 & 74.6 & 1.4 & 24.0 & 18.0 & 17.6 & 79.2 & 50/50 \\
\midrule
All tasks & 5{,}666 & 2{,}962 & 43.9 & 47.4 & 8.7 & 59.8 & 39.1 & 68.2 & 59/59 \\
\bottomrule
\end{tabular}
}
\end{table}

\FloatBarrier
\section{Implementation details}\label{app:implementation}

\paragraph{Models.} Qwen3-30B-A3B-Instruct-2507 and NVIDIA-Nemotron-3.5-Lightning-30B-A3B-BF16, at pinned Hugging Face
revisions. Nemotron's multi-token-prediction layer is not used, so base and compressed models decode the same way.

\paragraph{REAP.} llm-compressor~0.15's \texttt{REAPPruningModifier}, run with the sequential pipeline:
\begin{itemize}
  \item each MoE layer is calibrated on the outputs of the already compressed layers before it;
  \item the same number of experts is removed in every MoE layer;
  \item the router rows of the removed experts are deleted.
\end{itemize}

\paragraph{REAM port.} The original REAM code supports only Qwen-style layers: the MoE block at \texttt{layer.mlp},
gated experts, a softmax router and attention-only decoder layers. Our port re-implements it as an llm-compressor
modifier on top of REAP's calibration code, so it also covers Nemotron: Mamba layers, gateless squared-ReLU experts, a
sigmoid router with bias and a shared expert.

The algorithm, per MoE layer:
\begin{enumerate}
  \item one batched pass computes router logits, gate-weighted outputs and activations of every expert on every
        calibration token;
  \item the most salient experts become centres;
  \item in order of saliency, each centre takes the most similar unassigned experts, up to a group of $C$ including
        itself ($C=16$ at 25\%, 32 at 50\%);
  \item members are aligned to their centre by Hungarian matching on activation and weight costs and averaged with
        saliency weights;
  \item the router keeps the centres' rows;
  \item the next layer is calibrated on the merged layer's outputs.
\end{enumerate}

\textbf{Deliberate deviations from the original code:}
\begin{itemize}
  \item router scores use the model's own scoring function (sigmoid for Nemotron);
  \item all experts of a layer are compared on the same sampled tokens;
  \item saliency is computed from the router's actual top-$k$ weights;
  \item the group size counts the centre (REAM's text reads ``up to $C$ unassigned non-centroid experts'');
  \item grouped routing and multi-GPU calibration are not supported; neither model needs them.
\end{itemize}

\textbf{Validation:}
\begin{itemize}
  \item unit tests check that grouping and the principal-component computations match the original functions;
  \item a neuron-shuffled copy of an expert is aligned back and merges into the original;
  \item all compressed checkpoints load without missing or unexpected weights.
\end{itemize}

\paragraph{Determinism.} Without deterministic kernels, two REAM runs of the same configuration agreed in only 18 of 48
Qwen3-30B layers. With \texttt{torch.use\_deterministic\_algorithms(True)} two runs are bit-identical, at about 25\%
extra time, and all runs use it.

\paragraph{Generation and scoring.}
\begin{itemize}
  \item \textbf{Serving:} vLLM~0.32 with BF16 weights.
  \item \textbf{Decoding:} greedy for website cues, brand names and categorisation. Each model's default sampling for
        domain descriptions and email templates: Qwen3-30B temperature 0.7, top-$p$ 0.8, top-$k$ 20; Nemotron
        temperature 1.0, top-$p$ 0.95. The seed is derived from the row identifier.
  \item \textbf{Structured outputs:} brand names and categorisation use the production JSON schemas with structured
        outputs (XGrammar).
  \item \textbf{Length caps:} 1,024 answer tokens for the schema-constrained tasks, 32,768 for the others.
  \item \textbf{Judging:} categorisation and domain descriptions use the production judge with the product's judge
        instructions. A cache keyed by row and answer text gives identical answers identical verdicts.
  \item \textbf{Email templates:} scored with the production rubric library, with rendered checks in headless Chromium.
\end{itemize}

\paragraph{Hardware and software.} 4~$\times$ NVIDIA RTX PRO 6000 Blackwell (96\,GB); every compression and generation
run used one GPU. Python 3.12 throughout, with these stacks:
\begin{itemize}
  \item compression: llm-compressor~0.15, PyTorch~2.14, Transformers~5.21;
  \item generation: vLLM~0.32, PyTorch~2.13, Transformers~5.19;
  \item probe and analysis: PyTorch~2.14, Transformers~5.21.
\end{itemize}

\FloatBarrier
\end{appendix}

\end{document}